\documentclass[review]{elsarticle}
\graphicspath{ {./figures/} }
\usepackage{float}
\usepackage{verbatim} 
\usepackage{apalike}
\restylefloat{figure}
\floatstyle{plaintop} 
\restylefloat{table}

\usepackage{lineno}
\modulolinenumbers[5]
\usepackage{amsfonts}
\usepackage[colorlinks = true,
            linkcolor = blue,
            urlcolor  = blue,
            citecolor = blue,
            anchorcolor = blue]{hyperref}

\usepackage{enumitem}
\usepackage{graphicx}
\usepackage{caption}
\usepackage{subfig}
\usepackage{multirow}
\usepackage{multicol}
\usepackage{makecell}
\usepackage{longtable}
\usepackage{lipsum}
\usepackage{url}
\usepackage{booktabs}
\usepackage[table,xcdraw]{xcolor}
\usepackage{makecell}
\usepackage{mathtools}
\usepackage[left=2.5cm, right=2.5cm, top=2.5cm]{geometry}
\usepackage{algorithm}
\usepackage{algorithmic}
\newcolumntype{P}[1]{>{\centering\arraybackslash}p{#1}}

\usepackage{float}
\usepackage{multirow}
\usepackage{longtable}
\usepackage{tabulary}
\usepackage{array}
\usepackage{booktabs}
\usepackage[referable]{threeparttablex}
\usepackage{caption}

\makeatletter
\newcommand*\TY@cap@gobble[2][]{\\}
\def\ltabulary{%
    \def\caption{
        \@ifstar\TY@cap@gobble\TY@cap@gobble}
    \def\endfirsthead{\\}%
    \def\endhead{\\}%
    \def\endfoot{\\}%
    \def\endlastfoot{\\}%
    \def\tabulary{%
        \def\TY@final{%
    \def\endfirsthead{\LT@end@hd@ft\LT@firsthead}%
    \def\endhead{\LT@end@hd@ft\LT@head}%
    \def\endfoot{\LT@end@hd@ft\LT@foot}%
    \def\endlastfoot{\LT@end@hd@ft\LT@lastfoot}%
    \longtable}%
        \let\endTY@final\endlongtable
        \TY@tabular}%
    \dimen@\columnwidth
    \advance\dimen@-\LTleft
    \advance\dimen@-\LTright
    \tabulary\dimen@}

\makeatother
\newcolumntype{L}[1]{>{\raggedright\let\newline\\\arraybackslash\hspace{0pt}}m{#1}}
\newcolumntype{C}[1]{>{\centering\let\newline\\\arraybackslash\hspace{0pt}}m{#1}}
\newcolumntype{R}[1]{>{\raggedleft\let\newline\\\arraybackslash\hspace{0pt}}m{#1}}

\journal{Biotechnology Advances}

\biboptions{authoryear}

\begin{document}
\begin{frontmatter}











\title{Machine Learning Methods for Small Data and Upstream Bioprocessing Applications: A Comprehensive Review}

\author[uts]{Johnny Peng \corref{cor1}}
\ead{johnny.peng@student.uts.edu.au}

\author[uts]{Thanh Tung Khuat}
\ead{thanhtung.khuat@uts.edu.au}

\author[uts]{Katarzyna Musial}
\ead{Katarzyna.Musial-Gabrys@uts.edu.au}

\author[uts]{Bogdan Gabrys}
\ead{bogdan.gabrys@uts.edu.au}

\cortext[cor1]{Corresponding author.}
\address[uts]{Complex Adaptive Systems Laboratory, The Data Science Institute, University of Technology Sydney, NSW 2007, Australia}

\begin{abstract}
Data is crucial for machine learning (ML) applications, yet acquiring large datasets can be costly and time-consuming, especially in complex, resource-intensive fields like biopharmaceuticals. A key process in this industry is upstream bioprocessing, where living cells are cultivated and optimised to produce therapeutic proteins and biologics. The intricate nature of these processes, combined with high resource demands, often limits data collection, resulting in smaller datasets. This comprehensive review explores ML methods designed to address the challenges posed by small data and classifies them into a taxonomy to guide practical applications. Furthermore, each method in the taxonomy was thoroughly analysed, with a detailed discussion of its core concepts and an evaluation of its effectiveness in tackling small data challenges, as demonstrated by application results in the upstream bioprocessing and other related domains. 
By analyzing how these methods tackle small data challenges from different perspectives, this review provides actionable insights, identifies current research gaps, and offers guidance for leveraging ML in data-constrained environments.
\end{abstract}


\end{frontmatter}

\section{Introduction}
\subsection{Small Data}
Machine learning (ML) has been increasingly utilised in various domains due to its ability to learn from data and make accurate predictions or decisions. As the volume of available data grows, the performance of ML models often improves, leading to more precise and reliable outcomes. Similarly, the accuracy and reliability of ML models often decrease as the available data becomes limited, and once the data size is smaller than a certain threshold, building a reliable and accurate ML model becomes difficult or even impossible with standard ML methods, and more domain-specific methods with potential additional usage of meta- or transfer learning mechanisms \cite{lebu15,albu20}, if possible, might be required. But how "small" is "small"? And how can we define small data?  A rule of thumb proposed in \citet{va98} is that if the ratio of data size to Vapnik–Chervonenkis (VC) dimension is smaller than 20, then the dataset can be considered as small data. Other ways to define small data are discussed in \citet{alcr18}, such as for every prediction class, there should be at least 50 to 1000 data points, or for every prediction class, there should be at least 10 to 100 data points, or for every weight in a neural network, there should be at least 10 data points. A recent study published by DeepMind \citep{hobo22} pushed the limits of these rules of thumb and proved that they are still applicable for highly complex models with billions of parameters, such as Large Language Models (LLM). In this paper, the Chinchilla scaling law was proposed. It suggests that for LLM, every model parameter would require roughly 20 data points (text tokens) to train it optimally. In that sense, billions of text tokens are still considered small for many LLMs with billions of parameters. For example, even though GPT-3 was trained on 300 billion text tokens, given it has 175 billion parameters, for it to be optimally trained, it would need to be trained on roughly 3.5 trillion tokens. 

Despite these guidelines and rules of thumb, determining whether a dataset is small for an ML task in practice is quite difficult. However, these papers provided important insights into what needs to be considered in a definition of small data. Firstly, as we saw in the Chinchilla paper \citep{hobo22}, whether a dataset could be considered small is relative to the complexity of the underlying ML model. Selecting a suitable model with complexity matching the available training data is important when building an ML model. Otherwise, the performance of the final model might be compromised. Secondly, whether a dataset could be considered small should also be relative to the complexity of the underlying ML task. Several billion data points might be a lot for many common ML tasks. However, it is insufficient for training LLM models to perform tasks involving mathematics or requiring strong reasoning capabilities \citep{hobo22}. Thus, instead of providing a rigid threshold to distinguish small data from non-small data, we define small data in the context of machine learning (ML) as follows: "For an ML task, small data refers to a dataset that contains fewer data points than what is typically required to accomplish such a task." While the term "typical" might appear subjective and influenced by individual experience, this definition is flexible and robust enough to serve as a guiding principle for identifying instances where small data limitations are present, enabling the application of appropriate ML methods to address these challenges. 


Following this definition, it is not difficult to find small data in many real-world applications, as acquiring large datasets can be challenging due to various factors such as privacy concerns, resource constraints, or the rarity of certain events. This limitation of data is the number one barrier that limits the adoption of AI and ML models in various industries \citep{Ng22}. The biopharmaceutical industry is an example of a sector that suffers from this small data issue. In biopharmaceutical companies, novel bioproducts often have limited process data \citep{tuga18, baal21}. At the same time, a wide range of measurements are collected during the process, contributing to a Low-N problem \citep{tuga18} where the number of training samples is much smaller than the data dimensions, which makes an ML model training task much more difficult and increases the risk of training data overfitting. The issue is exacerbated by products being moved to various production sites during their lifecycle, leading to a limited number of historical experimental data points at each site. At the same time, the increased use of real-time Process Analytical Technology (PAT) tools leads to a large dimensionality of collected data with a limited number of offline measurements as labels \citep{baal21}, further worsening the Low-N problem. 


\subsection{Prior Work}
Given the importance of addressing small data problems, there is a growing interest in the ML research community for developing specialised ML methods that can effectively learn from limited data, with many surveys and reviews on this topic being carried out in recent years. \citet{cabu22} discussed various approaches for learning from small data, such as active learning and few-shot learning, emphasising the need for theoretical guarantees on their generalisation performance. The authors introduced a PAC (Probably Approximately Correct) framework to analyse small data learning, focusing on label complexity and generalisation error bounds. A key contribution is their exploration of geometric representation perspectives, categorising small data learning models into Euclidean and non-Euclidean (hyperbolic) representations. The survey also highlighted the impact of small data learning on multiple AI domains, such as transfer learning, contrastive learning, and meta-learning, and identified challenges like weak supervision and imbalanced distributions. \citet{shxu18} presented a comprehensive survey on Small Sample Learning (SSL) or Few Sample/Shot Learning (FSL). The authors categorized SSL techniques into two main approaches: \textit{concept learning}, which focuses on recognizing and forming new concepts from limited observations by leveraging prior knowledge (such as recognition, generation, and imagination), and \textit{experience learning}, which works alongside large-sample learning to optimize performance when training data is scarce. The paper explored neuroscience evidence supporting SSL, highlighting similarities with human cognition, such as episodic memory, imagination, and compositionality. It also discussed various SSL methodologies, including meta-learning, transfer learning, data augmentation, and generative models, and identified key challenges and future research directions in the field. \citet{lugo23} provided a comprehensive review of FSL, covering over 300+ FSL-related papers from 2000 to 2019, categorising FSL methods into generative model-based and discriminative model-based approaches. A key focus of this paper is meta-learning, which has become the dominant paradigm for FSL, with strategies such as Learn-to-Measure, Learn-to-Finetune, Learn-to-Parameterise, Learn-to-Adjust, and Learn-to-Remember. The survey also covered emerging extensions of FSL, including semi-supervised, unsupervised, cross-domain, generalised, and multimodal FSL. Additionally, the paper highlighted FSL applications across various domains like computer vision, natural language processing, audio processing, reinforcement learning, and robotics, discussing benchmark performance and future research directions.

Some reviews have focused on certain types of techniques, such as data augmentation, for addressing small data issues. \citet{lamu17} provided a comprehensive review of the Fuzzy Theory-based methods for generating artificial data, which discusses the commonly used Fuzzy Theory-based techniques such as Mega-Trend Diffusion and Fuzzy C-Means clustering for generating synthetic data to solve small data problems. Bootstrapping was also mentioned as an alternative method, with the obvious disadvantage of not being able to add sample diversity to the data or reduce the information gaps between samples. In \citep{lide18}, the usage of Mega Trend Diffusion models, Bootstrapping, and the Radial Basis Function Neural Network for generating artificial data was again discussed. In addition, this review also discussed the pros and cons of supervised learning models that work well on small data, such as Support Vector Machine, K-Nearest Neighbour, and Bayesian methods. \citet{ra18} discussed techniques for model evaluation and selection under limited data scenarios, which are particularly important when working with small datasets. It covered methods like variants of cross-validation and bootstrapping, providing practical guidance for choosing appropriate models and algorithms when data is limited. 

There have also been reviews that focus on applications of small data ML methods, \citet{pach22} provided a comprehensive review of various Generative Adversarial Network (GAN)  methods and their application in mechanical fault diagnosis with small samples. \citet{rofa22} and \citet{mosu22} provided a comprehensive review of ML methods addressing small data problems in automatic speech processing and skin disease analysis, respectively, which includes many other emerging methods such as transfer learning and meta-learning. However, there is a lack of reviews on applications of small data ML methods for the applications in the domain of upstream bioprocessing, an application area where data is usually limited. Thus, this review paper effectively addresses this research gap and provides practical guidance for addressing small data issues in the domain of upstream bioprocessing and other related fields.

The rest of the paper is organised as follows: In Section \ref{sec:introduction}, we begin by introducing upstream bioprocessing and explore the types of data involved, along with the underlying causes of the small data problem in this context. Then, in section \ref{sec:taxonomy}, we introduce a novel taxonomy that offers a comprehensive perspective on various machine learning methods applicable to addressing small data challenges at different stages of a typical ML workflow. 
In sections~\ref{sec:data}, \ref{sec:model-training} and \ref{sec:model-retraining}, we examine the key ideas of each method, evaluate their efficacy in tackling limited data challenges, and discuss their existing and potential applications for addressing small data challenges in the context of upstream bioprocess and other related fields. Section \ref{discusions} summarises the key methods for handling the small data problem and proposes potential research directions. Lastly, in section~\ref{sec:conclusion}, we conclude this study and highlight potential future research directions for ML methods discussed in this study.

\section{Small Data Problem in Upstream Bioprocessing}
  \label{sec:introduction}
    Upstream bioprocessing encompasses the initial stages of biopharmaceutical production, focusing on cultivating cells or microorganisms to generate the desired product, such as a protein or antibody. This involves crucial steps like cell line selection, media optimisation, cell cultivation within bioreactors, and harvesting \citep{khba24}. This complex process involves various controlled variables, including temperature, ph, and nutrient levels, to maximise product yield and quality \citep{rofe10}. Process Analytical Technology (PAT) is crucial in monitoring and controlling these critical process parameters in real-time, providing valuable insights for optimisation \citep{stma13}. One powerful PAT tool is Raman spectroscopy, a non-invasive technique that provides detailed chemical information about the bioprocess by analysing the interaction of light with molecules \citep{escu17}. Raman spectroscopy offers the potential for real-time monitoring of various critical components in the bioreactor, including nutrients, metabolites, and even the product itself when integrated with advanced analytics like ML \citep{escu17, khba24}. 

    \begin{ThreePartTable}
          \begin{ltabulary}{P{2cm}P{3cm}P{9cm}}
          \caption{Type of data generated by bioreactor runs. \label{comparisons_table}}\\
            \toprule
            \textbf{Data Type} & \textbf{Example Variables} & \textbf{Description} \\
            \midrule
            Raman Spectrums $(R(t))$ & Spectral data from the Raman spectrometer often ranges from 100 to 3325 $cm^{-1}$ & Raman spectrometer can measure Raman spectrums of the culture medium in near real-time, generating a large volume of Raman data from each run. However, meaningful information or insights cannot be read or extracted from them easily. A separate Raman model (typically ML models) is required to convert these raw data into meaningful measurements. \\
            \midrule
            Online Measurements $(O(t))$ & Temperature, ph, Dissolved Oxygen, Weight, Agitator power, Off-gas O2 \& CO2 & These measurements are generated in real-time from hardware online sensors in the bioreactor. These sensors ensure the process is going as expected, and adjustments can be made as needed. Some of these online measurements, such as Substrate flow rate, Acid/Base flow rate, cooling water flow rate, etc., are controlled directly and precisely by manual or automatic controls. These controls ensure that the behaviour of the controlled variables is aligned with the instructions from the system controls. Otherwise, it could be an indication of faulty controls. \\
            \midrule
            Offline Measurements $(Y(t))$ & Cell Concentration and Viability, Biomass \& Metabolite Concentration & These are the measurements that cannot be observed directly from hardware online sensors. Instead, skilled labour is required to manually take a sample from the bioreactor and perform offline analysis to obtain these measurements. Alternatively, these measurements can be inferred by Raman spectra with a properly calibrated Raman model.  \\
            \bottomrule
          \end{ltabulary} 
          \label{table:datatype}
        \end{ThreePartTable}
    
    However, the full potential of Raman spectroscopy and other PATs is often hampered by the limited data available in upstream bioprocessing. The high cost and time investment required for each bioreactor run restrict the amount of data that can be generated \citep{suge22, siwo20}. While PAT provides valuable real-time measurements, the overall dataset size, especially offline analytical measurements used as ground truth,  remains relatively small compared to the requirements of data-intensive machine learning algorithms \citep{hehe23,dubo23}. This data scarcity limits the development of accurate predictive models and hinders the optimisation of control strategies based on PAT data. The complexity of biological systems further exacerbates this challenge, making it difficult to generalise from limited data even with the insights provided by PAT. This highlights the need for innovation in developing and leveraging machine learning methods specifically designed to learn effectively from small datasets, enabling more robust and reliable applications of PAT and advanced analytics in upstream bioprocessing. To provide more background on the small data problem in upstream bioprocessing, we have summarised the common types of time series data generated from upstream bioprocesses in Table \ref{table:datatype}, where $t$ represents the time step at which the data was captured.
    
    Out of these types of data, the most important one is the offline analytical results $(Y)$ that measure the Critical Quality Attributes (CQAs) and Key Performance Indicators (KPIs), which are the key indicators of the quality of the products. Naturally, the most important learning tasks focus on understanding and predicting these offline analytical results. For example, given the data generated from historical and current bioreactor runs ($R, O, Y$ time series), one potential ML application is to explore and develop soft sensors that accurately map $R(t)$, $O(t)$, $Y(t-1,t-2,t-3...)$ to $Y(t)$ by learning from historical data and current feedback. As frequently illustrated in the broader process industry area \cite{kaga09,kagr11}, such soft sensors enable near real-time monitoring for rapid identification of process issues, such as missed metabolic shifts and potential contamination, allowing for automatic, adaptive controls to improve product quality and consistency. Furthermore, it offers a cheaper, less labour-intensive solution than traditional offline methods, reducing manual effort and waste. However, these soft sensors are difficult to build as offline analysis is often performed infrequently due to the high labour cost, which leads to limited labelled data for related learning tasks, which could lead to poorly trained ML models that are potentially overfitted. Thus, with limited data and feedback, accurate and robust machine-learning models are challenging to construct for upstream bioprocess applications.

\section{Proposed Taxonomy for ML Methods for Small Data}
\label{sec:taxonomy}
\begin{figure}
                \centering
                \includegraphics[width=1\linewidth]{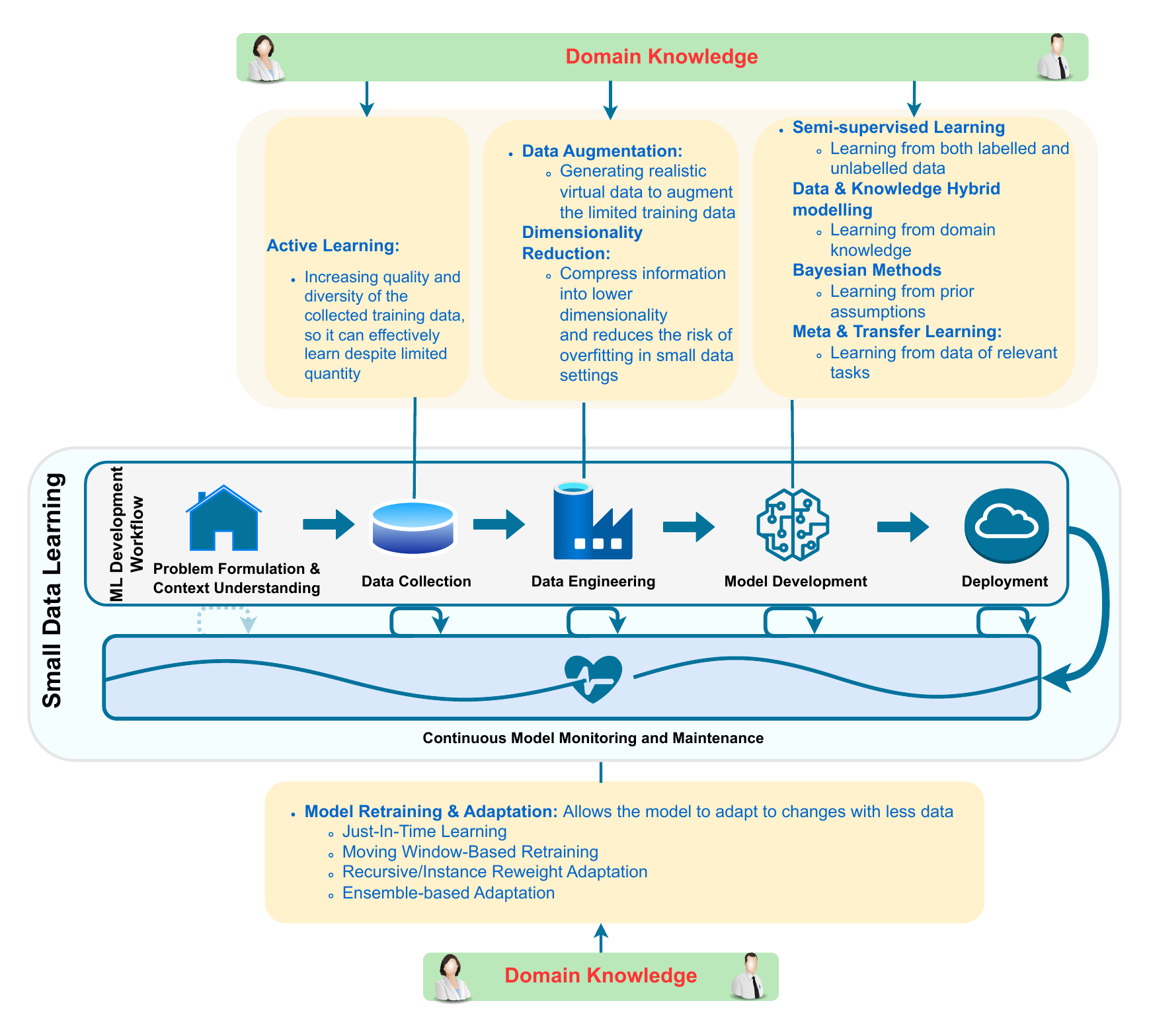}
                \caption{Machine learning methods for small data across ML development cycle}
                \label{our_taxonomy}
    \end{figure}

The existing surveys have already introduced various taxonomies for ML methods that focus on addressing small data issues. \citet{pach22} broadly classified the methods for handling small data by the type of techniques: data augmentation-based, transfer learning/domain adaptation-based and model-based strategies. In \citet{waya20}, the authors classified the FSL methods based on how they incorporated prior knowledge and helped reduce the hypothesis space. However, all these taxonomies focus on the technical differences between individual methods rather than how they could be applied together in practice, creating difficulties for ML practitioners to apply these methods holistically. Furthermore, as shown in the review \citep{mosu22} on skin disease analysis, many studies use more than one ML method to handle small data problems and achieve better model performance. Thus, it is important to review ML methods for small data holistically across all aspects of the ML workflow, allowing the identification of more opportunities and aspects that can address the small data issue. For this reason, we introduce a new taxonomy (Figure~\ref{our_taxonomy}) that classifies these ML methods based on the three main ML steps of a typical ML workflow \citep{kemu24,khke23,boho22}, including Data Collection \& Engineering, Model Development, and Continuous Monitoring \& Maintenance. This taxonomy emphasises the integration of techniques across the entire ML process, enabling practitioners to systematically tackle the limitations of small data at different workflow stages.

To address small data issues during the data collection and engineering stage, the focus is on optimising the input data to improve the model's ability to learn from limited samples \citep{gape04}. Methods include \textit{Active Learning} \citep{se09}, which prioritises informative samples to minimise labelling effort while maximising information gained, making it especially useful when annotated data is scarce. Another key method is \textit{Data Augmentation} \citep{mumu25}, which expands the training dataset by creating plausible variations, such as synthetic data generation and data transformation, thereby enhancing the model's generalisation capabilities. Finally, \textit{Dimensionality Reduction} \citep{boda24, ghsa19}, using techniques such as Principal Component Analysis (PCA) or t-distributed Stochastic Neighbor Embedding (t-SNE), reduces the input feature space to ensure that only the most relevant information is used for training, helping to prevent overfitting in small data scenarios.

During the model training step, our focus is on developing strategies to train effective models despite limited data. One such approach is \textit{Semi-Supervised Learning} \citep{vaho20}, which leverages both labelled and unlabelled data to maximise the utility of available information, effectively bridging the gap between small and large datasets \cite{gape04}. Another promising method is \textit{Data \& Knowledge-Driven Hybrid modelling} \citep{brki22}, which integrates domain knowledge with data-driven techniques. By incorporating prior expertise, this approach helps compensate for insufficient training data while improving model reliability. Similarly, \textit{Bayesian Methods} \citep{gi15, shsw16} provide a probabilistic framework that is well-suited for small data settings, as they incorporate prior distributions to enhance learning with limited samples \citep{maio23,amnj23,kata22}. Beyond probabilistic techniques, \textit{Meta \& Transfer Learning} \citep{paya10, huva21} enables models to generalise effectively across tasks with minimal data by leveraging knowledge transferred from related learning tasks and datasets. 


The final stage focuses on methods that ensure continuous improvement and adaptability of the model, particularly in data-constrained scenarios where each new data point can significantly enhance model performance. To achieve this, one of the most common techniques is to use the \textit{Moving Window} \citep{likr09,lejo05} approach. In this case, a model can be fully or incrementally retrained using a sliding window of recent data, allowing it to adapt to evolving patterns while maintaining temporal relevance. Similarly, \textit{Online Learning} \citep{poud01} processes data sequentially, updating the model in real-time, making it well-suited for dynamic environments with small, incremental data streams. Additionally, \textit{Ensemble Learning} or the \textit{Mixture of Experts} approach \citep{saro18,doyu20} enhances adaptability by combining multiple models to reduce variance and bias, resulting in more robust predictions in changing conditions. Finally, \textit{Just-In-Time Learning} \citep{kafU13a, sa14, jich15,yuge14} refines predictions by dynamically selecting the most relevant subset of data for training, ensuring highly adaptable, instance-specific outputs.

With methods covering the end-to-end ML workflow, this taxonomy offers a comprehensive and workflow-oriented perspective for addressing small data challenges. Unlike previous taxonomies focusing solely on technical distinctions, this framework highlights the integration and interplay of techniques at different ML stages. For example, a practitioner could combine \textit{data augmentation} in the data engineering stage, \textit{transfer learning} during training, and \textit{ensemble learning} in the update phase to achieve robust results. Moreover, this classification emphasizes the iterative nature of ML workflows and the end-to-end lifecycle of ML models, discussing methods such as \textit{online learning}, \textit{moving window} and \textit{just-in-time learning} to ensure that models remain adaptive and effective over time, even as data limitations persist. This taxonomy empowers practitioners to design more holistic and effective solutions for small data problems.

In the following sections, we introduce each method in the taxonomy and explore its specific applications in upstream bioprocessing. To extend the relevance beyond this domain, we also present a summary table highlighting key results from applications of these methods in addressing small data challenges across various other fields. By providing this broader perspective, we emphasise the robustness and versatility of these approaches, showcasing their potential for diverse applications beyond bioprocessing.


\section{ML Methods for Addressing Small Data Challenges during Data Collection and Engineering}  
\label{sec:data}    
    Data collection and engineering is typically the first step of a standard ML project or workflow, and it involves collecting the data and then transforming it into a pre-processed dataset that can be used for model training. Data collection is crucial as the amount and diversity of the data collected would significantly impact downstream tasks. We already discussed many issues with small data, but having data that is not diverse and representative could worsen the issue. To prevent this from happening,  most experimental designs for bioprocess applications are done through the Design of Experiments (DoE), an approach which is becoming increasingly popular \citep{kaav21}. However, to generate data for building ML models, DoE could be subjective and may waste time experimenting with factors that are not useful for ML tasks. In the following sections, we will introduce an \textbf{Active Learning} (AL) technique, which allows the ML algorithms to activelly and incrementaly query humans or systems to generate the most useful data for the ML model to learn from, maximizing the performance and frequently reducing the associated uncertainty of the ML models trained based on such data.

    Although collecting more data is the most direct way to solve small data issues, collecting and labelling additional data could be costly, even with the help of AL. A cost-free alternative is \textbf{Data Augmentation} (DA), which can be applied during the Data Engineering stage to increase the volume and diversity of the dataset by automatically generating new and realistic data instances. Apart from small data volume, high dimensionality could also contribute to the Low-N problem and cause the model to overfit the data. Thus, we could further alleviate the small data problem by performing \textbf{Dimension Reduction} (DR), which reduces the number of dimensions in the data while retaining a large portion of the useful information for training ML models. After introducing AL in the following subsections, we will examine the technical details of both DA \& DR and their applications for handling small data. 
    
    
\subsection{Active Learning}    
Active learning (AL) is a machine learning method that aims to minimise the time, effort, and cost of labelling data by selectively querying and labelling the most informative instances from a pool of unlabelled data \citep{se09}. Given a learning algorithm $\mathcal{A}$, a hypothesis class $\mathcal{H}$, and a large pool of unlabelled data $\mathcal{U}$, the goal of active learning is to find an optimal querying strategy $\mathcal{Q}$ that minimizes the number of queries required for achieving a desired level of performance on the target task. In this context, each query will be sent to an oracle, an entity (e.g., a human annotator or a system) that provides labels for the selected instances, which often has a cost involved. Thus, in practice, a query budget $\mathcal{B}$ is also applied, which stops the queries when it is reached. Formally, let $D = \{(x_1, y_1), \dots, (x_n, y_n)\}$ be a set of labelled instances, where $x_i$ represents the feature vector of the $i$-th instance and $y_i$ is its corresponding label. The AL process is formally defined as Algorithm~\ref{alg:active_learning}.
  \begin{algorithm}
    \caption{Active Learning (AL) Process}
    \label{alg:active_learning}
    \begin{algorithmic}[1]
    \REQUIRE Learning algorithm $\mathcal{A}$, Hypothesis class $\mathcal{H}$, Unlabelled dataset $\mathcal{U}$, Query budget $\mathcal{B}$
    \ENSURE Trained model $h$ with satisfactory performance
    
    \STATE \textbf{Initialize:} labelled dataset $D_0$, Unlabelled dataset $\mathcal{U}_0 \gets \mathcal{U}$, $t \gets 0$
    
    \WHILE{$t < \mathcal{B}$}
        \STATE \textbf{Train the model:} Train $h_t \in \mathcal{H}$ using $\mathcal{A}$ and the labelled dataset $D_t$
        \STATE \textbf{Evaluate the model:} Compute performance metric (e.g., accuracy, F1 score) for $h_t$
        \IF{Performance of $h_t$ is satisfactory}
            \STATE \textbf{Stop and return:} $h_t$
        \ENDIF
        \STATE \textbf{Query new instance:} 
        \STATE Select instance $x_{t+1}^*$ from $\mathcal{U}_t$ using querying strategy $\mathcal{Q}$
        \STATE Query the label $y_{t+1}^*$ for $x_{t+1}^*$
        \STATE \textbf{Update datasets:}
        \STATE $D_{t+1} \gets D_t \cup \{(x_{t+1}^*, y_{t+1}^*)\}$
        \STATE $\mathcal{U}_{t+1} \gets \mathcal{U}_t \setminus \{x_{t+1}^*\}$
        \STATE Increment $t$: $t \gets t + 1$
    \ENDWHILE
    \end{algorithmic}
    \end{algorithm}
    The most crucial component of the AL process is the querying strategy (line 9 of algorithm~\ref{alg:active_learning}), which evaluates the informativeness of all or sampled unlabelled instances and decides which one should be sent to the Oracle for labelling next. In \citet{thsc23}, a taxonomy has been proposed, which classified query strategies into four categories: Random (Passive Learning), Information-Based, Representation-Based, and Meta-active Learning. Query strategy can significantly affect the number of queries required to achieve the desired model performance. Based on the theoretical analysis in \citet{va84}, to train a simple linear classifier with maximum desired error rate $\epsilon$ based on a random query strategy, it would require $O(1/\epsilon)$ labels. However, with a simple active learning strategy using binary search, only $O(log1/\epsilon)$ labels would be required to achieve the same performance. A high-level description of these query strategies frameworks will be discussed below, and more technical details can be found in \citet{thsc23}.

    \begin{itemize}
        \item \textbf{Information-based Query Strategies} - A group of techniques that evaluate the informativeness of instances based on the contribution they make to improving the learning model's performance or reducing uncertainty in predictions. These strategies often quantify informativeness through probabilistic measures, such as entropy (measuring the uncertainty of a predicted label distribution) or mutual information  (assessing how much an instance reduces uncertainty about the model parameters or predictions). For example, \textit{uncertainty sampling} selects instances for which the model has the highest uncertainty in its predictions, such as those closest to the decision boundary in a classifier. \textit{Query-by-Committee (QBC)} evaluates disagreement among a committee of models trained on the current data, selecting instances that maximise disagreement to reduce hypothesis space. Another example is \textit{Expected Model Change}, which queries instances that are predicted to induce the greatest change in model parameters, thereby encouraging faster convergence to the optimal hypothesis.
        \item \textbf{Representation-based Query Strategies} - A group of techniques that select representative instances of the underlying data distribution, thereby improving the model's ability to generalise by ensuring that the labelled dataset reflects the diversity and structure of the entire dataset. By selecting instances that capture the variability, structure, or density of the unlabelled data, these strategies reduce the risk of bias toward overrepresented patterns and ensure better coverage of the input space. This leads to a more robust model that can perform well on unseen data. For instance, clustering-based methods select data points near cluster centroids to ensure diversity, while density-based methods prioritise instances in high-density regions, which are more likely to represent the core of the data distribution.
        \item \textbf{Meta-active Learning Query Strategies} - A group of techniques that leverage prior knowledge or meta-information about the learning model's behaviour, instance informativeness, or task-specific data characteristics to guide the selection of informative instances for labelling by an oracle. Meta-active learning query strategies exploit the experience gained from solving previous learning tasks by learning a meta-model or meta-learner that captures patterns or relationships between instances, their informativeness, and the resulting model performance across various tasks. This meta-knowledge is then transferred to new tasks, enabling meta-active learning strategies to accelerate the learning process by efficiently prioritizing queries that are expected to yield the highest benefit for the model.
    \end{itemize}

\subsubsection{Applications of Active Learning on Small Data in Upstream Bioprocessing}

Active learning (AL) is increasingly used in upstream bioprocessing to accelerate the Design-Build-Test-Learn (DBTL) cycle, particularly in small data scenarios where experimental costs, resource limitations, and biological variability constrain large-scale data collection. AL minimises experimental workload while maximising information gain, allowing models to learn efficiently from limited data by strategically selecting the most informative experiments rather than relying on exhaustive trial-and-error approaches \citep{fafa21, liba15}.  

One of the earliest applications of AL in bioprocess optimisation demonstrated its ability to extract maximal value from limited experimental runs. In \citet{macr11}, Bayesian AL was applied to optimise Penicillin G production using in silico simulations, achieving near-optimal process parameters within just three fed-batch runs—a 300\% improvement over exploratory trials. This highlights AL’s strength in small data environments, where traditional methods would require significantly more experiments to converge on an optimal solution. Similarly, in cell-free protein synthesis, \citet{boko20} tackled the challenge of batch-to-batch variability and a vast combinatorial space (~4 million buffer compositions). By identifying just 20 key buffer formulations, their AL-driven approach improved protein production 34-fold, demonstrating how AL can efficiently explore high-dimensional design spaces with limited experimental data. In mammalian cell culture optimisation, data scarcity is a persistent challenge due to the high cost and time requirements of cell-based experiments. \citet{haoz23} addressed this issue by integrating gradient-boosting decision trees with high-throughput assays, revealing optimal adjustments in serum, vitamin, and amino acid compositions while reducing the need for large-scale experimental screening.  

The integration of AL with automated experimentation further mitigates small data constraints by maximising information gain per experiment and reducing reliance on manual interventions. In \citet{kiwh04}, a "robot scientist" applied AL to dynamically select the most informative biological experiments, significantly cutting costs threefold lower than the cheapest standard approach and 100-fold lower than passive learning. This ability to selectively acquire the most valuable data is critical in bioprocessing, where the cost of generating labelled data (e.g., running fermentation trials or cell culture experiments) is high. Similarly, in \citet{hach19}, BioAutomata, a fully automated AL-driven DBTL system, was used to optimise lycopene biosynthesis. By testing less than 1\% of possible designs, BioAutomata achieved a 1.77-fold yield improvement, demonstrating how AL can navigate complex biological landscapes efficiently despite limited training data.  

Beyond experimental design, AL also addresses the small data challenge in strain engineering by guiding model-driven decision-making. \citet{raco20} introduced the Automated Recommendation Tool (ART), which applies Bayesian ensemble modelling to predict optimal strain designs with minimal experimental data. ART was validated across diverse metabolic engineering applications—including biofuels, fatty acids, and hoppy beer flavours without hops—demonstrating that highly efficient strain optimisation can be achieved even in data-sparse conditions. However, the study also underscored the importance of model refinement and careful assumption handling, as small data models are more sensitive to biases and generalisation errors.  

Across these studies, a common theme emerges: AL enables targeted experimentation, reduces reliance on large datasets, integrates seamlessly with automation, and optimises strain and process parameters with minimal data. By intelligently selecting the most informative experiments, AL makes upstream bioprocessing more efficient, scalable, and cost-effective under data- and resource-constraints environment.

Apart from accelerating the DBTL cycle, one promising but under-researched area is the application of AL for building Raman-based soft sensors for cell culture process monitoring. As discussed in Section~\ref{sec:introduction}, offline labelling of Raman spectra is labour-intensive and costly. AL can optimise this process by selectively determining which Raman samples are worth labelling, ensuring that each labelled instance contributes meaningful improvements to model performance. This approach minimises labelling costs, maximises information gain, and enhances the diversity and representativeness of small datasets, making soft sensors more reliable under resource-constrained conditions.  

Despite the lack of research on AL for Raman-based soft sensors in bioprocessing, AL-driven soft sensor development has been extensively studied in other domains. A common strategy involves leveraging feature subspaces to evaluate the informativeness of unlabelled samples. In \citet{ge14}, an AL mechanism was integrated into a Principal Component Regression (PCR) based soft sensor using a representation-based query strategy. By assessing both the principal component and residual subspaces, the model prioritised samples that improved predictive accuracy and stability in an industrial case study. Similarly, \citet{wuch17} combined Kernel Partial Least Squares (KPLS) with a Gaussian Process (GP) model to measure uncertainty, triggering sample selection when uncertainty exceeded a predefined threshold. Instead of random selection, the system dynamically updated its training set with the most relevant labelled instances, improving adaptability in online predictions. Other approaches refine sample selection by incorporating both prediction uncertainty and representativeness. In \citet{tali18}, an objective function balancing GP-based uncertainty estimation and Kernel Principal Component Analysis (KPCA)-based representativeness was introduced, ensuring that the most informative sample was labelled from a given batch. Alternatively, \citet{shxi18} employed approximate linear dependence (ALD) to measure the relationship between unlabelled samples and the existing labelled dataset, providing another means of assessing sample representativeness beyond PCA or PLS.  Recent advancements in adversarial learning-based AL strategies offer a new perspective for soft sensor optimisation. In \citet{dazh22, daya23}, a discriminator network was trained to distinguish between labelled and unlabelled instances. The network’s uncertainty scores served as an indicator of representativeness—if an instance was poorly represented in the labelled set, it was prioritised for labelling. This adversarial approach enables adaptive sample selection, ensuring that the labelled dataset evolves to better cover the feature space with minimal data acquisition.  

\textit{Remark:} Given the increasing adoption of Raman spectroscopy for real-time bioprocess monitoring, integrating AL into Raman-based soft sensor development presents a promising research direction. By systematically reducing labelling costs and enhancing model robustness, AL-driven soft sensors could effectively mitigate small data challenges in upstream bioprocessing, particularly under budget and resource constraints.

    \subsection{Data Augmentation}
        Data Augmentation (DA) is a set of ML methods that increase the diversity and size of training datasets without collecting new data. DA could be crucial in developing effective ML models when data is scarce. DA techniques can be classified further into \textbf{Data Transformation} and \textbf{Data Synthesis} \citep{mumu25}. Both methods can be used to generate new data instances, and the key distinction between them is that the data generated from Data Transformation is mapped from the original data instances via transformations. In contrast, Data Synthesis generates new data instances directly by mixing existing data instances or sampling a learned data distribution. 
        
        Most DA techniques are domain-specific, with different transformations applied across domains and data types \citep{wawa24}. Thus, given that the application focus of this paper is in the domain of bioprocess, where data are typically collected in the form of tabular and time-series data, we will discuss DA techniques for tabular and time-series data only. In 2021, \citet{iwuc21} published a comprehensive survey focusing on DA techniques for tabular data and time series. We have summarised the key idea of each of the methods from the survey in the following subsections.

        \subsubsection{Data Transformation}
            
            For a given tabular data with $\mathbf{n}$ data points, the objective of data transformation is to apply transformation operations $\mathbf{g}$ to the original samples $\mathbf{X}$, generating additional training data $\mathbf{X'}$ without modifying the associated label set $\mathbf{Y}$, which means that the transformation $\mathbf{g(X_i)}$ yields a modified sample associated with the original label $\mathbf{Y_i}$, e.g., $\mathbf{g(X_i, Y_i) \rightarrow (X_i', Y_i)}$, where $\mathbf{i=1 \dots n}$, and each $\mathbf{x_i}$ could be either univariate or multivariate. The objective of transforming time series data is the same, except where $\mathbf{X=\{x_1, \dots, x_t, \dots,x_T\}}$, with $\mathbf{T}$ being the number of time steps from the training set. Table \ref{data_transformation_method} summarises the common data transformations for tabular and time-series data.

            The main assumption required for data transformation is that the transformed data instances are typical of the dataset. If data transformations are done wrongly, it can introduce noise to the training dataset and negatively impact the model performance \citep{iwuc21}. Multiple transformation operations can also be performed to further increase the transformed data's diversity. However, due to constraints on computational resources, cost, and potential negative effects of spurious data, selecting and performing only the most beneficial transformation is generally recommended \citep{iwuc21}. 

        \begin{ThreePartTable}
          \begin{ltabulary}{P{5cm}P{2cm}P{8.25cm}}
          \caption{Common Data Transformation Techniques for Tabular and Time Series Data.\label{data_transformation_method}}\\
            \toprule
            \textbf{Method} & \textbf{Applicable Data Type} & \textbf{Key Idea}\\
            \midrule
            Jittering \citep{fihs19, ralo19, umpf17, argu19} & Tabular \& Time Series & $\mathbf{x' = \{x_1 + \epsilon_1, \dots, x_T + \epsilon_T\}}$ where $\epsilon \sim N(0,\sigma^2)$. Assume the unseen data are a factor of Gaussian noise away from the training data. $\sigma$ is a hyperparameter. \\
            \midrule
            Rotation \citep{ohal17, fafo18, iwuc20, umpf17, ralo19} & Time Series & $\mathbf{x' = \{Rx_1, \dots, Rx_T\}}$ where $\mathbf{R}$ is the rotation matrix for angle $\theta \sim N(0,\sigma^2)$. $\sigma$ is a hyperparameter. \\
            \midrule
            Scaling \citep{trch20, deca19, ralo19, umpf17}& Tabular \& Time Series & $\mathbf{x' = \{\alpha x_1, \dots, \alpha x_T\}}$ where $\alpha \sim N(1,\sigma^2)$. $\sigma$ is a hyperparameter. Alternatively, $\alpha$ could be randomly selected from a predefined set of values.\\
            \midrule
            Magnitude Warping \citep{umpf17} & Time Series & $\mathbf{x' = \{\alpha_1 x_1, \dots, \alpha_T x_T}\}$, where $\mathbf{ \alpha_1, \dots, \alpha_T}$ is created by interpolating a cubic spline with knots
            $u = \{u_1, \dots, u_i, \dots, u_I\}$ with $u_i \sim N(1,\sigma^2)$. $\sigma$ is a hyperparameter.\\
            \midrule
            Window Slicing \citep{lema16} & Time Series & $\mathbf{x' = \{x_{\phi}, \dots, x_{W+\phi}\}}$, where $\phi$ is a random positive integer $\leq T - W$, and $W$ is the window size.\\
            \midrule
            Permutation \citep{odha18,umpf17, Pan2020} & Time Series & Generate $\mathbf{x'}$ by re-arranging segments of a time series without preserving time dependencies. The segments could be either equal or vary in size. When all segments have the size of 1, it is equivalent to random shuffling.\\
            \midrule
            Time Warping \citep{umpf17} & Time Series & Generate $\mathbf{x'}$ by first warps the original time domain $1, \dots ,T$ to new values $\tau(1), \dots, \tau(T)$ using a warping function $\mathbf{\tau}$. Then, the time series with the warped time domain $\mathbf{Z}$ is created and assigned with the same magnitude as $\mathbf{X}$ s.t. $\mathbf{x_{t}=z_{\tau(t)}}$ for all $t=1, \dots, T$. Finally, $x' = \{x_1, \dots, x_T\}$ is generated as the interpolated values for $t=1, \dots ,T$ based on time series $\mathbf{Z}$.\\
            \midrule
            Window Warping \citep{lema16} & Time Series & Generate $\mathbf{x'}$ by taking a random time series window and stretching or contracting it by a multiplier. $\mathbf{x'}$ is generated as the interpolated values for $t=1, \dots, T$ based on the stretched or contracted time series.\\
            \bottomrule
          \end{ltabulary}
        \end{ThreePartTable}
        
    \subsubsection{Data Synthesis}

            Data Synthesis generates synthetic data using algorithms or ML models designed to capture the original data's patterns, characteristics, and behaviours. \textbf{Pattern Mixing} and \textbf{Generative Models} are the two main types of methods for generating new data instances. Pattern Mixing generates new data instances by combining existing data instances. For time series data, many pattern-mixing techniques leverage Dynamic Time Warping (DTW) \citep{sach78} to generate synthetic variations. DTW allows non-linear alignments by dynamically warping the time axis to find the optimal match between two sequences, which enables pattern-mixing techniques to identify the key warping patterns in existing sequences and apply similar transformations to new or augmented samples. The common pattern mixing techniques are summarised in Table \ref{pattern_mixing}. On the other hand, generative models are ML models that aim to learn the true data distribution of the training set and generate realistic new data points with some variations. Formally, given training set $X = \{x^{(1)}, x^{(2)}, ..., x^{(n)}\}$ drawn i.i.d. from the real data distribution $p_{\text{data}}(x)$, a generative model tries to estimate the parameters $\theta$ of the model distribution $p_{\text{model}}(x; \theta)$ such that the model distribution is as close as possible to the real data distribution. The closeness of the model distribution to the real data distribution can be measured using various statistical distances or divergences. One common choice is the Kullback-Leibler (KL) divergence:
            
            \begin{equation}
            D_{KL}(p_{\text{data}}(x) || p_{\text{model}}(x; \theta)) = \mathbb{E}_{x \sim p_{\text{data}}(x)}\left[\log\left(\frac{p_{\text{data}}(x)}{p_{\text{model}}(x; \theta)}\right)\right]
            \end{equation}
            
            The goal then is to find the parameters $\theta$ that minimise the KL divergence:
            
            \begin{equation}
            \hat{\theta} = \arg\min_{\theta} D_{KL}(p_{\text{data}}(x) || p_{\text{model}}(x; \theta))
            \end{equation}

            \begin{table}[H]
            \vspace{2ex}
            \caption{Common Pattern Mixing Techniques for Tabular and Time Series Data.}\label{pattern_mixing}
            \centering
            \begin{tabulary}{0.9\textwidth}{P{2cm}P{2cm}P{11cm}}
            \toprule
            \textbf{Method} & \textbf{Applicable Data Type} & \textbf{Key Idea}\\
            \midrule
            SMOTE \citep{chbo02} & Tabular \& Time Series & In Synthetic Minority Over-sampling Technique (SMOTE), a random instance $x$ is chosen from a minority class, and another random instance $x_{NN}$ is selected from the k-nearest neighbours of $x$. Then synthetic data instance is generated as $x' = x +\lambda|x-x_{NN}|$, where $\lambda$ is a random value between 0 and 1. \\
            \midrule
            Guided Warping \citep{argu19} & Time Series & Guided warping leverages DTW that incorporates additional constraints or guidance to warp the time axis of existing time series and thus generate new time series. The main idea behind guided warping is to utilise prior knowledge to guide the alignment and ensure a meaningful warping\\
            \midrule
            SPAWNER \citep{kaka20} & Time Series & SuboPtimAl Warped time series geNEratoR (SPAWNER) 
            Leverages DTW's warping capability while introducing an additional constraint that directs the warping path through a random point and generates new time series by averaging aligned patterns.\\
            \midrule
            DBA \citep{peke11} & Time Series & DTW Barycentric Averaging (DBA) generates new time series as the barycenter of time elements aligned by DTW from multiple discrete time series.\\
            \bottomrule
            \end{tabulary}
            \vspace{-3ex}
            \end{table}
            \raggedbottom
 
             Generative models can be further broken down into \textbf{statistical} generative models, \textbf{neural network-based} generative models, and \textbf{first-principle models}. Statistics-based generative models refer to a class of models that use statistical principles to model the underlying distribution of the data. Examples include Gaussian Mixture Models (GMMs), Monte Carlo, Markov Chain Monte Carlo, and Bayesian Network \citep{muah23}. Neural-network-based generative models are a class of generative models that use neural networks to model data distribution. These models aim to learn the true data distribution of the training set to generate new data points. Two popular types of neural-network-based generative models are Generative Adversarial Networks (GANs) and Autoencoder \citep{muah23}:
            
            \begin{itemize}
            \item \textbf{Generative Adversarial Networks (GANs):} A GAN consists of two neural networks, the generator $G$ and the discriminator $D$. The generator generates synthetic data, and the discriminator distinguishes between real and synthetic data. The objective is to train the generator to generate data such that the discriminator cannot distinguish it from real data.
            
            \item \textbf{Encoder-Decoder Network:} The Encoder-Decoder architecture is a type of neural network design pattern composed of two main components: the encoder and the decoder. The encoder processes the input data and generates a compressed representation, which the decoder takes as input and generates the output. Some Encoder-decoder Networks, such as Variation Auto-Encoder (VAE) \citep{kiwe22}, can be trained to generate synthetic data. VAE learns to encode data into a lower-dimensional latent space and then decode it back to the original input. Once the model is trained, the encoder could be detached from the network, and then we could feed random noise (or sampling from a distribution) as the input to the decoder to generate synthetic data. 
            \end{itemize}
           
             Unlike statistical and neural network-based generative models, which learn the data distribution based on existing data, first-principle or mechanistic models generate new data based on the underlying physical, chemical, and biological principles governing a system. Thus, it can be used to create new data even without any existing data. In the context of bioprocessing, these models are derived from the fundamental understanding of the process and its components, such as cell metabolism, bioreactor dynamics, and mass transfer phenomena \citep{nalu22}. Another type of data synthesis method that can be used for time series is \textbf{Decomposition}, which decomposes the time series into components, then perturbs and recombines these components together to generate new time series \citep{nabu20, el02, behy16, gaso21}. Common decomposition method includes Empirical Mode Decomposition (EMD) \citep{hush98}, Independent Component Analysis (ICA) \citep{co94}, Seasonal and Trend decomposition using Loess (STL) \citep{clcl90}, and RobustSTL \citep{wega18}.

                
        \subsubsection{Applications of Data Augmentation in Upstream Bioprocessing}
        Data augmentation techniques have been widely adopted in upstream bioprocessing applications to address challenges posed by limited experimental data. These techniques can be broadly classified into synthetic data generation using \textit{first-principle models}, \textit{statistical simulation} methods, \textit{data transformation} techniques, and \textit{noise-based augmentation} strategies. By leveraging these methods, researchers have successfully enhanced predictive modelling accuracy, improved machine learning generalisation, and optimised bioprocess performance.
        
        A significant number of studies employ first-principle models to generate synthetic datasets that supplement limited real-world experimental data. For example, \citet{bech06} and \citet{roro08} utilised the Anaerobic Digestion Model No. 1 (ADM1) to create synthetic data for dimensionality reduction analysis using Principal Component Analysis (PCA). Their findings showed that a reduced number of biomasses and reactions could still capture key system dynamics, demonstrating that synthetic data can facilitate model simplification. Similarly, \citet{abna07} used the Sonnleitner-Kappeli model to generate synthetic data, which was then used to train Self-Organising Maps (SOM) for clustering metabolic states and forming a Multiple Local Linear Models (MLLM) framework for fed-batch yeast fermentation. In microbial factory assessment, \citet{oyli19} employed genome-scale metabolic models (iML1515) to simulate metabolic fluxes under bioprocess constraints, thereby augmenting training datasets for machine learning models used to predict titer, rate, and yield. Additionally, \citet{boba22} leveraged mechanistic digital twins to simulate fed-batch monoclonal antibody (mAb) production, enabling enhanced process optimisation despite limited experimental datasets.
        
        Another prevalent data augmentation approach is a statistical simulation and stochastic modelling, particularly Monte Carlo-based methods. \citet{stpa11} integrated Monte Carlo simulations with multivariate statistical analysis to model batch-to-batch variability in biopharmaceutical manufacturing, aiding facility fit assessments. Similarly, \citet{yafa14} employed stochastic discrete-event simulations to generate synthetic data representing batch-to-batch fluctuations in key bioprocess parameters, which were then used in conjunction with decision tree classification (CART) for facility debottlenecking analysis. In vaccine production cost modelling, \citet{chwi14} applied Monte Carlo stochastic simulations to create probability distributions of production costs, allowing researchers to assess the economic feasibility and optimise decision-making. These studies demonstrate that Monte Carlo-based augmentation techniques effectively mitigate uncertainty and enhance predictive modelling in bioprocesses.
        
        Several studies have utilised data transformation techniques to augment existing datasets, expanding the amount of available training data for machine learning applications. \citet{rosa22} applied polynomial smoothing and interpolation-based augmentation to increase the number of training samples for syngas fermentation modelling, significantly improving the predictive accuracy of machine learning models. In Raman spectroscopy-based bioprocess monitoring, \citet{lath24} proposed an innovative spectral augmentation approach that generates synthetic spectra with statistically independent labels, reducing biases caused by inherent correlations among bioprocess variables and improving CNN-based predictions.
        
        Lastly, noise-based data transformation strategies have been used to enhance the robustness of machine learning models trained on small datasets. \citet{dema16} introduced artificial noise augmentation by adding 3\% normally distributed random error to experimental data, increasing training diversity and improving artificial neural network (ANN) predictions for cyanobacterial C-phycocyanin production. Similarly, \citet{pime22} employed stochastic regularisation techniques such as dropout and minibatch sampling within a deep hybrid modelling framework, preventing overfitting and enhancing predictive performance in bioprocess simulations. These noise-based augmentation methods ensure that ML models generalise well to new data, even when experimental datasets are limited.
        
        Collectively, these studies have demonstrated the critical role of data augmentation in overcoming data scarcity challenges in upstream bioprocessing applications. By employing first-principle models, statistical simulations, and data transformation methods, researchers have successfully improved process modelling, optimised decision-making, and enhanced machine learning model performance in bioprocess development.

        Despite the growing adoption of data augmentation in upstream bioprocess applications, the field still lags behind other industries in utilising more advanced data augmentation techniques. This limitation is particularly evident in the development of soft sensors for bioprocess monitoring, where current approaches rely primarily on first-principle simulations, Monte Carlo-based statistical augmentation, and basic noise-based transformations. In contrast, other process industries have begun leveraging more sophisticated data augmentation strategies, such as GANs and VAEs, to enhance soft sensor performance under data-scarce conditions.
        
        For instance, in chemical process modelling, \citet{chji22} demonstrated the use of GANs to generate virtual data distributions that closely match the existing labelled dataset. A subset of the generated data was then selected using generalised local probability, ensuring that the augmented training set maintained a distribution consistent with the original data. Training soft sensors with this augmented dataset significantly improved their performance, highlighting the potential of GAN-based synthetic data generation for bioprocess monitoring applications. Similarly, \citet{zhho21} and \citet{zhxu22} proposed a more structured augmentation approach for soft sensor development in a cascade reaction process for high-density polyethylene production. Instead of randomly generating synthetic data, the method first identified scarce regions in the dataset using the K-Nearest Neighbours (KNN) algorithm. Then, a Wasserstein GAN (WGAN) with gradient penalty was employed to generate unlabeled samples filling these scarce regions. A conditional GAN with a cycle structure was used to assign meaningful labels to the synthetic data, ensuring that the augmented samples retained physical relevance. This approach significantly improved the soft sensor accuracy compared to conventional data synthesis techniques such as Mega-Trend Diffusion and SMOTE, demonstrating its effectiveness in enhancing data-limited models.
        
        Another promising direction for data augmentation in soft sensor development is the integration of causality-informed generative models. In \citet{jige22}, a novel causality-informed VAE was introduced to generate samples for a soft sensor predicting oxygen ($O_2$) concentration in a primary reformer. Unlike conventional VAEs, this method incorporated mechanistic knowledge into the sample generation process, ensuring that the synthetic data preserved the underlying causal relationships governing the chemical process. The resulting augmented dataset significantly improved the robustness and generalisation ability of the soft sensor.
        
        \textit{Remark:} These advanced data augmentation methods could greatly benefit soft sensor development in upstream bioprocessing applications. By leveraging deep generative models to create realistic, process-informed synthetic data, researchers could overcome the constraints of small datasets and improve bioprocess monitoring, control, and optimisation. However, the application of these techniques in upstream bioprocessing remains largely unexplored, and further research is required to assess their effectiveness in real-world bioprocess environments. Future studies should investigate how GAN- and VAE-based virtual data generation can be adapted to upstream bioprocess monitoring.        
        
    \subsection{Dimensionality Reduction}
       \textbf{Dimensionality reduction} (DR) is a technique used to reduce the number of features in a dataset while preserving critical information for downstream tasks. This is particularly valuable when working with small datasets, where a high-dimensional feature space can lead to issues such as overfitting, increased model complexity, and poor generalisation. By reducing dimensionality, DR helps mitigate these challenges by improving model stability, reducing computational costs, and enhancing learning efficiency \citep{saaz21, boda24, ghsa19}. The two primary approaches to dimensionality reduction are \textbf{Feature Selection} and \textbf{Feature Extraction} \citep{boda24, ghsa19}.
            
        
        \subsubsection{Feature Extraction}

        Feature extraction reduces the data dimensionality while preserving meaningful structure by transforming the original feature set into a smaller set of new features that retain most of the original data's variability. This method is particularly useful in small data settings because it compresses information into a lower-dimensional space, allowing the model to learn more effectively from limited data \citep{boda24, ghsa19}. We have broadly classified the feature extraction techniques shown below based on the recent reviews \citep{rare21, boda24, ghsa19}. Furthermore, we also analysed the suitability of each approach in data-scarce scenarios, as the effectiveness of these methods can vary significantly when dealing with small datasets.

        \begin{itemize}
        \item \textbf{PCA Based} - PCA-based feature extraction methods are variants of PCA \citep{pe01}, which transform features into uncorrelated variables called principal components (PCs), which represent the directions of maximum variance in the dataset. Variants such as Sparse PCA, Kernel PCA, and Incremental PCA extend their applicability to feature selection, nonlinear data, and streaming data. PCA is generally a data- and compute-efficient method \citep{algr22,foal19}, as it relies on computing the covariance matrix rather than requiring extensive training. However, in cases where the dataset is highly sparse or dimensionality exceeds the number of available samples, PCA may suffer from unstable covariance estimation and overfitting. Regularisation techniques such as Regularised PCA or Sparse PCA can help mitigate this issue and improve robustness in small data scenarios. There are also other robust PCA approaches utilising resampling techniques and constructing PCA ensembles \cite{gaba06} which are particularly suitable for small data environments.
        \item \textbf{Linear Discriminant Analysis (LDA) Based} - LDA aims to optimise class separability by maximising between-class variance while minimising within-class variance. However, LDA performs poorly in small data scenarios, particularly when the number of samples per class is limited or class distributions deviate from Gaussian assumptions. The estimation of class covariance matrices becomes unreliable, and overfitting may occur when the number of features exceeds the number of samples. To address these limitations, Local Fisher Discriminant Analysis (LFDA) \citep{su06} introduces local structure preservation, improving performance when dealing with small datasets. Additionally, regularised LDA techniques can stabilise covariance estimation and enhance generalisation.
        \item \textbf{Neural Network Based} - Neural network-based dimensionality reduction methods, such as Autoencoders and Self-Organising Maps (SOMS), have gained popularity due to their ability to learn nonlinear representations. However, these models typically require large datasets to generalise effectively, and they are considered less data and compute-efficient than PCA-based methods \citep{algr22,foal19}. Thus, in small data scenarios, neural networks are prone to overfitting, leading to poor generalisation. Regularisation techniques such as dropout and weight decay can be employed to counteract this. Alternatively, transfer learning with pre-trained models can help mitigate the need for extensive data by leveraging features learned from larger datasets. Nevertheless, simpler methods, such as shallow autoencoders or SOMs with a limited number of neurons, may be preferable in small data contexts.
        \item \textbf{Canonical Correlation Analysis Based} - CCA \citep{ho36} is a statistical method that finds maximally correlated linear projections between two multivariate datasets. It is particularly useful for understanding the shared structure between different feature sets. However, in small data settings, CCA may suffer from unstable correlation estimation, leading to poor performance. Regularised CCA (RCCA) introduces constraints to stabilise correlation estimation and improve robustness when sample sizes are limited. CCA-based approaches work best when a strong relationship exists between the two datasets and when there are sufficient samples to reliably estimate correlations.
        \item \textbf{Partial Least Squares Regression Based} - PLSR \citep{wo75} is closely related to CCA but focuses on maximising covariance rather than correlation between predictor variables ($X$) and response variables ($Y$). Unlike standard regression, PLSR extracts latent components that best explain the variation in $Y$, making it well-suited for predictive modelling in high-dimensional, low-sample-size settings. PLSR is particularly advantageous in small data scenarios because it does not require estimating full covariance matrices, reducing the risk of overfitting. Additionally, sparse and regularised PLSR variants further enhance stability by enforcing sparsity on the latent variables. Compared to PCA, which only captures variance in the feature space, PLSR ensures that dimensionality reduction is optimised for predictive performance. 
        \item \textbf{Non-negative Matrix Factorisation (NMF) Based} - Non-negative Matrix Factorisation (NMF) \citep{lese99} decomposes a non-negative matrix into lower-dimensional factors while maintaining interpretability. Unlike PCA, which captures global variance, NMF focuses on extracting meaningful underlying components. This approach is particularly advantageous in applications such as image processing and text mining, where data is naturally non-negative. NMF can perform well with small datasets if the data is well-structured, though factorisation instability may arise when sample sizes are too small. In such cases, sparse NMF or regularised NMF techniques can be applied to enhance stability and prevent overfitting.
        \item \textbf{Manifold Based} - Manifold learning methods, including Isomap \citep{tesi00}, Locally Linear Embedding (LLE) \citep{rosa00}, and Laplacian Eigenmaps \citep{beni01}, aim to represent data in a lower-dimensional space while preserving geometric and topological properties. These methods generally require a sufficiently large dataset to construct a reliable neighbourhood graph, which forms the basis for dimensionality reduction. If data is sparse, the estimated manifold structure may be distorted, leading to poor generalisation. As a result, manifold learning is typically not well-suited for small data scenarios. In cases where labelled data is available, semi-supervised manifold learning approaches can improve performance by leveraging available class information to guide the transformation.
        \end{itemize}
        
        Overall, PCA, NMF, and PLSR are generally more stable in small data settings, particularly when regularised versions are employed. LDA and neural network-based approaches tend to struggle due to unstable covariance estimation and overfitting, respectively. CCA can be useful if properly regularised, while PLSR is particularly effective in small datasets where predictive modelling is required. Manifold learning methods typically require more data to construct meaningful representations. By understanding these limitations, researchers can select the most appropriate dimensionality reduction method based on the size and structure of their datasets.
    
        \subsubsection{Feature Selection}
            
        Feature selection is the process of selecting a subset of features in a given dataset for training and optimising the performance of ML models. For small datasets, eliminating redundant or irrelevant features helps focus the model on the most meaningful data points, improving generalisation and reducing the risk of overfitting \citep{boda24, ghsa19}. The major branches of feature selection methods include:
        
        \begin{enumerate}
        \item \textbf{Filter Methods}: These methods apply a statistical measure to assign a score to each feature, which represents the importance of this feature and is used to determine whether a feature should be kept or removed from the dataset. Examples of such methods include ReliefF \citep{wazh16}, Minimum Redundancy Maximum Relevance (MRMR) \citep{naja19, zhzh19, wati18, wayu18, chya17}.
        
        \item \textbf{Wrapper Methods}: In these methods, different combinations of features are prepared, evaluated, and compared to other combinations. A predictive model is used to evaluate the combination of features and assign a score based on model accuracy. Examples include recursive feature elimination \citep{gyre05}, forward feature selection, backward feature elimination, and exhaustive feature selection. Some methods also use heuristics to determine the feature set, including Genetic Algorithms (GAs), Particle Swarm Optimisation (PSO), and their hybridisation \citep{alka07}. Another heuristic that can be efficiently computed in high-dimensional data is Simultaneous Perturbation Stochastic Approximation (SPSA) \citep{yead18}. 
        
        \item \textbf{Embedded Methods}: These methods learn which features best contribute to the model's accuracy while the model is being created. The most common type of embedded method is the LASSO (Least Absolute Shrinkage and Selection Operator) \cite{ro96} regularisation method, where the penalty applied over the size of the coefficients can drive some of them to zero, effectively selecting the features.
        
        \end{enumerate}
                
       Feature extraction methods may retain more information than feature selection since they combine existing features rather than discard them. However, most feature extraction methods only preserve certain structures of the feature, meaning there is information loss. Thus, depending on the implementation details, feature selection may retain more information than feature extraction.
       
       Another advantage of feature selection is that it offers better interpretability, as extracted features may be difficult to understand due to their composite nature. By leveraging dimensionality reduction techniques, small datasets can be transformed into more compact and informative representations, improving model performance while addressing the challenges associated with limited data availability.
       
        \subsubsection{Applications of Dimensionality Reduction for Small Bioprocess Data}
    
        A significant portion of the reviewed studies employ feature extraction techniques, where high-dimensional input data is transformed into a new, lower-dimensional space while retaining essential variance information. Among these techniques, PCA and its extensions (Multi-way PCA (MPCA), Principal Component Regression (PCR), and Partial Least Squares (PLS)) are the most widely used. Studies such as \citet{iggl97}, \citet{kuch04}, and \citet{zhwa21} applied PCA to reduce the dimensionality of fermentation and glycosylation process data, allowing for better fault detection, process monitoring, and forecasting. MPCA and PLS-based approaches were frequently used to enhance predictive models for key process variables, particularly in metabolic modelling \citep{babe22}, time-series bioprocess analysis \citep{alsu22}, and hybrid ML-bioprocess models \citep{rapa23}. These methods help mitigate overfitting, computational complexity, and multicollinearity, making ML-based soft sensors more reliable despite limited experimental data.
        
        Moreover, advanced feature extraction methods such as UMAP are emerging in bioprocess analytics. \citet{oddw22} integrated UMAP with PCA to analyse high-dimensional multi-omics datasets, demonstrating that non-linear feature extraction methods can further improve ML model generalisation in bioprocess optimisation. Additionally, studies such as \citet{kuma15} introduced novel Compromise Whitening (CWH), which balances PCA and Data Whitening (DWH) techniques to extract robust latent features for clustering-based bioprocess classification. These findings suggest PCA remains the dominant approach as it is highly data- and computationally efficient. However, non-linear and customised feature extraction techniques are gradually being explored to improve bioprocess monitoring. It is also worth noting that more complex dimensionality reduction techniques, such as neural network-based methods, are rarely applied to upstream bioprocess applications due to limited available data. 
        
        Unlike feature extraction, feature selection techniques are less frequently applied in upstream bioprocessing applications. The reviewed studies indicate that LASSO regression \citep{fala13} and Random Forest (RF) feature selection \citep{huzh23} are among the few feature selection methods utilised in bioprocess applications. In \citet{fala13}, LASSO was used to remove redundant operational parameters in bioprocess scale-up modelling, allowing for improved generalisation in cytotoxic compound production predictions. Similarly, \citet{huzh23} applied RF-based feature selection to identify relevant auxiliary variables for training an LSTM-based soft sensor in penicillin fermentation, preventing overfitting and enhancing predictive accuracy.
        
        The limited adoption of feature selection techniques in bioprocessing contrasts sharply with other industrial applications, where automated feature selection (e.g., Recursive Feature Elimination, Mutual Information, and Bayesian Optimisation-based selection) is widely used to optimise soft sensors \cite{kaga09,kaga09a}. The heavy reliance on PCA-based methods in bioprocessing may be attributed to the need to capture latent biological and process interactions, at which PCA excels. However, since feature selection methods are less prone to information loss and retain original process parameters, they present a significant opportunity for improving ML models trained on extremely small bioprocess datasets. For example, feature selection can be applied based on prior knowledge to remove certain regions of the spectral data, such as Cosmic Ray Removal and Dark Spectrum Subtraction \citep{gava15}.
        
        
        \textit{Remark:} Future research should explore hybrid approaches combining feature extraction and selection methods to improve model interpretability and robustness. For example, integrating PCA with LASSO, RF, or knowledge-based feature selection could ensure that the most relevant extracted features are retained while discarding non-informative ones, reducing model complexity while maintaining predictive performance. Additionally, applying non-linear feature extraction techniques such as Kernel PCA may provide better latent representations than standard linear PCA.

\section{ML Methods for Addressing Small Data Challenges during Model Development}
\label{sec:model-training}
    During the model development phase, the collected, augmented, and feature-engineered dataset will be used to train, optimise, and evaluate a set of candidate models. The candidate model pool could be arbitrarily constructed based on experience, common industrial practice, or previous benchmark results. It could also be explored as part of the fully automated machine learning (AutoML) execution, as is becoming more and more common \cite{kemu24,khke23,scke23} and including applications in the broader process industry \cite{kaga09d,sabu16,sabu16a,sabu19}. The best-performing candidate model will be deployed if it meets the ML task's performance requirement. However, even the performance of the best model may still be unsatisfactory if trained on small data \cite{kaga09b,kaga10a}. Thus, we must apply other ML methods to augment the model's performance with additional learning sources. In the following subsections, we will introduce a set of ML methods that allow the model to learn from additional sources other than the labelled data.
    
    Firstly, Semi-supervised learning (SSL) will be introduced, which tackles the small data challenge by learning from unlabelled data, if available \cite{gape04}. Then, we will discuss two well-known and widely adapted approaches in the bioprocess domain and broader process industry - Bayesian Methods and Data- \& Knowledge-driven Hybrid (DKH) Modelling, which tackle this problem by combining prior domain knowledge with data-driven ML models \cite{kaga09}. Lastly, Meta \& Transfer Learning (Meta-L \& TL) are introduced. They attack the small data problem by leveraging knowledge learned from other potentially related datasets and tasks \cite{lebu15,albu20}. Although both Meta-L and TL are relatively new for bioprocess applications and are not widely adopted by the industry and bioprocess research community, they have shown promising results in many other domains. 
    
        
    \subsection{Semi-Supervised Learning}

        Semi-supervised learning (SSL) is a machine learning paradigm that uses labelled and unlabelled data during training. It is particularly useful when labelled data is scarce or expensive to obtain, while unlabelled data remains abundant. A comprehensive review of SSL was presented in \citet{vaho20}, which discussed the following four main branches of SSL:
        
        \begin{itemize}
        \item \textbf{Wrapper Methods}: This method first trains models with labelled data. The trained models are then used to label the unlabelled data. The most confident predictions for the unlabelled data are then added to the labelled dataset with their predicted labels, and the models are retrained on this larger, augmented labelled dataset. The wrapper methods include self-training, co-training, and tri-training, with one, two, and three ML models trained and used for label estimation.
         \item \textbf{Unsupervised Preprocessing}: These methods either derive additional features from unlabelled data, cluster the data, or set starting parameters for a supervised learning model in an unsupervised way. Like wrapper methods, they can be applied along with supervised learning models. However, unlike the wrapper method, it does not generate additional labelled data for training supervised models.
        \item \textbf{Intrinsically Semi-supervised}: Intrinsically semi-supervised methods, in the context of SSL, are algorithms inherently designed to use both labelled and unlabelled data during the training process. Unlike methods that apply unsupervised techniques as a pre-processing step before supervised learning or methods that combine supervised and unsupervised steps, intrinsically semi-supervised methods simultaneously use both data types within one unified learning framework. These methods are created to naturally handle the combination of labelled and unlabelled data, effectively using the information from the unlabelled data to improve generalisation performance. The central premise of these methods is that the unlabelled data, when used alongside the labelled data, can provide useful information that helps to improve the learning algorithm's overall performance.

        \item \textbf{Graph-based methods}: These methods construct a graph where each node represents a data point, and edges connect similar data points. The labelled data points propagate their labels to nearby unlabelled points via the edges. There are many techniques for this propagation, including Label Propagation, Label Spreading, and graph-based regularisation methods like Laplacian Regularisation.
        \end{itemize}

        The usefulness of unlabelled data depends on the distribution of both feature and label space.  In general, at least one of the following assumptions needs to be satisfied for unlabelled data to be useful \citep{chsc10}:

        \begin{itemize}
            \item{\textbf{Continuity / Smoothness Assumption}: Data satisfies this assumption if data points that are closer to each other in the input space are more likely to have the same or similar label.}
            \item{\textbf{Cluster Assumption}: This assumption states that the data naturally form discrete clusters and that the data points in the same cluster are more likely to have the same or similar label. Since data points from the same cluster imply they are close in the feature space, the cluster assumption implies the continuity assumption.}
            \item{\textbf{Manifold Assumption}: This assumption states that the data lie approximately on a manifold with a much lower dimension than the input space. Thus, the manifold can be learned from labelled and unlabelled data's distances and densities relationship (defined on the manifold). The manifold assumption also implies the continuity assumption, as data close to each other on the manifold should also have the same or similar labels.}
        \end{itemize}

        Most of these assumptions can be assessed using domain knowledge and data analysis methods such as clustering and dimension reduction techniques. In addition, the characteristics of the data also need to be considered. For example, when dealing with multivariate time series data, rather than evaluating the distance between two data points purely based on their feature spaces, the time element must also be accounted for when checking these assumptions.    
        
        \subsubsection{Applications of Semi-Supervised Learning for Small Bioprocess Data}

        
        A paper published in 2012 \citep{jiwa12} proposed a semi-supervised learning model named recursive weighted kernel regression (RWKR) to be used as the soft sensor for monitoring biomass and penicillin concentration. The experiments use simulated data with and without 3\% noise added. In this simulated dataset, the sampling interval is every 0.5h. However, the sample label is only available every eight hours, leaving the other 15 samples unlabelled to mimic the infrequent offline analysis in practice. Three benchmark models are trained using the simulated data, and they are Relevance Vector Machine (RVM), Harmonic Function (HF), and Harmonic Function without unlabelled data (HF0). The result shows that the RWKR is superior to all three models when noise is added to the data. Furthermore, the HF0 model consistently performs the worst with or without noise added to the simulated data, highlighting the benefit of leveraging unlabelled data. In \citet{qiwa22}, a different SSL approach was examined with a similar data set-up - simulated data with Gaussian noise but with different labels to unlabelled data ratios. The authors proposed a localised semi-supervised algorithm that estimates the labels of unlabelled data points based on comprehensive similarity. These unlabelled data with estimated labels are then combined with the labelled data to form the new training data for building RVM to predict penicillin concentration. Furthermore, a Sequence-constrained fuzzy c-means algorithm is applied to break the bioprocess data into three phases, and three different RVMs were built using data from each phase. The results show that the proposed approach is superior to other benchmark methods that could not leverage the unlabelled data or take the multi-phase approach, highlighting the effectiveness of the proposed SSL algorithm. In \citet{goti18}, a deep learning-based SSL approach was proposed for building soft sensors that estimate crucial parameters in Streptokinase and Penicillin fermentation processes. The proposed approach applied Self-Organising Map (SOM) onto unlabeled data to initialise the weights of the feed-forward neural network that is used to estimate the process parameters. With weights initialised by SOM instead of random initialisation, the feed-forward neural network can learn faster and generalise better using limited labelled data. It has shown significantly better performance compared to the conventional SVR model. \citet{esta22} explores semi-supervised regression (SSR) as a tool for developing soft sensors in (bio-)chemical process monitoring, particularly in cases where quality measurements are infrequent and expensive to obtain. The authors apply SSR to two case studies—a Williams-Otto process and a bioethanol production process—and compare them against traditional regression models. By leveraging unlabeled process data alongside a limited set of labelled quality measurements, the semi-supervised approach improves prediction accuracy while requiring fewer labelled samples. The study demonstrated that semi-supervised deep kernel learning (SSDKL), which integrates neural networks with GP, outperforms traditional models in reducing root mean square error (RMSE). The method enhances predictive robustness, particularly in low-data scenarios, making it valuable for real-time process optimisation and control in upstream bioprocess applications. However, the results indicated that while SSDKL effectively captures steady-state behaviours, additional methods or more frequent measurements may be needed to fully model process dynamics. The paper suggested further research on integrating SSR with dynamic models to enhance its applicability in real-world bioprocessing environments.
        
        Outside of bioprocess applications, plenty of research has been done on applying SSL to develop soft sensors. In \citet{fezh20}, a novel divergence-based wrapper SSL method called adversarial tri-regression was introduced for soft sensor applications in cigarette production. It generates adversarial samples based on the consideration of maximum disturbance, which, combined with original labelled samples, are then used to train three initial regressors. Each regressor labels an unlabelled sample when the other two agree. As more samples are mutually labelled, the final model averages the prediction of the three base regressors and yields increasingly accurate predictions. On the other hand, an intrinsic SSL approach for PCR with Bayesian regularisation named Semisupervised Bayesian Probabilistic PCR (SBPPCR) was proposed and developed in \citet{geso11}. The paper proposed a new maximum likelihood function that includes the likelihood of unlabelled data. This new likelihood function is optimised based on the Expectation-Maximisation (EM) algorithm. Based on the sulphur recovery unit (SRU) process data and the debutanizer distillation process, the semi-supervised PCR model showed significant performance improvement compared to a vanilla PLSR model and the other traditional semi-supervised learning approaches. After three years, the same authors developed a Mixture form of the Semi-Supervised Probabilistic PCR (SSPPCR) model, which can perform better when data are generated from different operation modes \citep{gehu14}. A year later, in \citet{ge15}, they expanded on this mixture form of the SSPPCR model with a modified likelihood function incorporating the Bayesian regularisation term, which helps determine the optimal number of principal components used for modelling. Similarly, in \citet{zhge18}, the authors expanded the mixture form of the SSPPCR model by modifying its likelihoods, which incorporates the quantum effect for both labelled and unlabelled modelling situations. Both of these methods showed significant improvement in prediction performance and stability compared to the original mixture form of the SSPPCR model. With the promising results shown in this research, it is worth exploring their applicability for developing soft sensors for bioprocess applications. Some other examples of using both labelled and unlabelled data for the development of soft sensors and their adaptation over time can be found in the following two review papers in the broader process industry area \cite{kaga09,kagr11}.

    \subsection{Bayesian Methods}

    Bayesian methods refer to statistical techniques based on Bayes' theorem. These methods allow us to integrate our prior beliefs about uncertain events or parameters into ML models, which are used as prior knowledge and can guide and fast-track the learning process of the model. In general, Bayesian methods consist of the following three components:

    \begin{itemize}
        \item \textbf{Prior}: This represents what we know about the value of several certain parameters before observing the data. For example, we might assume the target variable's distribution to be normal based on our prior belief.
        \item \textbf{Likelihood}: This function tells us how probable it is to observe a specific data point given the distribution we have assumed as the prior.
        \item \textbf{Posterior}: The posterior is the updated belief about the distribution of the parameter, taking into account both the prior and the likelihood of the observed data. The posterior distribution is computed using Bayes' theorem, allowing us to make probabilistic statements about the parameters.
    \end{itemize}

    During the learning process of Bayesian methods, the distributions of these components will be explicitly modelled. Hence, Bayesian methods are inherently probabilistic and output a distribution of predictions, which is useful for measuring the uncertainty associated with the predictions. Bayesian methods can be applied to many tasks, including Bayesian Modelling, Bayesian Optimisation, Bayesian Inference, and Bayesian Decision Theory \citep{gi15, shsw16}. Their key ideas are explained below:

    \begin{itemize}
        \item \textbf{Bayesian modelling} is a statistical paradigm that deals with uncertainty explicitly by expressing variables as probability distributions. It provides a principled and coherent framework for uncertainty quantification and incorporates prior knowledge, making it particularly useful in scenarios with significant uncertainty or when we want to combine existing beliefs with data to make predictions or inferences. These properties and capabilities are desirable for our purpose, as building an ML model based on small data will likely be subject to a significant degree of uncertainty and model variance. Furthermore, many existing mechanistic bioprocess models can be seen as existing beliefs. They can be combined with ML models, potentially improving the model performance even when limited data is available. More discussion on hybrid modelling will be discussed in the next section.
        \item \textbf{Bayesian Optimisation} is a sequential design strategy for the global optimisation of black-box functions. It builds a probability model of the objective function and uses it to select the most promising points to evaluate. It is particularly useful for the optimisation of hyperparameters in ML models, as well as for optimising parameters in mechanistic bioprocess models.
        \item \textbf{Bayesian Inference} is a method of statistical inference that updates the probability for a hypothesis as more evidence or information becomes available using Bayes' theorem, which provides a mathematical model for how to update our beliefs about an unknown quantity, such as a model parameter, based on observed data. In Bayesian inference, the degree of belief in a hypothesis before evidence is considered is referred to as the "prior". The evidence is then used to update this belief to form the "posterior." The posterior probability combines our prior belief, the observed data, and how likely we were to observe that data under different hypotheses (this is the "likelihood"). 
        \item \textbf{Bayesian Decision Theory} is a decision-making framework that applies probability principles to making decisions under uncertainty. It's based on the fundamental idea that decisions should be made by considering both the possible outcomes of a decision and their associated probabilities, then choosing the action that maximises the expected utility, often subject to a cost function.
    \end{itemize}

    \subsubsection{Applications of Bayesian Methods for Small Bioprocess Data} 

    The performance of bioprocess (either data- or knowledge-driven) models depends on the underlying model and the amount and quality of the data available. However, the predictions from these models often have large uncertainty due to a lack of data and understanding of the bioprocess. Under such uncertainty, the single "best fit" bioprocess model resulting from maximum likelihood parameter estimation using limited available data may perform poorly in the production environment. As discussed in the last session, Bayesian methods are well-suited for a situation like this because they explicitly allow for the incorporation of prior knowledge or assumptions and the ability to quantify the uncertainty associated with the prediction. Given these useful properties of Bayesian methods, they have been widely adopted by the industry and research community for various applications \citep{khhu13}. The descriptions below present some key research results from the last 20 years in this area.

    \begin{itemize}
        \item \textbf{Bayesian Bioprocess modelling}: Bayesian-based state estimators or soft-sensors are commonly used in a bioprocess for state estimation and control optimisation as shown in \citet{kahe07,quam08,quam09,stpe18,alca20,fepa21,holu21,sist22}. In \citep{ammo16, roam16}, substrate feeding strategy and controls are developed and integrated with a Bayesian-based biomass estimator. Recently, more advanced Bayesian modelling techniques, such as the Hierarchical Bayesian Model and Bayesian Network, have also been gaining attention in bioprocess applications. In \citet{voos22}, the Hierarchical Bayesian Model was used to estimate biomass-specific, absolute enzymatic activity and predict optimal expression conditions for the CAR protein in the E. coli process. In \citet{zhwa21}, the study converts the N-linked glycosylation metabolic network into a Bayesian Network (BN), computing fluxes of the glycosylation process as a joint probability with culture parameters as inputs. This model, tested with different Chinese Hamster Ovary cell culture data, shows excellent predictive abilities. This validates BN models as valuable tools in the upstream process and medium development for glycoprotein production. Similarly, BN was utilised to map complex causal relationships between process parameters and product quality in \citet{xiwa22}. In addition, a Shapley value-based sensitivity analysis is employed to quantify each input factor's influence on output variations, along with a Bayesian posterior distribution for quantifying model uncertainty. The proposed approach can identify production bottlenecks, guide process specifications, optimise data collection, and enhance production stability.
        
        \item \textbf{Bayesian Optimisation for Experimental Design}: In \citet{nadi21}, Bayesian optimisation is combined with AL for identifying the formulation that optimises the thermal stability of three tandem single-chain Fv variants of the commercially available antibody Humira. Remarkably, this was achieved using only 25 experiments, a significantly lower number than the trials required by traditional methods such as DoE, full screening, or grid search techniques. Similarly, \citet{bana23} proposed the use of a multi-objective batch Bayesian optimisation (MOBBO) algorithm to maximise the vesicle-to-protein ratio and the enzymatic activity of the MSC-EV protein CD73 while minimising calregulin impurities of extracellular vesicles (EVs) production from a 3D culture of mesenchymal stem cells (MSCs). They achieved the optimal combination of process parameters, addressing the intended objectives, with merely 32 experiments, comparable to or even lower than the number required by the classical DoE or traditional one-factor-at-a-time (OFAT) methods. 
        
        \item \textbf{Bayesian Optimisation for First Principle Model}: Besides experimental design, Bayesian optimisation was also used to optimise bioprocess models. \citet{macr11} explored the application of Bayesian AL and probabilistic tendency models to address the challenge of small data in bioprocess optimisation. The authors proposed a run-to-run optimisation strategy, where imperfect models are continuously updated using Bayesian inference, leveraging experimental data to iteratively refine model parameters and optimise bioreactor performance. The study demonstrated that probabilistic tendency models—which explicitly characterize their levels of confidence—allow for systematic exploration and exploitation of operating conditions, improving productivity despite significant parametric and structural uncertainties. By utilising global sensitivity analysis (GSA), only the most influential parameters are updated, ensuring efficient data use. The methodology was validated on a penicillin G fed-batch bioreactor, where Bayesian optimisation significantly improved productivity even in the presence of a process-model mismatch. The paper highlighted that Bayesian approaches can systematically reduce uncertainty, making them particularly valuable for optimising bioprocesses with limited experimental data. A more sophisticated Bayesian optimisation method for the bioprocess model was proposed in \citet{ligu17}. The authors first define the posterior distribution of the model parameters and then generate an ensemble of model parameters by performing uniform distributed sampling on the parameter confidence region. Optimisation is then performed to maximise the lower confidence bound of the model prediction for productivity measures. The authors applied this ensemble-based process optimisation in mAb batch production using mammalian hybridoma cell culture, and it achieved better performance than the ML model-based optimisation.

        \item \textbf{Bayesian Inference for Bioprocess Monitoring}: In \citet{yu12}, Bayesian inference and a two-stage support vector regression (SVR) method were proposed for batch bioprocess. In this approach, Bayesian inference first identifies biases and misalignments in measurements using posterior probabilities. Then, the measurements are calibrated and used to build the second-stage SVR model, predicting new output measurements. This Bayesian Inference-based Two-Stage Support Vector Regression (BI-SVR) approach, applied to a fed-batch penicillin cultivation process, outperformed the conventional SVR method in predicting various output measurements, demonstrating its efficacy in handling different levels of measurement uncertainty.
        
        \item \textbf{Bayesian Decision Theory for Bioprocess Monitoring}: In \citet{sina01}, the authors used a Bayesian decision strategy in conjunction with the Gompertz model and Probabilistic neural networks to classify the lag, logarithmic and stationary phases in a batch process of Bacillus subtilis. The model can correctly classify 99\% of the training data and 95\% of the testing data. This suggests that the methodology has the potential for real-time implementation and automation of growth phase classification using only a single turbidimeter sensor.
    \end{itemize}

    \textit{Remark:} As shown from the results of these studies, Bayesian methods have become essential in upstream bioprocess for addressing data limitations and uncertainty, enabling the integration of prior knowledge, uncertainty quantification, and improved decision-making. Their applications span experimental design, bioprocess modelling, and monitoring, proving their success in improving predictive accuracy and reducing experimental burden under data-constrained scenarios. In terms of future research directions, as we have seen from \citet{nadi21, macr11}, combining Bayesian methods with other methods for addressing small data issues, such as AL, can lead to promising results. Thus, future research should focus on combining Bayesian methods with other techniques in this taxonomy, such as Hybrid Modelling, to further enhance process adaptability and efficiency in data-limited bioprocess environments.
    
    \subsection{Data- and Knowledge-driven Hybrid modelling}

    Data- and knowledge-driven hybrid (DKH) modelling is an approach that combines both data-driven modelling and knowledge-driven modelling, aiming to take advantage of the strengths of both methodologies to improve overall predictive accuracy, model interpretability, and generalizability. \textbf{Knowledge-driven model} (KDM), sometimes referred to as the first-principles model or mechanistic model, is a type of model that aims to describe the underlying mechanisms of a system \citep{joco08}. This means it attempts to understand and model the system's individual components and interactions. While a mechanistic model is often based on first principles, it can also incorporate other types of information, such as details about the specific mechanisms or processes at play in the system. For instance, in biological or chemical systems, a mechanistic model might describe specific biochemical reactions or physiological processes, which might be based on first principles (like the principles of chemistry) but also include more specific information about the particular reactions or processes involved. On the other hand, \textbf{Data-driven models} (DDM), often based on machine learning methods, are trained using large amounts of data \cite{kaga09}. These models learn patterns within the data and use these patterns to make predictions or decisions without being explicitly programmed. Data-driven models can capture complex patterns in the data that may not be accounted for by knowledge-driven models. However, the data-driven model relies on the variety, quality, and amount of available training data, and it could perform poorly when available data is limited. Thus, by combining these two types of models to form a hybrid model, we can leverage the power of ML to find complex patterns in data while incorporating expert knowledge to improve model interpretability and provide guidance during the model-learning process, especially when data is limited and insufficient for the ML model to learn effectively. 

    There are many ways to construct DKH models, and the specific approach used depends on the balance of knowledge and data available, the complexity of the system being modelled, and the specific goals of the modelling effort. A recent review discusses three major hybrid modelling methodologies \citep{brki22}: Hybrid Submodelling (HSM), Physics-Informed Machine Learning P-IML, and Model Calibration (MC). The details about each branch are discussed below:     

        

    \begin{itemize}        
        \item \textbf{Hybrid Submodelling}: HSM is a widely used hybrid modelling approach that integrates data-driven components within a mechanistic framework. It is primarily applied when some aspects of the system are well understood and can be modelled using first principles, while others remain unknown or too complex to model explicitly. The two main strategies in HSM include \textit{mechanism estimation}, where a data-driven model approximates unknown process relationships while maintaining consistency with mechanistic laws, and \textit{mechanism correction}, where a data-driven component corrects the discrepancies between mechanistic model predictions and observed data. 
        \item \textbf{Physics-Informed Machine Learning (P-IML)} - P-IML enhances traditional machine learning models by embedding physical constraints into the training process. This approach ensures that data-driven models remain consistent with established engineering and scientific principles, thereby improving generalisation and reliability. P-IML incorporates physics constraints in three key ways: \textit{Physics-Constrained Training}, where physical equations are added as penalty terms in the loss function of a machine learning model; \textit{Physics-Guided Model Architectures}, where network structures are designed to inherently satisfy known physical constraints; and \textit{Soft vs. Hard Constraints}, where soft constraints penalise constraint violations during optimisation while hard constraints strictly enforce physical laws.
        \item \textbf{Model Calibration}: MC adjusts the parameters of a mechanistic model based on experimental data. It is particularly useful when a first-principles model exists but contains uncertain parameters that need to be fine-tuned based on real observations. Potential model calibration techniques include \textit{Bayesian Calibration}, which uses prior distributions and Bayesian inference to update model parameters; \textit{Surrogate-Based Calibration}, where computationally expensive mechanistic models are approximated using surrogate models like GP Regression; and \textit{Multi-Fidelity Calibration}, which integrates low-fidelity simulations with sparse high-fidelity experimental data to optimise parameter estimation.
    \end{itemize}
    
    Recent advancements in scientific machine learning have introduced methods such as \textit{neural operators and universal differential equations (UDEs)}, which aim to learn entire differential equations rather than just individual parameters, which extend beyond traditional P-IML frameworks such as Neural ODE for integrating mechanistic knowledge with data-driven learning. \citet{agra23} provides a comprehensive review of the evolution of Hybrid Neural Network modelling in bioprocessing. It traces the transition from shallow hybrid models, which typically combine simple feedforward neural networks (FFNNs) with first-principles models, to deep hybrid models, incorporating advanced architectures such as Convolutional Neural Networks (CNNs), Long Short-Term Memory Networks (LSTMs), and Physics-Informed Neural Networks (P-INNs). This review has discussed several applications that demonstrated how hybrid modelling has significantly improved ML performance under small-data conditions.
    
    \subsubsection{Applications of Data-driven and Knowledge-driven Hybrid Model for Small Bioprocess Data}
    Hybrid modelling has been increasingly adopted in upstream bioprocessing applications as an effective strategy to overcome small data challenges, particularly in cases where purely mechanistic models require extensive calibration and purely data-driven models suffer from poor generalisation. Among the different hybrid modelling approaches, HSM is one of the most widely used approaches in hybrid modelling, particularly in chromatographic separations, mammalian cell culture modelling, and soft sensing applications, where some system relationships are well understood while others remain complex or unknown. For instance, in protein chromatographic separation, \citet{name04} developed a hybrid model integrating a first-principle general rate model with an ANN-based empirical model, allowing for rapid scenario exploration and optimisation without requiring large experimental datasets. Similarly, in therapeutic protein production, \citet{naso19} combined mass balance equations with ANN-based kinetic parameter estimation, improving model adaptability and predictive accuracy. Mechanism correction strategies have also been employed to improve real-time process monitoring, as seen in \citet{fega19}, where an Extended Kalman Filter (EKF) corrected mechanistic model predictions using Raman spectroscopy data, resulting in more reliable monoclonal antibody (mAb) concentration estimations. Additionally, hybrid modelling has been successfully applied in process scale-up and chromatography modelling. \citet{nase21} introduced an LKM-ANN hybrid model for protein chromatography breakthrough curve predictions, where the mechanistic Lumped Kinetic Model (LKM) was refined using an ANN to estimate adsorption behaviour, achieving three times lower prediction errors compared to using mechanistic models alone. Similarly, in bioprocess scale-up applications, \citet{badu21} integrated mechanistic mass balances with ANN-based learning to predict viable cell concentration and titer across different production scales, from 300 mL shake flasks to 15 L bioreactors, significantly reducing experimental burden while maintaining strong predictive performance. These studies demonstrate the broad applicability of HSM in handling process complexities, making it particularly effective in chromatography, real-time monitoring, and multi-scale bioprocess modelling.
    
    P-IML is relatively new and has gained traction as a strategy to enforce physical consistency in data-driven models while improving generalisation and reliability. In bioreactor system modelling, \citet{pime22} incorporated mass balance constraints into a deep learning framework, improving prediction accuracy by 18.4\% while reducing computational cost by 43.4\%. Other studies have explored the integration of Neural Ordinary Differential Equations (Neural ODEs) to bridge the gap between mechanistic and ML models. \citet{chdu24} developed a Neural ODE-based model to capture continuous temporal dynamics in bioprocesses, improving predictions of viable cell concentration, glucose consumption, and lactate production while reducing overfitting risks in data-limited environments. Similarly, \citet{sape23} extended Neural ODEs to optimal control applications, embedding ML-based feedback policies within a dynamical system model, resulting in improved robustness and predictive accuracy despite sparse training data. These studies highlight the growing role of PI-ML in bioprocess modelling, particularly in applications requiring real-time adaptability and physics-aware learning.
    
    Lastly, MC remains a critical approach for refining mechanistic models by adjusting parameters based on experimental data, ensuring robust and transferable process models. Bayesian Calibration has been particularly effective in optimising bioprocess models under uncertainty, as demonstrated by \citet{ligu17}, who employed a Bayesian ensemble-based optimisation framework to refine mammalian hybridoma cell culture models, improving prediction reliability. Surrogate-based calibration has also proven beneficial for computational efficiency, as seen in \citet{bobo24}, where a deep neural network-assisted mechanistic model calibration enabled fast parameter inference, significantly reducing computational costs while maintaining high accuracy. These approaches have been especially relevant in bioprocess control and optimisation, where parameter uncertainties can impact decision-making. 
    
    \textit{Remark}: The increasing adoption of Hybrid Submodelling, Physics-Informed Machine Learning, and Model Calibration in bioprocess applications reflects a growing industry need for hybrid models that balance interpretability, adaptability, and predictive accuracy under small-data conditions. Among these, HSM is the most widely applied, particularly in chromatography, cell culture modelling, and soft sensing, where unknown process dynamics need to be estimated or corrected. Meanwhile, PI-ML approaches are gaining momentum, especially in Neural ODE-based process modelling and bioreactor system control, ensuring physically consistent data-driven learning. Model Calibration remains crucial in parameter estimation and process scale-up modelling, ensuring that mechanistic models remain robust despite data scarcity. Future research should focus on hybrid modelling frameworks that integrate multiple strategies, such as combining HSM with Bayesian calibration to enhance predictive accuracy or embedding PI-ML into real-time adaptive control for dynamic bioprocess environments. Given the demonstrated effectiveness of DKH modelling in reducing data requirements and improving process generalisation, continued advancements in hybrid modelling architectures and uncertainty-aware learning will be critical for scaling up bioprocess production and optimising biopharmaceutical manufacturing under data-limited conditions.

    \subsection{Meta \& Transfer Learning}
    Both Meta Learning (Meta-L) and Transfer Learning (TL) aim to transfer knowledge from a source task, where a lot of data is available, to a target task, where less data might be available. The learned model, feature representations, or data instances from the source task are leveraged to improve learning efficiency and performance on the target task. The key distinction between Meta-L and TL lies in what knowledge is transferred. TL typically focus on the transfer of knowledge in data instances, such as fine-tuning a pre-trained neural network trained on a larger dataset to quickly adapt to new tasks with less data. In contrast, Meta-L focus on teaching the model how to learn based on its learning from experiences from other related tasks, such as optimising hyperparameters or architectures based on previous model learning experiences in other tasks. However, in the research community, they are often used loosely or interchangeably. Thus, we have organised them under the same session so their applications can be discussed together.

    \subsubsection*{Transfer Learning}  
\begin{figure}[h]
      \centering
      \includegraphics[width=\linewidth]{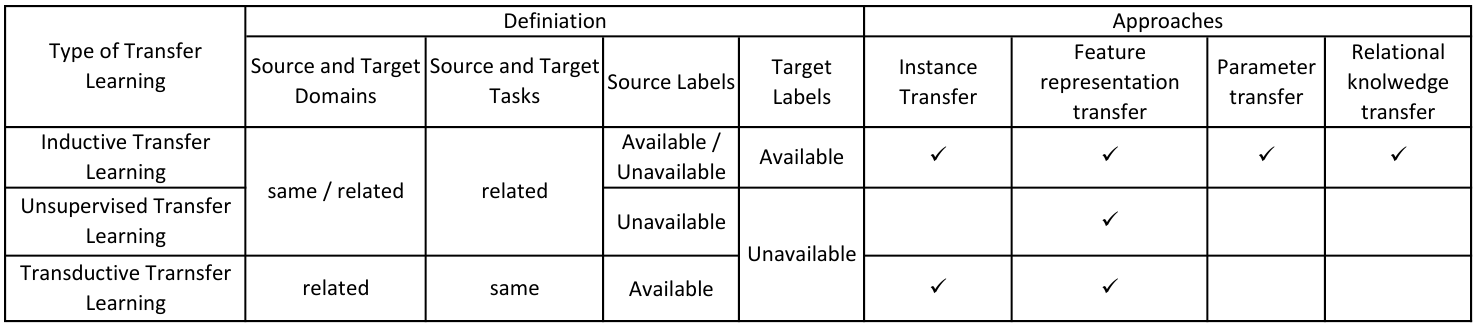}
      \caption{Major Branches and Approaches of Transfer Learning}\label{transfer_learning_methods}
    \end{figure}

    To better illustrate the learning problem of TL, we formally define it below: 
    \begin{itemize}
        \item Let \(D = \{X, P(X)\}\) represent a domain, where $X$ is the feature space, and correspondingly $P(X)$ is marginal probability distribution of the feature space.
        \item Let \(T = \{Y, P(Y|X)\}\) represent a task, where $Y$ is the label space, and $P(Y|X)$ is the conditional probability distribution of the label.
        \item Given Source domain \(D_s\), Source task \(T_s\), Target domain \(D_t\), Target task \(T_t\), TL aims to improve the learning of the target predictive function \(f_t\) in \(D_t\) using the knowledge in \(D_s\) and \(T_s\). Furthermore, \(D_s \neq D_t\) or \(T_s \neq T_t\). Otherwise, it would simply be a traditional ML problem. 
    \end{itemize}
    \citet{paya10} conducted a comprehensive survey on TL approaches, and their proposed taxonomy has been widely recognised by the research community, with more than 20,000 citations. Although it has been more than 10+ years since the survey was published, and some information may be outdated now, it still provides essential foundational knowledge about TL. As a result, we have briefly summarised it here in Figure \ref{transfer_learning_methods}.

    Based on the summarised results in Figure \ref{transfer_learning_methods}, we can conclude that, for the TL problem, $D_s$ has to be the same or at least related to $D_t$, and similarly, $T_s$ has to be related to or the same as $T_t$. This assumption is important as relatedness between domains and tasks determines the success of the TL. Suppose that no relationship exists between the source and target domains or tasks. In that case, the negative transfer would likely happen, which means that applying transfer learning hurts the performance of the target predictive function \(f_t\) in \(D_t\). For two domains $D_s$ and $D_t$ to be considered related, an explicit or implicit relationship exists between the feature spaces of the source and target domains. Similarly, for two tasks $T_s$ and $T_t$ to be considered related, an explicit or implicit relationship must exist between their label space. The TL approaches that can be applied would differ depending on whether target and source labels exist. Below is a high-level summary of the key idea of each of these TL approaches.

    \begin{itemize}
        \item \textbf{Instance Transfer}: This approach assumes certain parts of the data in the source domain can be reused together with the labelled data in the target domain for learning. Instance reweighting and importance sampling are the two main approaches to performing instance transfer with a different focus. 
            \begin{itemize}
                \item Instance Reweighting: The key idea behind instance reweighting is that not all instances in the source domain are equally useful for learning a model that generalises well to the target domain. By assigning higher weights to the instances that are more similar to those of the target domain, the model is encouraged to pay more attention to these instances during training and vice versa. Instance reweighting prevents the model from picking up noise or learning from low-relevance instances in the source domain, thus improving the model's ability to generalise in the target domain.
                \item Importance Sampling: Importance sampling reweights the source data instances according to the ratio of target and source instance densities (importance weights), which shifts the source distribution to be similar to the target distribution. Formally, if $p_s(x)$ and $p_t(x)$ are the source and target data distributions, respectively, the importance weight for an instance x is given by the ratio $p_t(x) / p_s(x)$. This requires an estimation of the source and target data densities, which can be challenging in high-dimensional spaces. 
            \end{itemize}
        
         While both methods aim to adjust the source distribution to behave similarly to the target distribution, both techniques involve weighting instances. The key difference is that instance reweighting weights instances based on the similarity between the individual instances. In contrast, importance sampling weights the instances based on the similarity between data densities. The appropriate method depends on the specific characteristics of the source and target domains and the data availability in each domain.
    
        \item \textbf{Feature Representation Transfer}: The key idea of this approach is to learn a transformation that maps features in the source and target domains to a common representation. The assumption here is that while the data distributions in the source and target domains are different, they can be made similar by learning a mapping or transformation between their feature spaces. An advantage of this approach is that it can be applied even if no labelled data is available in both source and target domains, while there are also supervised feature construction methods that can leverage labelled data if available. 
        \item \textbf{Parameter Transfer}: This approach transfers knowledge from the source task to the target task by sharing parameters or priors between the source and target models. It operates under the assumption that different tasks share some common parameters or priors. Transferring these parameters or priors can improve the learning performance on the target task, especially when the target task does not have sufficient labelled data. Popular transfer learning techniques such as Fine-Tuning, Multitask Learning, and Neural Architecture Search belong to this category. 
        \item \textbf{Relational knowledge Transfer}: This approach transfers the relational knowledge (i.e., relationships between instances or features) from the source domain to the target domain. This method assumes that although the source and target data may reside in different feature spaces, the relationships or interactions between instances or features may be similar across domains. Relational Knowledge Transfer is useful in tasks where the data are structured and relationships between instances or features are important. This includes natural language processing, social network analysis, and bioinformatics.
    \end{itemize}

    \subsubsection*{Meta Learning}
      Meta-learning, or "learning to learn", aims to design models that can learn new skills or rapidly adapt to new tasks with a few training examples \cite{huva21}. Meta-learning algorithms learn the underlying structure of the data across multiple tasks during the meta-training phase. During the meta-testing phase, they can quickly adapt to new, unseen tasks using only a small number of samples. To better illustrate the learning process of meta-learning, we have formally defined the learning objective of Meta-Learning as follows:
        
        \begin{itemize}
          \item Given a set of tasks $T_i\in\mathcal{T}$ with individual task $T_i$ sampled from a task distribution $p(\mathcal{T})$. Each task $T_i$ consists of a learning problem with a corresponding dataset $\mathcal{D}_i$.
        
          \item For each task $T_i$, the dataset $\mathcal{D}_i$ can be partitioned into two subsets: a support set $S_i$ and a query set $Q_i$. The support set $S_i$ is used for task-specific adaptation, while the query set $Q_i$ is used to evaluate the performance of the adapted model on the task $T_i$.
        
          \item The goal of the meta-learning algorithm is to learn a set of parameters $\theta$ of meta-learner $f$ or a learning procedure that can efficiently adapt to new tasks $T_i$ from the distribution $p(\mathcal{T})$ using the support set $S_i$. The success of the adaptation is measured based on the generalisation error of model $f$ with updated parameters $\theta'$, which is calculated using a selected loss function $\mathcal{L}$ and the query set $Q_i$. In mathematical terms, we could define $\theta$ as:
          \begin{equation}
             \theta = \min_{\theta} \mathbb{E}_{T_i \sim p(\mathcal{T})} \left[ \mathcal{L}_{Q_i} \left( f_{\theta\xrightarrow[S_i]{}\theta'} \right) \right],
          \end{equation}        
        \end{itemize}

    To achieve the learning objective defined above, three main types of Meta-Learning approaches can be applied, which are Metrics-based, Optimisation-based, and Model-based  \cite{pale22}. The high-level description of these methods is discussed below, along with the advantages and disadvantages of each method summarised in Table \ref{meta_learning_methods}.

    \begin{ThreePartTable}
          \vspace{2ex}
          \begin{ltabulary}{P{1.5cm}P{3.5cm}P{4.5cm}P{4cm}}
          \caption{Comparison of Meta-Learning Methods. \emph{S} is the support set\label{meta_learning_methods}}\\
                \toprule
                  & \textbf{Metrics-based} & \textbf{Optimization-based} & \textbf{Model-based} \\
                \midrule
                \textbf{Key idea} & Metrics Learning. & Gradient Descent. & Memory; RNN. \\
                \midrule
                \textbf{Advantages} & Faster Inference. & Flexibility to optimise in dynamic environments. &                                Faster inference with memory models. \\ 
                                    & Easy to deploy. & \emph{S} can be discarded post-optimization. & Eliminates the need for defining a metric or optimising at test. \\
                \midrule
                \textbf{Disadvantage} & Less adaptive to optimisation in dynamic environments. & Optimisation at inference is undesirable for real-world deployment. & Less efficient to hold data in memory as \emph{S} grows. \\ 
                & Computational complexity grows linearly with size of \emph{S} at test. & Prone to over-fitting. & Hard to design. \\
                \bottomrule
          \end{ltabulary}
        \end{ThreePartTable}
        
        \begin{itemize}
            \item \textbf{Metrics-based} method is an approach that aims to learn a distance metric or similarity function to compare and relate data points from different tasks. This learned metric can determine how similar new data points are to the existing ones, allowing for efficient adaptation and generalisation to new tasks with limited labelled data.
            \item \textbf{Optimisation-based} methods in the context of meta-learning refer to approaches that focus on learning an optimal set of initial parameters, model architectures, or learning algorithms to adapt quickly and effectively to new tasks. By leveraging prior knowledge and experience from multiple tasks, optimisation-based meta-learning systems aim to find a suitable starting point or an adaptable learning process that can be fine-tuned with minimal training data or updates for new tasks. 
            \item \textbf{Model-based} methods approach that utilise an internal model or memory mechanism to learn and store information across different tasks. These methods focus on learning a model that can efficiently encode, store, and retrieve knowledge from prior experiences, allowing the model to quickly adapt and generalise to new tasks with limited data. 
        \end{itemize} 

        By applying these meta-learning methods, a meta-learner can be learned during the process, and "meta-features" that are useful across all tasks can also be identified. Meta-features refer to a feature or attribute that describes certain data characteristics or the learning task that can be captured across tasks. For instance, meta-features can include characteristics such as the number of instances in the dataset, the number of features, class distribution, measures of feature correlation, feature entropy, and other statistical properties. In the context of machine learning tasks, meta-features can also include elements like the type of the task (classification, regression, clustering, etc.), the performance of certain baseline models, or the computational cost of different learning algorithms. Meta-features are particularly useful in meta-learning scenarios such as algorithm selection and hyperparameter optimisation, where the goal is to learn how to select the best algorithm or optimise the hyperparameters for a given task based on its meta-features. By effectively capturing and utilising these meta-features, a meta-learner can make more informed decisions about the best approach to a given learning task, potentially leading to improved performance and efficiency.
        
    \subsubsection{Applications of Meta \& Transfer Learning for Small Bioprocess Data} 

    In recent years, Meta-L and TL have been gaining popularity and adopted across various types of domains and industries, including bioprocessing \citep{isco24}. The reviewed studies highlight four primary bioprocessing applications where transfer learning has been successfully implemented: \textbf{cell-line and bioprocess condition generalisation}, \textbf{batch-to-batch process optimisation}, \textbf{strain-specific knowledge transfer}, and \textbf{soft sensor recalibration for real-time monitoring}.

    One of the most impactful applications of transfer learning in upstream bioprocessing is cell-line and bioprocess condition generalisation, where ML models trained on one bioprocess setup are adapted to new products, media compositions, or reactor configurations. \citet{hust21} introduced a hybrid GP model with entity embedding vectors, allowing knowledge transfer across cell lines and process conditions. Unlike conventional one-hot encoding, which treats each product separately, embedding vectors capture latent similarities between different production systems, enabling more accurate predictions for novel bioprocess conditions with minimal new data. By integrating prior knowledge from historical datasets, this approach significantly reduces the number of required wet-lab experiments, making it particularly useful for accelerating process development and scale-up across multiple bioproducts.

    Another key bioprocess application of transfer learning is batch-to-batch process optimisation, where ML models leverage prior knowledge from previous production runs to improve process control in new batches. \citet{pesa20} developed a two-stage reinforcement learning (RL) framework that first pre-trains an RL controller on simulated mechanistic data before fine-tuning it with real experimental data from small-batch runs. This transfer learning-based batch-to-batch adaptation allowed for faster policy convergence, reduced data requirements, and improved process control compared to Nonlinear Model Predictive Control (NMPC). By freezing lower network layers and fine-tuning only the final control layers, this study demonstrated that transfer learning significantly accelerates process adaptation and enhances robustness in batch bioprocess control. Such an approach is particularly valuable in bioprocess industries where each batch is unique and models must continuously adapt to changing process conditions. Similar results have been demonstrated in \citet{hewi24}, where Meta-L was leveraged for improving bioprocess characterisation in scenarios with limited data. The study explores the effectiveness of two knowledge transfer methods - meta-learning and one-hot encoding combined with GP models to enhance model performance when experimental data for a new bioprocess are scarce. Specifically, the PAC-optimal hyper-posterior (PACOH) meta-learning framework was adapted to GP regression models, enabling informed initialisation of GP parameters based on historical bioprocess data. This meta-learning approach significantly reduced prediction errors compared to models trained solely on new process data, particularly excelling with heterogeneous datasets. Additionally, the authors introduced the concept of "calibration design," an experimental strategy that optimally selects a minimal set of experiments for model adaptation by maximising the dissimilarity in process responses of historical data. Tested using simulated mammalian cell culture data, this method achieved accuracy comparable to traditional Design of Experiments (DoE) approaches while requiring up to four times fewer experiments. The study concluded that systematic knowledge transfer and calibration design could streamline bioprocess development, substantially reducing experimental requirements while improving predictive accuracy.
    
    Transfer learning has also proven highly effective in strain-specific knowledge transfer, where ML models trained on one microbial strain are adapted to predict bioprocess behaviour in a less-characterised strain. \citet{rove22} developed a predictive modelling framework that transfers kinetic knowledge from well-studied microbial strains to new, less-characterised strains, reducing the data requirements for predictive modelling. The study presented two case studies: transferring bioprocess knowledge from Desmodesmus sp. to Chlorella sorokiniana for lutein production and extending an ANN model trained on salicylic acid-stimulated Desmodesmus sp. to predict succinic acid-stimulated growth. By applying partial model retraining and penalty-based parameter adaptation, the study demonstrated that transfer learning significantly improved model accuracy while reducing uncertainty, confirming that pre-trained kinetic models can be efficiently adapted to new microbial systems. This technique is particularly beneficial for scaling up new microbial-based bioproduction platforms, where data availability for novel strains is often limited.
    
    A fourth major application of transfer learning in bioprocessing is soft sensor recalibration for real-time monitoring, where pre-trained ML models dynamically adapt to new bioprocess conditions through similarity-based calibration updates. \citet{sika23} introduced a generalist soft sensor concept, where multiway principal component analysis (MPCA) and k-nearest neighbour (KNN) similarity analysis enabled knowledge transfer across different bioprocesses, media compositions, and microbial strains. Instead of training a separate soft sensor for each condition, the approach identifies similar historical process data and recalibrates the soft sensor dynamically, allowing for real-time adaptation with minimal new experimental data. The study demonstrated successful knowledge transfer between Pichia pastoris and Bacillus subtilis bioprocesses, achieving high predictive accuracy for biomass and protein concentration (relative errors of 6.9\%–12.8\% across different conditions). These results highlighted that similarity-driven recalibration can significantly enhance soft sensor generalisation, making it particularly valuable for real-time bioprocess monitoring in variable production environments. 

   Besides bioprocess applications, numerous Meta-L and transfer learning (TL) methods have been developed for soft sensor construction in various industries. For multi-grade industrial processes, TL effectively transfers learned knowledge between different soft sensor models, particularly beneficial when data for a new operating grade is limited. Liu et al. \cite{liya19} introduced Domain Adaptation Extreme Learning Machine (DAELM), combining data from source and target domains while using a regularisation term to adjust the influence of each domain. This approach was further enhanced by adversarial transfer learning (ATL-DAELM), employing GANs to align source domain data distribution with the target domain \cite{liya20}. Although effective, DAELM and ATL-DAELM can handle only a single source domain. To address this, Wang et al. \cite{wazh20} proposed incorporating multiple source models into the regularization term, allowing flexible knowledge transfer based on domain relevance. Another straightforward approach, Frustratingly Easy Domain Adaptation (FEDA), was verified by Yamada et al. \cite{yaka22} to efficiently predict product quality with minimal computational effort, suitable for online learning scenarios.

    Beyond instance transfer, parameter and feature representation transfers also benefit soft sensor development. Hsiao et al. \cite{hska21} demonstrated parameter transfer effectiveness by fine-tuning a Feed Forward Network (FFN) with limited target domain data for a distillation column, significantly outperforming models trained solely on the target domain. Liu et al. \cite{liho21} introduced high-order partial least squares (HOPLS) to extract common and specialised features from multi-grade process data, significantly enhancing prediction accuracy over standard PLS models. Chai et al. \cite{chzh22} developed a deep probabilistic transfer regression (DPTR) method, integrating a deep generative regression model (DGRM) with adversarial learning for feature alignment across domains, showing superior accuracy compared to DAELM. 
    
    Meta-learning could similarly enhance soft sensor development by training a meta-learner across diverse batches and quickly adapting it to new conditions with minimal feedback. Although unexplored in bioprocessing contexts, Xiaoyong et al. \cite{xiya22} successfully applied a modified Model-Agnostic Meta-Learning (MAML) combined with K-Means clustering to enhance soft sensor accuracy in the Purified Terephthalic Acid (PTA) solvent system, achieving over 70\% improvement in prediction accuracy compared to conventional methods.

    
    
    \textit{Remark:} Future research in transfer learning and meta-learning for bioprocess applications should explore hybrid approaches that integrate multiple learning strategies to enhance model flexibility and predictive accuracy further. For instance, combining entity embedding-based knowledge transfer with phase-aware recalibration and meta-learning frameworks could enable more robust bioprocess monitoring across multiple production scales and different process conditions. Additionally, integrating batch-to-batch reinforcement learning, meta-learning for rapid parameter initialisation, and strain-specific parameter adaptation could optimise bioprocess control strategies for microbial fermentation and therapeutic protein production more effectively. Advances in self-supervised transfer learning techniques, combined with meta-learning algorithms that autonomously identify and rapidly adapt relevant features for cross-process adaptation, could significantly enhance knowledge generalisation and data efficiency in upstream bioprocess development. Given the demonstrated success of meta-learning and transfer learning in reducing data requirements and improving process generalisation, continued advancements in multi-domain transfer, meta-learning initialisation, and process-aware adaptation techniques will be essential for accelerating bioprocess scale-up and improving real-time decision-making under limited data constraints.

\section{ML Methods for Addressing Small Data Challenges during Continuous Monitoring and Maintenance}
\label{sec:model-retraining}
Continuous Monitoring and Maintenance are important steps in the Machine Learning Operation (MLOps) lifecycle. ML models' performance may deteriorate over time due to changes in data and environment. Thus, continuous monitoring allows for early detection of model deterioration, so that necessary model maintenance action can be taken to improve the model's performance. Typically, there are two ways to improve model performance: model retraining and model adaptation. Model retraining involves training a model from scratch instead of improving on the existing model's learned structure and parameters. On the other hand, model adaptation involves taking a pre-trained model and slightly modifying its structure or parameters by learning it from the new data.

There are two main reasons for models to be retrained or adapted. Firstly, it allows the model to adapt to changes in the underlying data generation process or environment \cite{kaga09b,kaga09c,kaga09d,kaga10,kaga10a,kagr11,baga15,sabu16a,baga17,baga18,kemu24}. There are many reasons for such a change to occur in industrial processes, ranging from a change of equipment, addition, removal or failure of sensors, or simply changing operating conditions affecting the observed data distribution. Such gradual data distribution change in the form of either \textbf{data drift} or \textbf{concept drift} \citep{wiku96}, while one of the easiest to tackle, has been one of the most often studied forms of change in the existing literature. The former occurs when there is a change in input data distributions, and the latter occurs when the relationship between the input data and the output target variable has changed. Model performance may deteriorate because of these changes. Thus, when we are expecting or have detected data and/or concept drifts, model retraining or other forms of adaptation allow the model to adapt to the changes in data and achieve better performance. Secondly, even if no changes in data are detected or anticipated, retraining it or continuous incremental learning with additional data could improve the model's performance if the model was trained on insufficient data. Note that insufficient data could even result from gradual changes such as data or concept drift, as when data distribution changes, the relevance of historical data is reduced, as they are no longer representative of future data. As a result, the information in some of the data points may no longer be useful or even harmful for the model to learn from, which causes insufficient data available for model training.     
    
Many model adaptation and retraining strategies exist, and different strategies could suit different scenarios and may have varying impacts on the performance of the model \cite{sabu16a}. Picking the right re-training strategy is particularly important for models built on small data because the smaller the data, the more weight the new data will carry during retraining, causing a bigger impact on the model. The most naive and intuitive retraining strategy would be re-running the whole training pipeline with new data included in the training data. It could be effective when the underlying data distribution does not change. However, it is not the most optimal strategy when the underlying data distribution changes, as the new data may now contain significantly more relevant and useful information for improving the model's ability to generalise compared to the historical data. It would be naive to include all historical data in the training set and assume all data carries the same weight. 

In the following section, we will introduce different sets of model retraining and adaption methods, which focuses on constructing ML models in a way that can adapt to new data and changed data distribution as fast as possible, thus requiring the least amount of new data to complete the adaptation, e.g. the model has reached to the point that new data can no longer improve the model performance. These methods are crucial for many practical applications where getting new data is expensive, and we will review their efficacy and relevance to small data applications in detail.

\subsection{Model Retraining}
    \subsubsection*{Moving Window Retraining} 
    This technique involves retraining the model by using the data points within a moving time window. As new data is gathered, the moving window is often set to incorporate the newest data while eliminating the oldest data for modelling purposes, as the most recent data points are believed to be the most pertinent to the current context. Model retraining can happen when either:

    \begin{itemize}
        \item A new data point arrived, known as sample-wise \citep{likr09}.
        \item Or after gathering a specific number of data points (step size), known as block-wise \citep{lejo05}. 
    \end{itemize}
    
    Models can also be retrained using two methods. The first involves completely retraining the model using the data within the moving window. The merit of this technique is that the initial learning algorithm can be reused during the online phase, eliminating the need for a unique online training method. This universality makes it compatible with any ML model. Hence, this approach is prevalent in adaptive model applications. The second method utilises a two-phase process. It starts by "downdating" to eliminate the oldest data point and then "updating" to integrate the most recent data samples \citep{likr09}. 
    The main challenge of the moving window approach is the storage requirement of all data within the moving window, which can pose issues for large windows in memory-constrained settings. Two crucial parameters significantly affect model efficacy: the adaptation window's size and the intervals between updates, known as the step size. Incorrect settings can diminish performance instead of enhancing it \citep{ka09,baga18}. Ideally, the window size should match the data's concept length. Failing to meet this can result in the model adapting to noise with a short window or limited adaptability to abrupt changes with a lengthy one. Fixed window and step sizes also bypass the task of concept drift detection. Without assurances of correct window and interval settings that match the underlying process dynamics, this approach can yield suboptimal results. An alternative is the adaptive window size technique \citep{heya08, zhwa07}, which dynamically reduces the window size when concept drifts are detected. However, this technique still does not address the catastrophic forgetting issue that moving window-based techniques might have, as continuously removing the oldest data within the time window might inadvertently discard data that contains beneficial information.
   
    \subsubsection*{Just-In-Time Learning}   
    Just-In-Time Learning (JITL) has many other names, such as lazy learning, instance-based learning, and on-demand modelling \citep{kafU13a, sa14, jich15,yuge14}. Concepts of local modelling inspire JIT learning; the idea behind JITL is to store many past observations and then, when faced with new input data, dynamically craft predictive models based on the most similar (to the new input data) subset of these historical data. In contrast with other adaptive models, crafted in advance and then adapted to new data as they come through, JIT models are developed from scratch in real-time as new data arrive. This real-time approach allows JIT models to represent the current state of a process better and gives JIT models the ability to handle gradual and sudden operational changes. 
    
    The major limitation of JITL is its reliance on historical data that mirrors new incoming data. The "similarity" between old and new data is often determined by various metrics, including how close data points are to each other, with methods like Euclidean distance being popular due to their straightforwardness. Some algorithms, like the correlation-based JIT (CoJIT) algorithm \citep{fuka12}, use correlation to determine similarity and have shown even better results in prediction accuracy. There are numerous JIT algorithms differentiated by how they measure similarity or the local models they employ. These models can be categorised based on the nature of the data they handle, such as their distribution type (e.g. Gaussian), linearity, and data completeness. Examples of these local models span a wide range, from PCR and PLSR to locally weighted regression and PLS algorithms. In a nutshell, JITL is an adaptive approach where predictive models are crafted as and when new data arrives, ensuring that they are always tailored to the most recent data trends.
    
\subsection{Model Adaptation}    
        
    \subsubsection*{Recursive / Instance weighting techniques}  

    Recursive or instance weighting techniques are used to adapt models to new data by leveraging their previous state or predictions, and this allows models to adjust dynamically as they encounter more data. Many online/incremental learning algorithms, such as Least Squares \citep{dama97}, Principal Component Analysis \citep{liyu00}, and Partial Least Squares methods \citep{qi98}, leverage this technique to enable model adaptation. Recursive algorithms are particularly useful in scenarios where the data is sequential or has temporal dependencies, as these models can carry information from previous data points to make predictions on new ones, and they typically involve giving less importance to the older model using a "forgetting factor." Like the moving window approach, these methods can incorporate a new data point (for sample-wise operations) or a block (for block-wise operations) to perform model adaptation. 
    
    Recursive methods also face challenges related to parameter estimation. Taking the example where weights get assigned based on the data sample's age, a critical determinant becomes the rate at which these weights decrease over time. Like the moving window approach, recursive methods inherently lack a mechanism for detecting concept drift unless the decay rate adapts to the speed of concept changes. However, there are strategies to adjust the forgetting factor dynamically. One prevalent method for dynamically adjusting the forgetting factor within the recursive linear least squares framework is to adjust the forgetting factor's value based on the product of Hotelling’s $T^2$ value and the Squared Prediction Error (SPE) to maintain the estimator's consistent information content \citep{foke81}. Another limitation of recursive methods is that, due to their sequential update nature \citep{kafu13}, they are sensitive to abrupt changes in data \citep{kagr11,sa14,geso13} and could perform unnecessary updates and adapt to noise. One potential way to address this issue is to adapt the model with data in mini-batches of a moving window, e.g., the last 100 data points, or data in the last 7 days. This ensures the models can adapt to the latest changes while not overreacting to abrupt changes.      
    
    \subsubsection*{Ensemble-based methods}     
    The third model adaptation technique discussed here hinges on the ensemble learning framework \citep{saro18,doyu20}. Within this framework, multiple models are simultaneously utilised and managed to predict a hard-to-measure variable. For the ensemble's final prediction, individual model predictions must be merged. A straightforward and popular merging technique takes the weighted sum of the predictions from individual models as the final prediction. Compared to age-based methods, it offers additional adaptability and enables concept drift detection and handling by adjusting the prediction weights. When the prediction weights of a model are set to zero, we essentially remove that model from the ensemble, e.g. model instance selection. Another strategy at this level involves introducing and integrating new models into the ensemble when a new data concept is identified, especially during sudden process shifts. Some more details and further discussion on multiple adaptive mechanisms within predictive systems utilising multiple predictor frameworks can be found here \cite{baga17,baga18}.
    
    The major downside of ensemble-based methods for concept drift detection and handling is that they increase model complexity and computational load due to managing multiple models and their respective weights. Despite this disadvantage, in adaptive soft sensing, ensemble methods like Neuro-Fuzzy Systems (NFS) \cite{ga04}, Incremental Local Learning-based Soft sensing Algorithms (ILLSA), and other hybrid techniques \citep{li07,kaga09d} are still prevalent due to their superior performance. 
    
    \subsection{Applications of Model Retraining and Adaptation for Small Bioprocess Data}

    The research on model retraining and adaptation for bioprocess development has been around for a long time. \citet{syho97} published in 1997 developed a time-sequenced moving window approach for adaptive neural networks in 2,3-butanediol fermentation, allowing the model to continuously adjust its weights in response to changing process conditions. Similarly, \citet{esmo22} proposed an optimisation-based moving window approach for model adaptation, which refines metabolic state estimations in Escherichia coli fermentation by integrating past and present process data in a memory effect-based model adaptation framework. These studies highlight that moving window is a simple yet effective technique to incorporate AM into ML models, improving prediction accuracy in scenarios where process dynamics evolve over time, such as those observed in bioprocesses.
    
    Recursive and instance weighting approaches have been particularly effective in real-time bioprocess control and fault-tolerant state estimation, as they dynamically reweight or update model parameters in response to process variability. Several studies have explored recursive Levenberg-Marquardt (L-M) learning algorithms to adaptively estimate process states in continuous stirred tank reactor (CSTR) bioprocesses. \citet{baes07} and \citet{bama09} developed a Recurrent Neural Network (RNN) architecture with recursive L-M learning, which enabled real-time model adaptation for wastewater treatment bioprocesses by continuously updating network weights based on process deviations. A similar Kalman Filter-based Recurrent Neural Network (KFRNN) framework was introduced in \citet{bama09, bama09a, bama10}, where Kalman filtering and recursive neural learning improved process state estimation under noisy and uncertain conditions. Additionally, adaptive control applications, such as \citet{chli13}, demonstrated how recursive kernel learning could continuously adjust nonlinear model parameters to improve predictive control of time-varying bioprocesses. These findings indicate that recursive learning approaches are highly effective in adaptive bioprocess control and real-time process monitoring, allowing models to self-correct based on evolving process trends even when trained on limited data. A recent study by \citet{turo2024} explored the application of OL for real-time monitoring of nutrient and metabolite concentrations in monoclonal antibody (mAb) production using Raman spectroscopy as input data. The author leveraged recursive-based OL algorithms—such as online support vector regression (OSVR) and recursive partial least squares regression (RPLSR) to continuously update model parameters based on newly available data, adapting to dynamic process conditions. The results indicate that this OL algorithm significantly outperforms static models, particularly when confronted with shifting process conditions, such as glucose feeding events, that induce non-stationary changes in metabolite behaviour. These findings reinforce the effectiveness of recursive-based model adaptation strategies in enhancing real-time bioprocess monitoring and improving model robustness while reducing the dependency on large training datasets.
    
    Ensemble-based model adaptation techniques have also been widely adopted in soft sensor development and sensor fault detection, as they dynamically adjust the contribution of different models based on their reliability. One notable application is adaptive soft sensing for biomass monitoring, where \citet{sibr22} developed an ensemble-based adaptive soft sensor that integrates three independent submodels for biomass concentration estimation in Pichia pastoris fermentation. The ensemble dynamically adjusts submodel weights using moving window regression and variance-based adaptation, ensuring that unreliable submodels are down-weighted in the presence of sensor noise or drift. Similar ensemble-based model adaptation methods have been applied in real-time sensor calibration, such as \citet{brkl20}, where swarm intelligence techniques were used to dynamically select the best predictive models at different bioprocess phases. These findings suggest that the ensemble-based model adaptation method is particularly useful for fault-tolerant bioprocess monitoring, as it enables real-time model adaptation without requiring frequent manual recalibration.
    
    JITL has emerged as one of the most effective model retraining approaches in Raman spectroscopy-based bioprocess monitoring, offering superior flexibility in real-time model calibration under small data constraints. Unlike traditional global calibration models, which require large historical datasets, JITL dynamically selects the most relevant local training data for each new query, ensuring that models remain relevant to changing bioprocess conditions. \citet{tusc19, tuwa20, tukh21} introduced Real-Time JITL (RT-JITL) and Spatiotemporal JITL (ST-JITL) for cell culture monitoring, where GP was used to enhance real-time adaptation, improving glucose and lactate concentration predictions by over 60\% compared to traditional global models. Further advancements in JITL include hybrid approaches, such as the integration of JITL with Variational Autoencoders (VAEs), as demonstrated in \citet{rakh24}, where VAE-extracted Raman spectral features improved similarity-based data selection, enhancing model robustness in heterogeneous bioprocess conditions. These studies highlight the superiority of JITL techniques in bioprocess monitoring applications with high variability, where continuous recalibration is necessary to account for shifts in raw materials, cell lines, and process conditions.
    
    Among the different model retraining and adaptation techniques, JITL approaches have recently gained the most traction in upstream bioprocess applications, primarily due to their effectiveness in soft sensing and real-time calibration. Making JITL the leading retraining and adaptation strategy in Raman-based bioprocess monitoring, as it enables real-time local calibration updates, significantly outperforming traditional global models in predictive accuracy. Meanwhile, recursive/instance weighting and ensemble-based approaches remain valuable in real-time bioprocess control and sensor fault detection, where they help mitigate process variability and improve model reliability.
    
    \textit{Remark:} Future research in model retraining and adaptation methods for upstream bioprocessing should explore hybrid approaches that combine multiple model retraining and adapation techniques to further enhance model generalisation and fault tolerance as those explored elsewhere in the broader process industry applications \cite{baga17,baga18,kaga09b,kaga09d,kaga10,kaga09c}. For instance, integrating JITL with recursive learning algorithms could enable continuous state estimation in metabolic modelling, while ensemble-based JITL frameworks could enhance real-time sensor calibration across different bioprocess scales. Additionally, the application of self-supervised adaptation—where models autonomously refine their calibration without requiring labelled data—could further improve bioprocess decision-making under extreme data scarcity. Given the demonstrated benefits of model retraining and adaptation in bioprocess optimisation, real-time control, and soft sensing, continued advancements in multi-modal adaptive frameworks will be crucial for enabling more resilient and data-efficient bioprocess monitoring systems.

\section{Discussions and Future Research Directions}\label{discusions}

The application of machine learning (ML) in upstream bioprocessing is gaining significant traction, particularly in the context of small data challenges. As discussed throughout this review, ML models typically require large datasets to perform effectively. However, in biopharmaceutical manufacturing, the cost and feasibility of generating large-scale data are limited. This necessitates innovative ML approaches that can extract meaningful insights from limited data, ensuring robust and reliable bioprocess monitoring and optimisation.

One of the key findings of this review is the necessity of hybrid modelling techniques that integrate domain knowledge with ML approaches. Traditional mechanistic models have long been used for process understanding and control, but their predictive capabilities can be enhanced by data-driven methods such as transfer learning and semi-supervised learning. By leveraging prior domain knowledge in the form of physics-based models or Bayesian priors, these hybrid models improve generalisation and mitigate overfitting, making them particularly suitable for small data applications.

Transfer and Meta-learning have also shown promising results in domains where rapid adaptation to new tasks is required. Recent studies have demonstrated the effectiveness of optimisation-based meta-learning, such as Transferrable Model-Agnostic Meta-Learning (T-MAML), in predicting outcomes with limited training data \citep{helu22}. Thus, A similar approach could be applied to upstream bioprocessing by training meta-learners on historical bioreaction batch data, allowing them to quickly adapt to new bioreaction runs with minimal labelled data. This could be particularly useful for predicting CQAs and CPPs at early bioprocessing stages. Furthermore, transfer and meta-learning could be leveraged in soft sensor development for bioprocess monitoring, where a meta-learner trained on multiple bioreactor runs could be deployed and fine-tuned for new environments with limited feedback. Although no study has been done on this topic for bioprocess applications, prior research has demonstrated the effectiveness of meta-learning in soft sensor development \citep{xiya22}, where a modified Model-Agnostic Meta-Learning (MAML) approach based on K-Means (KM) improved soft sensor performance in industrial applications.

Another significant aspect explored in this review is the role of active learning in reducing experimental burden while maximising information gain. Active learning strategies have shown promise in selecting the most informative data points for labelling, thereby improving model performance with minimal data acquisition. This is particularly relevant in upstream bioprocessing, where experimental setups are resource-intensive, and acquiring labelled data is often costly and time-consuming. Future research should explore adaptive active learning strategies that dynamically adjust selection criteria based on evolving process conditions.

Furthermore, the use of ensemble learning and just-in-time learning (JITL) has emerged as a promising avenue for improving predictive performance in dynamic bioprocess environments. JITL, in particular, has demonstrated its effectiveness in Raman spectroscopy-based soft sensor development, allowing real-time model adaptation to new process conditions. Combining ensemble learning with JITL could further enhance model robustness by dynamically selecting and weighting the most reliable predictors.

Due to a lack of applications in upstream bioprocessing, Federated learning (FL) is an additional ML technique that has not been discussed so far. However, it could potentially offer valuable solutions to small data challenges as its success has been demonstrated in other industries. FL has been widely applied in various domains where data privacy and governance concerns limit data sharing \citep{po20, elso22, puch21, citr20, riha20}. For example, in the healthcare industry, FL has enabled collaborative model training across institutions without exposing sensitive patient data \citep{daro21}. Thus, for upstream bioprocessing, FL could potentially facilitate knowledge sharing across multiple biopharmaceutical companies, research labs, and production sites, allowing models to be trained on diverse datasets while maintaining data privacy. This approach has the potential to address the data silos across organisations and hence improve process optimization, yield prediction, and overall industry-wide bioprocess efficiency. 

While ML methods for small data are rapidly advancing, challenges remain in achieving regulatory acceptance and industry-wide adoption. Model interpretability, reproducibility, and uncertainty quantification must be prioritised to gain trust among bioprocess engineers and regulatory bodies. Explainable AI techniques, combined with Bayesian uncertainty estimation, could help bridge this gap by providing confidence intervals around predictions and justifying model decisions. Thus, although ML methods for small data play a transformative role in upstream bioprocessing, the success of these approaches hinges on their integration with domain knowledge, model retraining and adaptation strategies, and explainability. Future research should focus on refining these techniques and ensuring their seamless adoption in real-world biomanufacturing workflows.

\section{Conclusion}
\label{sec:conclusion}
This review comprehensively examined the key idea and major branches of various ML methods that can effectively address the challenges of building ML models using small data, along with discussions on existing research that applied these methods to various bioprocess development applications. To provide a holistic view of how small data problems are handled throughout the ML workflow, we proposed a taxonomy of methods for dealing with small data that is based on the steps of typical ML workflows. This can guide the design of the ML model development process of an ablation study or ML project where multiple methods need to be applied together or in sequence. Adopting these methods can significantly enhance the ability to gain valuable insights from limited data and allow robust and accurate ML models to be built, which can help optimise various upstream bioprocess components for improving the quality and productivity of the final products, such as cell culture yield, product purity, and process efficiency. However, as with any ML application, care must be taken in selecting the appropriate methods for each specific case, considering the nature of the data, the context, and the specific objectives. Future research in this field can continue exploring, experimenting, and refining these methods with real bioprocess data and applying them to solve practical issues in the upstream bioprocess.

\section*{CRediT authorship contribution statement}
\textbf{Johnny Peng:} Conceptualisation, Methodology, Investigation, Validation, Visualisation, Writing – original draft. \textbf{Thanh Tung Khuat:} Conceptualisation, Methodology, Investigation, Validation, Writing – review \& editing. \textbf{Katarzyna Musial:} Conceptualisation, Investigation, Validation, Supervision, Writing – review \& editing. \textbf{Bogdan Gabrys:} Conceptualisation, Methodology, Investigation, Validation, Project administration, Funding acquisition, Writing – review \& editing. 

\section*{Declaration of Competing Interest}
All authors declare no competing interests, including no known competing financial interests or personal relationships that could have appeared to influence the work reported in this paper.

\section*{Acknowledgements}
This research was supported under the Australian Research Council's Industrial Transformation Research Program (ITRP) funding scheme (project number IH210100051). The ARC Digital Bioprocess Development Hub is a collaboration between The University of Melbourne, University of Technology Sydney, RMIT University, CSL Innovation Pty Ltd, Cytiva (Global Life Science Solutions Australia Pty Ltd) and Patheon Biologics Australia Pty Ltd.


\bibliography{ref}

\begin{thebibliography}{250}
\expandafter\ifx\csname natexlab\endcsname\relax\def\natexlab#1{#1}\fi
\providecommand{\url}[1]{\texttt{#1}}
\providecommand{\href}[2]{#2}
\providecommand{\path}[1]{#1}
\providecommand{\DOIprefix}{doi:}
\providecommand{\ArXivprefix}{arXiv:}
\providecommand{\URLprefix}{URL: }
\providecommand{\Pubmedprefix}{pmid:}
\providecommand{\doi}[1]{\href{http://dx.doi.org/#1}{\path{#1}}}
\providecommand{\Pubmed}[1]{\href{pmid:#1}{\path{#1}}}
\providecommand{\bibinfo}[2]{#2}
\ifx\xfnm\relax \def\xfnm[#1]{\unskip,\space#1}\fi
\bibitem[{Abaineh et~al.(2007)Abaineh, Nayak, Ranjan \& Gomes}]{abna07}
\bibinfo{author}{Abaineh, B.}, \bibinfo{author}{Nayak, R.}, \bibinfo{author}{Ranjan, A.~P.}, \& \bibinfo{author}{Gomes, J.} (\bibinfo{year}{2007}).
\newblock \bibinfo{title}{On-line control of a fed-batch fermentation by using som based multiple local linear models}.
\newblock In {\it \bibinfo{booktitle}{Proceedings of the 10th International Conference on Engineering Applications of Neural Networks}\/} (pp. \bibinfo{pages}{368--375}).
\bibitem[{Agharafeie et~al.(2023)Agharafeie, Ramos, Mendes \& Oliveira}]{agra23}
\bibinfo{author}{Agharafeie, R.}, \bibinfo{author}{Ramos, J. R.~C.}, \bibinfo{author}{Mendes, J.~M.}, \& \bibinfo{author}{Oliveira, R.} (\bibinfo{year}{2023}).
\newblock \bibinfo{title}{From shallow to deep bioprocess hybrid modeling: Advances and future perspectives}.
\newblock {\it \bibinfo{journal}{Fermentation}\/},  {\it \bibinfo{volume}{9}\/}, \bibinfo{pages}{922}.
\bibitem[{Al-Digeil et~al.(2022)Al-Digeil, Grinberg, Daniele, Dezfouli, Schmid, Cheben, Janz \& Xu}]{algr22}
\bibinfo{author}{Al-Digeil, M.}, \bibinfo{author}{Grinberg, Y.}, \bibinfo{author}{Daniele, M.}, \bibinfo{author}{Dezfouli, M.}, \bibinfo{author}{Schmid, J.}, \bibinfo{author}{Cheben, P.}, \bibinfo{author}{Janz, S.}, \& \bibinfo{author}{Xu, D.} (\bibinfo{year}{2022}).
\newblock \bibinfo{title}{Pca-boosted autoencoders for nonlinear dimensionality reduction in low data regimes}.
\newblock {\it \bibinfo{journal}{arXiv preprint arXiv:2205.11673}\/}, .
\bibitem[{Al-Rawi \& Karajeh(2007)}]{alka07}
\bibinfo{author}{Al-Rawi, M.}, \& \bibinfo{author}{Karajeh, H.} (\bibinfo{year}{2007}).
\newblock \bibinfo{title}{Genetic algorithm matched filter optimization for automated detection of blood vessels from digital retinal images}.
\newblock {\it \bibinfo{journal}{Comput Methods Programs Biomed}\/},  {\it \bibinfo{volume}{87}\/}, \bibinfo{pages}{248--253}.
\bibitem[{Alexander et~al.(2020)Alexander, Campani, Dinh \& Lima}]{alca20}
\bibinfo{author}{Alexander, R.}, \bibinfo{author}{Campani, G.}, \bibinfo{author}{Dinh, S.}, \& \bibinfo{author}{Lima, F.~V.} (\bibinfo{year}{2020}).
\newblock \bibinfo{title}{Challenges and opportunities on nonlinear state estimation of chemical and biochemical processes}.
\newblock {\it \bibinfo{journal}{Processes}\/},  {\it \bibinfo{volume}{8}\/}, \bibinfo{pages}{1 – 27}.
\bibitem[{Ali et~al.(2020)Ali, Budka \& Gabrys}]{albu20}
\bibinfo{author}{Ali, A.~R.}, \bibinfo{author}{Budka, M.}, \& \bibinfo{author}{Gabrys, B.} (\bibinfo{year}{2020}).
\newblock \bibinfo{title}{A review of meta-level learning in the context of multi-component, multi-level evolving prediction systems}.
\newblock {\it \bibinfo{journal}{CoRR}\/},  {\it \bibinfo{volume}{abs/2007.10818}\/}.
\bibitem[{Alinaghi et~al.(2022)Alinaghi, Surowiec, Scholze, McCready, Zehe, Johansson, Trygg \& Cloarec}]{alsu22}
\bibinfo{author}{Alinaghi, M.}, \bibinfo{author}{Surowiec, I.}, \bibinfo{author}{Scholze, S.}, \bibinfo{author}{McCready, C.}, \bibinfo{author}{Zehe, C.}, \bibinfo{author}{Johansson, E.}, \bibinfo{author}{Trygg, J.}, \& \bibinfo{author}{Cloarec, O.} (\bibinfo{year}{2022}).
\newblock \bibinfo{title}{Hierarchical time-series analysis of dynamic bioprocess systems}.
\newblock {\it \bibinfo{journal}{Biotechnology Journal}\/},  {\it \bibinfo{volume}{17}\/}.
\bibitem[{Alwosheel et~al.(2018)Alwosheel, {van Cranenburgh} \& Chorus}]{alcr18}
\bibinfo{author}{Alwosheel, A.}, \bibinfo{author}{{van Cranenburgh}, S.}, \& \bibinfo{author}{Chorus, C.~G.} (\bibinfo{year}{2018}).
\newblock \bibinfo{title}{Is your dataset big enough? sample size requirements when using artificial neural networks for discrete choice analysis}.
\newblock {\it \bibinfo{journal}{Journal of Choice Modelling}\/},  {\it \bibinfo{volume}{28}\/}, \bibinfo{pages}{167--182}.
\bibitem[{Ameur et~al.(2023)Ameur, Njah \& Jamoussi}]{amnj23}
\bibinfo{author}{Ameur, H.}, \bibinfo{author}{Njah, H.}, \& \bibinfo{author}{Jamoussi, S.} (\bibinfo{year}{2023}).
\newblock \bibinfo{title}{Merits of bayesian networks in overcoming small data challenges: a meta-model for handling missing data}.
\newblock {\it \bibinfo{journal}{International Journal of Machine Learning and Cybernetics}\/},  {\it \bibinfo{volume}{14}\/}, \bibinfo{pages}{229--251}.
\bibitem[{Amicarelli et~al.(2016)Amicarelli, Montoya \& Sciascio}]{ammo16}
\bibinfo{author}{Amicarelli, A.}, \bibinfo{author}{Montoya, L.~Q.}, \& \bibinfo{author}{Sciascio, F.~D.} (\bibinfo{year}{2016}).
\newblock \bibinfo{title}{Substrate feeding strategy integrated with a biomass bayesian estimator for a biotechnological process}.
\newblock {\it \bibinfo{journal}{International Journal of Chemical Reactor Engineering}\/},  {\it \bibinfo{volume}{14}\/}, \bibinfo{pages}{1187 – 1200}.
\bibitem[{Arslan et~al.(2019)Arslan, Guzel, Demirci \& Ozdemir}]{argu19}
\bibinfo{author}{Arslan, M.}, \bibinfo{author}{Guzel, M.}, \bibinfo{author}{Demirci, M.}, \& \bibinfo{author}{Ozdemir, S.} (\bibinfo{year}{2019}).
\newblock \bibinfo{title}{Smote and gaussian noise based sensor data augmentation}.
\newblock In {\it \bibinfo{booktitle}{2019 4th International Conference on Computer Science and Engineering (UBMK)}\/} (pp. \bibinfo{pages}{1--5}).
\bibitem[{Bader et~al.(2023)Bader, Narayanan, Arosio \& Leroux}]{bana23}
\bibinfo{author}{Bader, J.}, \bibinfo{author}{Narayanan, H.}, \bibinfo{author}{Arosio, P.}, \& \bibinfo{author}{Leroux, J.-C.} (\bibinfo{year}{2023}).
\newblock \bibinfo{title}{Improving extracellular vesicles production through a bayesian optimization-based experimental design}.
\newblock {\it \bibinfo{journal}{European Journal of Pharmaceutics and Biopharmaceutics}\/},  {\it \bibinfo{volume}{182}\/}, \bibinfo{pages}{103 – 114}.
\newblock \bibinfo{note}{; All Open Access, Green Open Access, Hybrid Gold Open Access}.
\bibitem[{Bakirov et~al.(2015)Bakirov, Gabrys \& Fay}]{baga15}
\bibinfo{author}{Bakirov, R.}, \bibinfo{author}{Gabrys, B.}, \& \bibinfo{author}{Fay, D.} (\bibinfo{year}{2015}).
\newblock \bibinfo{title}{On sequences of different adaptive mechanisms in non-stationary regression problems}.
\newblock In {\it \bibinfo{booktitle}{2015 International Joint Conference on Neural Networks ({IJCNN})}\/}.
\newblock \bibinfo{publisher}{{IEEE}}.
\bibitem[{Bakirov et~al.(2017)Bakirov, Gabrys \& Fay}]{baga17}
\bibinfo{author}{Bakirov, R.}, \bibinfo{author}{Gabrys, B.}, \& \bibinfo{author}{Fay, D.} (\bibinfo{year}{2017}).
\newblock \bibinfo{title}{Multiple adaptive mechanisms for data-driven soft sensors}.
\newblock {\it \bibinfo{journal}{Computers {\&} Chemical Engineering}\/},  {\it \bibinfo{volume}{96}\/}, \bibinfo{pages}{42--54}.
\bibitem[{Bakirov et~al.(2018)Bakirov, Gabrys \& Fay}]{baga18}
\bibinfo{author}{Bakirov, R.}, \bibinfo{author}{Gabrys, B.}, \& \bibinfo{author}{Fay, D.} (\bibinfo{year}{2018}).
\newblock \bibinfo{title}{Generic adaptation strategies for automated machine learning}.
\newblock {\it \bibinfo{journal}{ArXiv}\/}, .
\bibitem[{Banner et~al.(2021)Banner, Alosert, Spencer, Cheeks, Farid, Thomas \& Goldrick}]{baal21}
\bibinfo{author}{Banner, M.}, \bibinfo{author}{Alosert, H.}, \bibinfo{author}{Spencer, C.}, \bibinfo{author}{Cheeks, M.}, \bibinfo{author}{Farid, S.~S.}, \bibinfo{author}{Thomas, M.}, \& \bibinfo{author}{Goldrick, S.} (\bibinfo{year}{2021}).
\newblock \bibinfo{title}{A decade in review: use of data analytics within the biopharmaceutical sector}.
\newblock {\it \bibinfo{journal}{Current Opinion in Chemical Engineering}\/},  {\it \bibinfo{volume}{34}\/}, \bibinfo{pages}{100758}.
\bibitem[{Barberi et~al.(2022)Barberi, Benedetti, Diaz-Fernandez, Sévin, Vappiani, Finka, Bezzo, Barolo \& Facco}]{babe22}
\bibinfo{author}{Barberi, G.}, \bibinfo{author}{Benedetti, A.}, \bibinfo{author}{Diaz-Fernandez, P.}, \bibinfo{author}{Sévin, D.}, \bibinfo{author}{Vappiani, J.}, \bibinfo{author}{Finka, G.}, \bibinfo{author}{Bezzo, F.}, \bibinfo{author}{Barolo, M.}, \& \bibinfo{author}{Facco, P.} (\bibinfo{year}{2022}).
\newblock \bibinfo{title}{Integrating metabolome dynamics and process data to guide cell line selection in biopharmaceutical process development}.
\newblock {\it \bibinfo{journal}{Metabolic Engineering}\/},  {\it \bibinfo{volume}{72}\/}, \bibinfo{pages}{353--364}.
\bibitem[{Baruch et~al.(2009)Baruch, Mariaca-Gaspar \& Barrera-Cortes}]{bama09a}
\bibinfo{author}{Baruch, I.}, \bibinfo{author}{Mariaca-Gaspar, C.-R.}, \& \bibinfo{author}{Barrera-Cortes, J.} (\bibinfo{year}{2009}).
\newblock \bibinfo{title}{Direct adaptive soft computing neural control of a continuous bioprocess via second order learning}.
\newblock {\it \bibinfo{journal}{Lecture Notes in Computer Science (including subseries Lecture Notes in Artificial Intelligence and Lecture Notes in Bioinformatics)}\/},  {\it \bibinfo{volume}{5845 LNAI}\/}, \bibinfo{pages}{500 – 511}.
\bibitem[{Baruch et~al.(2010)Baruch, Mariaca-Gaspar, Barrera-Cortes \& Castillo}]{bama10}
\bibinfo{author}{Baruch, I.}, \bibinfo{author}{Mariaca-Gaspar, C.-R.}, \bibinfo{author}{Barrera-Cortes, J.}, \& \bibinfo{author}{Castillo, O.} (\bibinfo{year}{2010}).
\newblock \bibinfo{title}{Direct and indirect neural identification and control of a continuous bioprocess via marquardt learning}.
\newblock {\it \bibinfo{journal}{Studies in Computational Intelligence}\/},  {\it \bibinfo{volume}{318}\/}, \bibinfo{pages}{81 – 102}.
\bibitem[{Baruch et~al.(2007)Baruch, Escalante, Mariaca-Gaspar \& Barrera-Cortes}]{baes07}
\bibinfo{author}{Baruch, I.~S.}, \bibinfo{author}{Escalante, S.~F.}, \bibinfo{author}{Mariaca-Gaspar, C.~R.}, \& \bibinfo{author}{Barrera-Cortes, J.} (\bibinfo{year}{2007}).
\newblock \bibinfo{title}{Recurrent neural control of wastewater treatment bioprocess via marquardt learning}.
\newblock {\it \bibinfo{journal}{IFAC Proceedings Volumes}\/},  {\it \bibinfo{volume}{40}\/}, \bibinfo{pages}{289--294}.
\bibitem[{Baruch \& Mariaca-Gaspar(2009)}]{bama09}
\bibinfo{author}{Baruch, I.~S.}, \& \bibinfo{author}{Mariaca-Gaspar, C.~R.} (\bibinfo{year}{2009}).
\newblock \bibinfo{title}{A levenberg-marquardt learning applied for recurrent neural identification and control of a wastewater treatment bioprocess}.
\newblock {\it \bibinfo{journal}{International Journal of Intelligent Systems}\/},  {\it \bibinfo{volume}{24}\/}, \bibinfo{pages}{1094 – 1114}.
\bibitem[{Bayer et~al.(2021)Bayer, Duerkop, Striedner \& Sissolak}]{badu21}
\bibinfo{author}{Bayer, B.}, \bibinfo{author}{Duerkop, M.}, \bibinfo{author}{Striedner, G.}, \& \bibinfo{author}{Sissolak, B.} (\bibinfo{year}{2021}).
\newblock \bibinfo{title}{Model transferability and reduced experimental burden in cell culture process development facilitated by hybrid modeling and intensified design of experiments}.
\newblock {\it \bibinfo{journal}{Frontiers in bioengineering and biotechnology}\/},  {\it \bibinfo{volume}{9}\/}, \bibinfo{pages}{740215--740215}.
\newblock \bibinfo{note}{Krist V. Gernaey, Technical University of Denmark, Denmark}.
\bibitem[{Belkin \& Niyogi(2001)}]{beni01}
\bibinfo{author}{Belkin, M.}, \& \bibinfo{author}{Niyogi, P.} (\bibinfo{year}{2001}).
\newblock \bibinfo{title}{Laplacian eigenmaps and spectral techniques for embedding and clustering}.
\newblock In \bibinfo{editor}{T.~Dietterich}, \bibinfo{editor}{S.~Becker}, \& \bibinfo{editor}{Z.~Ghahramani} (Eds.), {\it \bibinfo{booktitle}{Advances in Neural Information Processing Systems}\/}.
\newblock \bibinfo{publisher}{MIT Press} volume~\bibinfo{volume}{14}.
\bibitem[{Bergmeir et~al.(2016)Bergmeir, Hyndman \& Benítez}]{behy16}
\bibinfo{author}{Bergmeir, C.}, \bibinfo{author}{Hyndman, R.~J.}, \& \bibinfo{author}{Benítez, J.~M.} (\bibinfo{year}{2016}).
\newblock \bibinfo{title}{Bagging exponential smoothing methods using stl decomposition and box–cox transformation}.
\newblock {\it \bibinfo{journal}{International Journal of Forecasting}\/},  {\it \bibinfo{volume}{32}\/}, \bibinfo{pages}{303--312}.
\bibitem[{Bernard et~al.(2006)Bernard, Chachuat, Helias \& Rodriguez}]{bech06}
\bibinfo{author}{Bernard, O.}, \bibinfo{author}{Chachuat, B.}, \bibinfo{author}{Helias, A.}, \& \bibinfo{author}{Rodriguez, J.} (\bibinfo{year}{2006}).
\newblock \bibinfo{title}{Can we assess the model complexity for a bioprocess: Theory and example of the anaerobic digestion process}.
\newblock {\it \bibinfo{journal}{Water Science and Technology}\/},  {\it \bibinfo{volume}{53}\/}, \bibinfo{pages}{85 – 92}.
\bibitem[{Borah et~al.(2024)Borah, Das, Seth, Mallick, Rahaman \& Mallik}]{boda24}
\bibinfo{author}{Borah, K.}, \bibinfo{author}{Das, H.~S.}, \bibinfo{author}{Seth, S.}, \bibinfo{author}{Mallick, K.}, \bibinfo{author}{Rahaman, Z.}, \& \bibinfo{author}{Mallik, S.} (\bibinfo{year}{2024}).
\newblock \bibinfo{title}{A review on advancements in feature selection and feature extraction for high-dimensional ngs data analysis}.
\newblock {\it \bibinfo{journal}{Functional {\&} Integrative Genomics}\/},  {\it \bibinfo{volume}{24}\/}, \bibinfo{pages}{139}.
\bibitem[{Borisyak et~al.(2024)Borisyak, Born, Neubauer \& Cruz-Bournazou}]{bobo24}
\bibinfo{author}{Borisyak, M.}, \bibinfo{author}{Born, S.}, \bibinfo{author}{Neubauer, P.}, \& \bibinfo{author}{Cruz-Bournazou, M.~N.} (\bibinfo{year}{2024}).
\newblock \bibinfo{title}{Deep learning for fast inference of mechanistic models’ parameters}.
\newblock In {\it \bibinfo{booktitle}{Computer Aided Chemical Engineering}\/} (pp. \bibinfo{pages}{3043--3048}).
\newblock \bibinfo{publisher}{Elsevier} volume~\bibinfo{volume}{53}.
\bibitem[{Borkowski et~al.(2020)Borkowski, Koch, Zettor, Pandi, Batista, Soudier \& Faulon}]{boko20}
\bibinfo{author}{Borkowski, O.}, \bibinfo{author}{Koch, M.}, \bibinfo{author}{Zettor, A.}, \bibinfo{author}{Pandi, A.}, \bibinfo{author}{Batista, A.~C.}, \bibinfo{author}{Soudier, P.}, \& \bibinfo{author}{Faulon, J.-L.} (\bibinfo{year}{2020}).
\newblock \bibinfo{title}{Large scale active-learning-guided exploration for in vitro protein production optimization}.
\newblock {\it \bibinfo{journal}{Nature Communications}\/},  {\it \bibinfo{volume}{11}\/}, \bibinfo{pages}{1872}.
\bibitem[{Bosch et~al.(2022)Bosch, Holmstr{\"o}m~Olsson \& Crnkovic}]{boho22}
\bibinfo{author}{Bosch, J.}, \bibinfo{author}{Holmstr{\"o}m~Olsson, H.}, \& \bibinfo{author}{Crnkovic, I.} (\bibinfo{year}{2022}).
\newblock \bibinfo{title}{Chapter 13 engineering ai systems}.
\newblock In \bibinfo{editor}{J.~Bosch}, \bibinfo{editor}{J.~Carlson}, \bibinfo{editor}{H.~Holmstr{\"o}m~Olsson}, \bibinfo{editor}{K.~Sandahl}, \& \bibinfo{editor}{M.~Staron} (Eds.), {\it \bibinfo{booktitle}{Accelerating Digital Transformation: 10 Years of Software Center}\/} (pp. \bibinfo{pages}{407--425}).
\newblock \bibinfo{address}{Cham}: \bibinfo{publisher}{Springer International Publishing}.
\bibitem[{Botton et~al.(2022)Botton, Barberi \& Facco}]{boba22}
\bibinfo{author}{Botton, A.}, \bibinfo{author}{Barberi, G.}, \& \bibinfo{author}{Facco, P.} (\bibinfo{year}{2022}).
\newblock \bibinfo{title}{Data augmentation to support biopharmaceutical process development through digital models—a proof of concept}.
\newblock {\it \bibinfo{journal}{Processes}\/},  {\it \bibinfo{volume}{10}\/}.
\bibitem[{Bradley et~al.(2022)Bradley, Kim, Kilwein, Blakely, Eydenberg, Jalvin, Laird \& Boukouvala}]{brki22}
\bibinfo{author}{Bradley, W.}, \bibinfo{author}{Kim, J.}, \bibinfo{author}{Kilwein, Z.}, \bibinfo{author}{Blakely, L.}, \bibinfo{author}{Eydenberg, M.}, \bibinfo{author}{Jalvin, J.}, \bibinfo{author}{Laird, C.}, \& \bibinfo{author}{Boukouvala, F.} (\bibinfo{year}{2022}).
\newblock \bibinfo{title}{Perspectives on the integration between first-principles and data-driven modeling}.
\newblock {\it \bibinfo{journal}{Computers \& Chemical Engineering}\/},  {\it \bibinfo{volume}{166}\/}, \bibinfo{pages}{107898}.
\bibitem[{Brunner et~al.(2020)Brunner, Klöckner, Kerpes, Geier \& Becker}]{brkl20}
\bibinfo{author}{Brunner, V.}, \bibinfo{author}{Klöckner, L.}, \bibinfo{author}{Kerpes, R.}, \bibinfo{author}{Geier, D.~U.}, \& \bibinfo{author}{Becker, T.} (\bibinfo{year}{2020}).
\newblock \bibinfo{title}{Online sensor validation in sensor networks for bioprocess monitoring using swarm intelligence}.
\newblock {\it \bibinfo{journal}{Analytical and Bioanalytical Chemistry}\/},  {\it \bibinfo{volume}{412}\/}, \bibinfo{pages}{2165 – 2175}.
\bibitem[{Cao et~al.(2022)Cao, Bu, Huang, Zhang, Tsang, Ong \& Kwok}]{cabu22}
\bibinfo{author}{Cao, X.}, \bibinfo{author}{Bu, W.}, \bibinfo{author}{Huang, S.}, \bibinfo{author}{Zhang, M.}, \bibinfo{author}{Tsang, I.~W.}, \bibinfo{author}{Ong, Y.~S.}, \& \bibinfo{author}{Kwok, J.~T.} (\bibinfo{year}{2022}).
\newblock \bibinfo{title}{A survey of learning on small data: generalization, optimization, and challenge}.
\newblock {\it \bibinfo{journal}{arXiv preprint arXiv:2207.14443}\/}, .
\bibitem[{Chai et~al.(2022)Chai, Zhao, Huang \& Chen}]{chzh22}
\bibinfo{author}{Chai, Z.}, \bibinfo{author}{Zhao, C.}, \bibinfo{author}{Huang, B.}, \& \bibinfo{author}{Chen, H.} (\bibinfo{year}{2022}).
\newblock \bibinfo{title}{A deep probabilistic transfer learning framework for soft sensor modeling with missing data}.
\newblock {\it \bibinfo{journal}{IEEE Transactions on Neural Networks and Learning Systems}\/},  {\it \bibinfo{volume}{33}\/}, \bibinfo{pages}{7598 – 7609}.
\bibitem[{Chapelle et~al.(2010)Chapelle, Scholkopf \& Zien}]{chsc10}
\bibinfo{author}{Chapelle, O.}, \bibinfo{author}{Scholkopf, B.}, \& \bibinfo{author}{Zien, A.} (\bibinfo{year}{2010}).
\newblock {\it \bibinfo{title}{Semi-supervised learning}\/}.
\newblock Adaptive computation and machine learning series.
\newblock \bibinfo{address}{Cambridge, Massachusetts}: \bibinfo{publisher}{MIT Press}.
\bibitem[{Chawla et~al.(2002)Chawla, Bowyer, Hall \& Kegelmeyer}]{chbo02}
\bibinfo{author}{Chawla, N.~V.}, \bibinfo{author}{Bowyer, K.~W.}, \bibinfo{author}{Hall, L.~O.}, \& \bibinfo{author}{Kegelmeyer, W.~P.} (\bibinfo{year}{2002}).
\newblock \bibinfo{title}{Smote: synthetic minority over-sampling technique}.
\newblock {\it \bibinfo{journal}{Journal of artificial intelligence research}\/},  {\it \bibinfo{volume}{16}\/}, \bibinfo{pages}{321--357}.
\bibitem[{Che et~al.(2017)Che, Yang, Li, Bai, Zhang \& Deng}]{chya17}
\bibinfo{author}{Che, J.}, \bibinfo{author}{Yang, Y.}, \bibinfo{author}{Li, L.}, \bibinfo{author}{Bai, X.}, \bibinfo{author}{Zhang, S.}, \& \bibinfo{author}{Deng, C.} (\bibinfo{year}{2017}).
\newblock \bibinfo{title}{Maximum relevance minimum common redundancy feature selection for nonlinear data}.
\newblock {\it \bibinfo{journal}{Information sciences}\/},  {\it \bibinfo{volume}{409-410}\/}, \bibinfo{pages}{68--86}.
\bibitem[{Chen et~al.(2022)Chen, Jiao \& Li}]{chji22}
\bibinfo{author}{Chen, H.}, \bibinfo{author}{Jiao, L.}, \& \bibinfo{author}{Li, S.} (\bibinfo{year}{2022}).
\newblock \bibinfo{title}{A soft sensor regression model for complex chemical process based on generative adversarial nets and vine copula}.
\newblock {\it \bibinfo{journal}{Journal of the Taiwan Institute of Chemical Engineers}\/},  {\it \bibinfo{volume}{138}\/}.
\bibitem[{Chen \& Liu(2013)}]{chli13}
\bibinfo{author}{Chen, K.}, \& \bibinfo{author}{Liu, Y.} (\bibinfo{year}{2013}).
\newblock \bibinfo{title}{Adaptive control of continuous time-varying bioprocesses using recursive kernel learning controller with polynomial form}.
\newblock {\it \bibinfo{journal}{IFAC Proceedings Volumes}\/},  {\it \bibinfo{volume}{46}\/}, \bibinfo{pages}{359--364}.
\bibitem[{Chiu \& Du(2024)}]{chdu24}
\bibinfo{author}{Chiu, K.-C.}, \& \bibinfo{author}{Du, D.} (\bibinfo{year}{2024}).
\newblock \bibinfo{title}{A neural ordinary differential equation model for predicting the growth of chinese hamster ovary cell in a bioreactor system}.
\newblock {\it \bibinfo{journal}{Biotechnology and bioprocess engineering}\/}, .
\bibitem[{Chuan et~al.(2014)Chuan, Wibowo, Lua \& Middelberg}]{chwi14}
\bibinfo{author}{Chuan, Y.~P.}, \bibinfo{author}{Wibowo, N.}, \bibinfo{author}{Lua, L.~H.}, \& \bibinfo{author}{Middelberg, A.~P.} (\bibinfo{year}{2014}).
\newblock \bibinfo{title}{The economics of virus-like particle and capsomere vaccines}.
\newblock {\it \bibinfo{journal}{Biochemical Engineering Journal}\/},  {\it \bibinfo{volume}{90}\/}, \bibinfo{pages}{255 – 263}.
\bibitem[{Cioffi et~al.(2020)Cioffi, Travaglioni, Piscitelli, Petrillo \& De~Felice}]{citr20}
\bibinfo{author}{Cioffi, R.}, \bibinfo{author}{Travaglioni, M.}, \bibinfo{author}{Piscitelli, G.}, \bibinfo{author}{Petrillo, A.}, \& \bibinfo{author}{De~Felice, F.} (\bibinfo{year}{2020}).
\newblock \bibinfo{title}{Artificial intelligence and machine learning applications in smart production: Progress, trends, and directions}.
\newblock {\it \bibinfo{journal}{Sustainability}\/},  {\it \bibinfo{volume}{12}\/}.
\bibitem[{Cleveland et~al.(1990)Cleveland, Cleveland \& Terpenning}]{clcl90}
\bibinfo{author}{Cleveland, R.~B.}, \bibinfo{author}{Cleveland, W.~S.}, \& \bibinfo{author}{Terpenning, I.} (\bibinfo{year}{1990}).
\newblock \bibinfo{title}{Stl: A seasonal-trend decomposition procedure based on loess}.
\newblock {\it \bibinfo{journal}{Journal of Official Statistics}\/},  {\it \bibinfo{volume}{6}\/}, \bibinfo{pages}{3}.
\bibitem[{Comon(1994)}]{co94}
\bibinfo{author}{Comon, P.} (\bibinfo{year}{1994}).
\newblock \bibinfo{title}{Independent component analysis, a new concept?}
\newblock {\it \bibinfo{journal}{Signal Processing}\/},  {\it \bibinfo{volume}{36}\/}, \bibinfo{pages}{287--314}.
\bibitem[{Dai et~al.(2023)Dai, Yang, Liu \& Yao}]{daya23}
\bibinfo{author}{Dai, Y.}, \bibinfo{author}{Yang, C.}, \bibinfo{author}{Liu, Y.}, \& \bibinfo{author}{Yao, Y.} (\bibinfo{year}{2023}).
\newblock \bibinfo{title}{Latent-enhanced variational adversarial active learning assisted soft sensor}.
\newblock {\it \bibinfo{journal}{IEEE Sensors Journal}\/},  {\it \bibinfo{volume}{23}\/}, \bibinfo{pages}{15762--15772}.
\bibitem[{Dai et~al.(2022)Dai, Zhang, Yao \& Liu}]{dazh22}
\bibinfo{author}{Dai, Y.}, \bibinfo{author}{Zhang, Y.}, \bibinfo{author}{Yao, Y.}, \& \bibinfo{author}{Liu, Y.} (\bibinfo{year}{2022}).
\newblock \bibinfo{title}{Variational adversarial active learning assisted process soft sensor method}.
\newblock In {\it \bibinfo{booktitle}{Proceedings of the 4th International Conference on Industrial Artificial Intelligence (IAI)}\/} (pp. \bibinfo{pages}{1--5}).
\newblock \bibinfo{organization}{IEEE}.
\bibitem[{Dayal \& MacGregor(1997)}]{dama97}
\bibinfo{author}{Dayal, B.~S.}, \& \bibinfo{author}{MacGregor, J.~F.} (\bibinfo{year}{1997}).
\newblock \bibinfo{title}{Improved pls algorithms}.
\newblock {\it \bibinfo{journal}{Journal of chemometrics}\/},  {\it \bibinfo{volume}{11}\/}, \bibinfo{pages}{73--85}.
\newblock \bibinfo{note}{ArticleID:CEM435}.
\bibitem[{Dayan et~al.(2021)Dayan, Roth, Zhong, Harouni, Gentili, Abidin, Liu \& et~al.}]{daro21}
\bibinfo{author}{Dayan, I.}, \bibinfo{author}{Roth, H.~R.}, \bibinfo{author}{Zhong, A.}, \bibinfo{author}{Harouni, A.}, \bibinfo{author}{Gentili, A.}, \bibinfo{author}{Abidin, A.~Z.}, \bibinfo{author}{Liu, A.}, \& \bibinfo{author}{et~al.} (\bibinfo{year}{2021}).
\newblock \bibinfo{title}{Federated learning for predicting clinical outcomes in patients with covid-19}.
\newblock {\it \bibinfo{journal}{Nature Medicine}\/},  {\it \bibinfo{volume}{27}\/}, \bibinfo{pages}{1735--1743}.
\bibitem[{Delgado-Escaño et~al.(2019)Delgado-Escaño, Castro, Cózar, Marín-Jiménez \& Guil}]{deca19}
\bibinfo{author}{Delgado-Escaño, R.}, \bibinfo{author}{Castro, F.~M.}, \bibinfo{author}{Cózar, J.~R.}, \bibinfo{author}{Marín-Jiménez, M.~J.}, \& \bibinfo{author}{Guil, N.} (\bibinfo{year}{2019}).
\newblock \bibinfo{title}{An end-to-end multi-task and fusion cnn for inertial-based gait recognition}.
\newblock {\it \bibinfo{journal}{IEEE Access}\/},  {\it \bibinfo{volume}{7}\/}, \bibinfo{pages}{1897--1908}.
\bibitem[{Dong et~al.(2020)Dong, Yu, Cao, Shi \& Ma}]{doyu20}
\bibinfo{author}{Dong, X.}, \bibinfo{author}{Yu, Z.}, \bibinfo{author}{Cao, W.}, \bibinfo{author}{Shi, Y.}, \& \bibinfo{author}{Ma, Q.} (\bibinfo{year}{2020}).
\newblock \bibinfo{title}{A survey on ensemble learning}.
\newblock {\it \bibinfo{journal}{Frontiers of Computer Science}\/},  {\it \bibinfo{volume}{14}\/}, \bibinfo{pages}{241--258}.
\bibitem[{Duong-Trung et~al.(2023)Duong-Trung, Born, Kim, Schermeyer, Paulick, Borisyak, Cruz-Bournazou, Werner, Scholz, Schmidt-Thieme, Neubauer \& Martinez}]{dubo23}
\bibinfo{author}{Duong-Trung, N.}, \bibinfo{author}{Born, S.}, \bibinfo{author}{Kim, J.~W.}, \bibinfo{author}{Schermeyer, M.-T.}, \bibinfo{author}{Paulick, K.}, \bibinfo{author}{Borisyak, M.}, \bibinfo{author}{Cruz-Bournazou, M.~N.}, \bibinfo{author}{Werner, T.}, \bibinfo{author}{Scholz, R.}, \bibinfo{author}{Schmidt-Thieme, L.}, \bibinfo{author}{Neubauer, P.}, \& \bibinfo{author}{Martinez, E.} (\bibinfo{year}{2023}).
\newblock \bibinfo{title}{When bioprocess engineering meets machine learning: A survey from the perspective of automated bioprocess development}.
\newblock {\it \bibinfo{journal}{Biochemical Engineering Journal}\/},  {\it \bibinfo{volume}{190}\/}, \bibinfo{pages}{108764}.
\bibitem[{von~den Eichen et~al.(2022)von~den Eichen, Osthege, Dölle, Bromig, Wiechert, Oldiges \& Weuster-Botz}]{voos22}
\bibinfo{author}{von~den Eichen, N.}, \bibinfo{author}{Osthege, M.}, \bibinfo{author}{Dölle, M.}, \bibinfo{author}{Bromig, L.}, \bibinfo{author}{Wiechert, W.}, \bibinfo{author}{Oldiges, M.}, \& \bibinfo{author}{Weuster-Botz, D.} (\bibinfo{year}{2022}).
\newblock \bibinfo{title}{Control of parallelized bioreactors ii: probabilistic quantification of carboxylic acid reductase activity for bioprocess optimization}.
\newblock {\it \bibinfo{journal}{Bioprocess and Biosystems Engineering}\/},  {\it \bibinfo{volume}{45}\/}, \bibinfo{pages}{1939 – 1954}.
\bibitem[{Elbir et~al.(2022)Elbir, Soner, Coleri, Gunduz \& Bennis}]{elso22}
\bibinfo{author}{Elbir, A.~M.}, \bibinfo{author}{Soner, B.}, \bibinfo{author}{Coleri, S.}, \bibinfo{author}{Gunduz, D.}, \& \bibinfo{author}{Bennis, M.} (\bibinfo{year}{2022}).
\newblock \bibinfo{title}{Federated learning in vehicular networks}.
\bibitem[{Eltoft(2002)}]{el02}
\bibinfo{author}{Eltoft, T.} (\bibinfo{year}{2002}).
\newblock \bibinfo{title}{Data augmentation using a combination of independent component analysis and non-linear time-series prediction}.
\newblock In {\it \bibinfo{booktitle}{Proceedings of the 2002 International Joint Conference on Neural Networks. IJCNN'02}\/} (pp. \bibinfo{pages}{448--453}).
\bibitem[{van Engelen \& Hoos(2020)}]{vaho20}
\bibinfo{author}{van Engelen, J.~E.}, \& \bibinfo{author}{Hoos, H.~H.} (\bibinfo{year}{2020}).
\newblock \bibinfo{title}{A survey on semi-supervised learning}.
\newblock {\it \bibinfo{journal}{Machine Learning}\/},  {\it \bibinfo{volume}{109}\/}, \bibinfo{pages}{373 – 440}.
\bibitem[{Esche et~al.(2022)Esche, Talis, Weigert, Brand~Rihm, You, Hoffmann \& Repke}]{esta22}
\bibinfo{author}{Esche, E.}, \bibinfo{author}{Talis, T.}, \bibinfo{author}{Weigert, J.}, \bibinfo{author}{Brand~Rihm, G.}, \bibinfo{author}{You, B.}, \bibinfo{author}{Hoffmann, C.}, \& \bibinfo{author}{Repke, J.-U.} (\bibinfo{year}{2022}).
\newblock \bibinfo{title}{Semi-supervised learning for data-driven soft-sensing of biological and chemical processes}.
\newblock {\it \bibinfo{journal}{Chemical Engineering Science}\/},  {\it \bibinfo{volume}{251}\/}, \bibinfo{pages}{117459}.
\bibitem[{Esmonde-White et~al.(2017)Esmonde-White, Cuellar, Uerpmann, Lenain \& Lewis}]{escu17}
\bibinfo{author}{Esmonde-White, K.~A.}, \bibinfo{author}{Cuellar, M.}, \bibinfo{author}{Uerpmann, C.}, \bibinfo{author}{Lenain, B.}, \& \bibinfo{author}{Lewis, I.~R.} (\bibinfo{year}{2017}).
\newblock \bibinfo{title}{Raman spectroscopy as a process analytical technology for pharmaceutical manufacturing and bioprocessing}.
\newblock {\it \bibinfo{journal}{Analytical and Bioanalytical Chemistry}\/},  {\it \bibinfo{volume}{409}\/}, \bibinfo{pages}{637--649}.
\bibitem[{Espinel-R{\'\i}os et~al.(2022)Espinel-R{\'\i}os, Morabito, Bettenbrock, Klamt \& Findeisen}]{esmo22}
\bibinfo{author}{Espinel-R{\'\i}os, S.}, \bibinfo{author}{Morabito, B.}, \bibinfo{author}{Bettenbrock, K.}, \bibinfo{author}{Klamt, S.}, \& \bibinfo{author}{Findeisen, R.} (\bibinfo{year}{2022}).
\newblock \bibinfo{title}{Soft sensor for monitoring dynamic changes in cell composition}.
\newblock {\it \bibinfo{journal}{IFAC-PapersOnLine}\/},  {\it \bibinfo{volume}{55}\/}, \bibinfo{pages}{98--103}.
\bibitem[{Farhan et~al.(2013)Farhan, Larjo, Yli-Harja \& Aho}]{fala13}
\bibinfo{author}{Farhan, M.}, \bibinfo{author}{Larjo, A.}, \bibinfo{author}{Yli-Harja, O.}, \& \bibinfo{author}{Aho, T.} (\bibinfo{year}{2013}).
\newblock \bibinfo{title}{Modeling bioprocess scale-up utilizing regularized linear and logistic regression}.
\newblock In {\it \bibinfo{booktitle}{2013 IEEE International Workshop on Machine Learning for Signal Processing (MLSP)}\/} (pp. \bibinfo{pages}{1--6}).
\newblock \bibinfo{organization}{IEEE}.
\bibitem[{Faulon \& Faure(2021)}]{fafa21}
\bibinfo{author}{Faulon, J.-L.}, \& \bibinfo{author}{Faure, L.} (\bibinfo{year}{2021}).
\newblock \bibinfo{title}{In silico, in vitro, and in vivo machine learning in synthetic biology and metabolic engineering}.
\newblock {\it \bibinfo{journal}{Current Opinion in Chemical Biology}\/},  {\it \bibinfo{volume}{65}\/}, \bibinfo{pages}{85 – 92}.
\bibitem[{Fawaz et~al.(2018)Fawaz, Forestier, Weber, Idoumghar \& Muller}]{fafo18}
\bibinfo{author}{Fawaz, H.~I.}, \bibinfo{author}{Forestier, G.}, \bibinfo{author}{Weber, J.}, \bibinfo{author}{Idoumghar, L.}, \& \bibinfo{author}{Muller, P.-A.} (\bibinfo{year}{2018}).
\newblock \bibinfo{title}{Data augmentation using synthetic data for time series classification with deep residual networks}.
\newblock In {\it \bibinfo{booktitle}{Proceedings of the 3rd ECML/PKDD Workshop on Advanced Analytics and Learning on Temporal Data}\/}.
\bibitem[{Feidl et~al.(2019)Feidl, Garbellini, Luna, Vogg, Souquet, Broly, Morbidelli \& Butt{\'e}}]{fega19}
\bibinfo{author}{Feidl, F.}, \bibinfo{author}{Garbellini, S.}, \bibinfo{author}{Luna, M.~F.}, \bibinfo{author}{Vogg, S.}, \bibinfo{author}{Souquet, J.}, \bibinfo{author}{Broly, H.}, \bibinfo{author}{Morbidelli, M.}, \& \bibinfo{author}{Butt{\'e}, A.} (\bibinfo{year}{2019}).
\newblock \bibinfo{title}{Combining mechanistic modeling and raman spectroscopy for monitoring antibody chromatographic purification}.
\newblock {\it \bibinfo{journal}{Processes}\/},  {\it \bibinfo{volume}{7}\/}, \bibinfo{pages}{683}.
\bibitem[{Feng \& Zhao(2020)}]{fezh20}
\bibinfo{author}{Feng, L.}, \& \bibinfo{author}{Zhao, C.} (\bibinfo{year}{2020}).
\newblock \bibinfo{title}{Adversarial sample based semi-supervised learning for industrial soft sensor}.
\newblock {\it \bibinfo{journal}{IFAC-PapersOnLine}\/},  {\it \bibinfo{volume}{53}\/}, \bibinfo{pages}{11644--11649}.
\bibitem[{Fernández et~al.(2021)Fernández, Pantano, Rodriguez \& Scaglia}]{fepa21}
\bibinfo{author}{Fernández, M.~C.}, \bibinfo{author}{Pantano, M.~N.}, \bibinfo{author}{Rodriguez, L.}, \& \bibinfo{author}{Scaglia, G.} (\bibinfo{year}{2021}).
\newblock \bibinfo{title}{State estimation and nonlinear tracking control simulation approach. application to a bioethanol production system}.
\newblock {\it \bibinfo{journal}{Bioprocess and Biosystems Engineering}\/},  {\it \bibinfo{volume}{44}\/}, \bibinfo{pages}{1755 – 1768}.
\bibitem[{Fields et~al.(2019)Fields, Hsieh \& Chenou}]{fihs19}
\bibinfo{author}{Fields, T.}, \bibinfo{author}{Hsieh, G.}, \& \bibinfo{author}{Chenou, J.} (\bibinfo{year}{2019}).
\newblock \bibinfo{title}{Mitigating drift in time series data with noise augmentation}.
\newblock In {\it \bibinfo{booktitle}{2019 International Conference on Computational Science and Computational Intelligence (CSCI)}\/} (pp. \bibinfo{pages}{227--230}).
\bibitem[{Fortescue et~al.(1981)Fortescue, Kershenbaum \& Ydstie}]{foke81}
\bibinfo{author}{Fortescue, T.}, \bibinfo{author}{Kershenbaum, L.}, \& \bibinfo{author}{Ydstie, B.} (\bibinfo{year}{1981}).
\newblock \bibinfo{title}{Implementation of self-tuning regulators with variable forgetting factors}.
\newblock {\it \bibinfo{journal}{Automatica}\/},  {\it \bibinfo{volume}{17}\/}, \bibinfo{pages}{831--835}.
\bibitem[{Fournier \& Aloise(2019)}]{foal19}
\bibinfo{author}{Fournier, Q.}, \& \bibinfo{author}{Aloise, D.} (\bibinfo{year}{2019}).
\newblock \bibinfo{title}{Empirical comparison between autoencoders and traditional dimensionality reduction methods}.
\newblock In {\it \bibinfo{booktitle}{Proceedings of the IEEE Second International Conference on Artificial Intelligence and Knowledge Engineering (AIKE)}\/} (pp. \bibinfo{pages}{211--214}).
\newblock \bibinfo{organization}{IEEE}.
\bibitem[{Fujiwara et~al.(2012)Fujiwara, Kano \& Hasebe}]{fuka12}
\bibinfo{author}{Fujiwara, K.}, \bibinfo{author}{Kano, M.}, \& \bibinfo{author}{Hasebe, S.} (\bibinfo{year}{2012}).
\newblock \bibinfo{title}{Development of correlation-based pattern recognition algorithm and adaptive soft-sensor design}.
\newblock {\it \bibinfo{journal}{Control Engineering Practice}\/},  {\it \bibinfo{volume}{20}\/}, \bibinfo{pages}{371--378}.
\bibitem[{Gabrys(2004)}]{ga04}
\bibinfo{author}{Gabrys, B.} (\bibinfo{year}{2004}).
\newblock \bibinfo{title}{Learning hybrid neuro-fuzzy classifier models from data: to combine or not to combine?}
\newblock {\it \bibinfo{journal}{Fuzzy Sets and Systems}\/},  {\it \bibinfo{volume}{147}\/}, \bibinfo{pages}{39--56}.
\bibitem[{Gabrys et~al.(2006)Gabrys, Baruque \& Corchado}]{gaba06}
\bibinfo{author}{Gabrys, B.}, \bibinfo{author}{Baruque, B.}, \& \bibinfo{author}{Corchado, E.} (\bibinfo{year}{2006}).
\newblock \bibinfo{title}{Outlier resistant pca ensembles}.
\newblock In \bibinfo{editor}{B.~Gabrys}, \bibinfo{editor}{R.~J. Howlett}, \& \bibinfo{editor}{L.~C. Jain} (Eds.), {\it \bibinfo{booktitle}{Knowledge-Based Intelligent Information and Engineering Systems}\/} (pp. \bibinfo{pages}{432--440}).
\newblock \bibinfo{address}{Berlin, Heidelberg}: \bibinfo{publisher}{Springer Berlin Heidelberg}.
\bibitem[{Gabrys \& Petrakieva(2004)}]{gape04}
\bibinfo{author}{Gabrys, B.}, \& \bibinfo{author}{Petrakieva, L.} (\bibinfo{year}{2004}).
\newblock \bibinfo{title}{Combining labelled and unlabelled data in the design of pattern classification systems}.
\newblock {\it \bibinfo{journal}{International Journal of Approximate Reasoning}\/},  {\it \bibinfo{volume}{35}\/}, \bibinfo{pages}{251--273}.
\newblock \bibinfo{note}{Integration of Methods and Hybrid Systems}.
\bibitem[{Gao et~al.(2021)Gao, Song, Wen, Wang, Sun \& Xu}]{gaso21}
\bibinfo{author}{Gao, J.}, \bibinfo{author}{Song, X.}, \bibinfo{author}{Wen, Q.}, \bibinfo{author}{Wang, P.}, \bibinfo{author}{Sun, L.}, \& \bibinfo{author}{Xu, H.} (\bibinfo{year}{2021}).
\newblock \bibinfo{title}{Robusttad: Robust time series anomaly detection via decomposition and convolutional neural networks}.
\bibitem[{Gautam et~al.(2015)Gautam, Vanga \& Ariese}]{gava15}
\bibinfo{author}{Gautam, R.}, \bibinfo{author}{Vanga, S.}, \& \bibinfo{author}{Ariese, F.} (\bibinfo{year}{2015}).
\newblock \bibinfo{title}{Review of multidimensional data processing approaches for raman and infrared spectroscopy}.
\newblock {\it \bibinfo{journal}{EPJ techniques and instrumentation}\/},  {\it \bibinfo{volume}{2}\/}, \bibinfo{pages}{1}.
\bibitem[{Ge(2014)}]{ge14}
\bibinfo{author}{Ge, Z.} (\bibinfo{year}{2014}).
\newblock \bibinfo{title}{Active learning strategy for smart soft sensor development under a small number of labeled data samples}.
\newblock {\it \bibinfo{journal}{Journal of Process Control}\/},  {\it \bibinfo{volume}{24}\/}, \bibinfo{pages}{1454 – 1461}.
\bibitem[{Ge(2015)}]{ge15}
\bibinfo{author}{Ge, Z.} (\bibinfo{year}{2015}).
\newblock \bibinfo{title}{Mixture bayesian regularization of pcr model and soft sensing application}.
\newblock {\it \bibinfo{journal}{Industrial Electronics, IEEE Transactions on}\/},  {\it \bibinfo{volume}{62}\/}, \bibinfo{pages}{4336--4343}.
\bibitem[{Ge et~al.(2014)Ge, Huang \& Song}]{gehu14}
\bibinfo{author}{Ge, Z.}, \bibinfo{author}{Huang, B.}, \& \bibinfo{author}{Song, Z.} (\bibinfo{year}{2014}).
\newblock \bibinfo{title}{Mixture semisupervised principal component regression model and soft sensor application}.
\newblock {\it \bibinfo{journal}{AIChE journal}\/},  {\it \bibinfo{volume}{60}\/}, \bibinfo{pages}{533--545}.
\bibitem[{Ge \& Song(2011)}]{geso11}
\bibinfo{author}{Ge, Z.}, \& \bibinfo{author}{Song, Z.} (\bibinfo{year}{2011}).
\newblock \bibinfo{title}{Semisupervised bayesian method for soft sensor modeling with unlabeled data samples}.
\newblock {\it \bibinfo{journal}{AIChE journal}\/},  {\it \bibinfo{volume}{57}\/}, \bibinfo{pages}{2109--2119}.
\bibitem[{Ge et~al.(2013)Ge, Song \& Gao}]{geso13}
\bibinfo{author}{Ge, Z.}, \bibinfo{author}{Song, Z.}, \& \bibinfo{author}{Gao, F.} (\bibinfo{year}{2013}).
\newblock \bibinfo{title}{Review of recent research on data-based process monitoring}.
\newblock {\it \bibinfo{journal}{Industrial \& Engineering Chemistry Research}\/},  {\it \bibinfo{volume}{52}\/}, \bibinfo{pages}{3543--3562}.
\bibitem[{Ghojogh et~al.(2019)Ghojogh, Samad, Mashhadi, Kapoor, Ali, Karray \& Crowley}]{ghsa19}
\bibinfo{author}{Ghojogh, B.}, \bibinfo{author}{Samad, M.}, \bibinfo{author}{Mashhadi, S.}, \bibinfo{author}{Kapoor, T.}, \bibinfo{author}{Ali, W.}, \bibinfo{author}{Karray, F.}, \& \bibinfo{author}{Crowley, M.} (\bibinfo{year}{2019}).
\newblock \bibinfo{title}{Feature selection and feature extraction in pattern analysis: A literature review}.
\newblock {\it \bibinfo{journal}{ArXiv preprint arXiv:1905.02845}\/}, .
\bibitem[{Gill(2015)}]{gi15}
\bibinfo{author}{Gill, J.} (\bibinfo{year}{2015}).
\newblock {\it \bibinfo{title}{Bayesian methods : a social and behavioral sciences approach}\/}.
\newblock Chapman {\&} Hall/CRC statistics in the social and behavioral sciences series. (\bibinfo{edition}{third edition.} ed.).
\newblock \bibinfo{address}{Boca Raton}: \bibinfo{publisher}{CRC Press}.
\bibitem[{Gopakumar et~al.(2018)Gopakumar, Tiwari \& Rahman}]{goti18}
\bibinfo{author}{Gopakumar, V.}, \bibinfo{author}{Tiwari, S.}, \& \bibinfo{author}{Rahman, I.} (\bibinfo{year}{2018}).
\newblock \bibinfo{title}{A deep learning based data driven soft sensor for bioprocesses}.
\newblock {\it \bibinfo{journal}{Biochemical Engineering Journal}\/},  {\it \bibinfo{volume}{136}\/}, \bibinfo{pages}{28 – 39}.
\bibitem[{Gysels et~al.(2005)Gysels, Renevey \& Celka}]{gyre05}
\bibinfo{author}{Gysels, E.}, \bibinfo{author}{Renevey, P.}, \& \bibinfo{author}{Celka, P.} (\bibinfo{year}{2005}).
\newblock \bibinfo{title}{Svm-based recursive feature elimination to compare phase synchronization computed from broadband and narrowband eeg signals in brain--computer interfaces}.
\newblock {\it \bibinfo{journal}{Signal processing}\/},  {\it \bibinfo{volume}{85}\/}, \bibinfo{pages}{2178--2189}.
\bibitem[{Hamedi~Rad et~al.(2019)Hamedi~Rad, Chao, Weisberg, Lian, Sinha \& Zhao}]{hach19}
\bibinfo{author}{Hamedi~Rad, S.}, \bibinfo{author}{Chao, R.}, \bibinfo{author}{Weisberg, S.}, \bibinfo{author}{Lian, J.}, \bibinfo{author}{Sinha, S.}, \& \bibinfo{author}{Zhao, H.} (\bibinfo{year}{2019}).
\newblock \bibinfo{title}{Towards a fully automated algorithm driven platform for biosystems design}.
\newblock {\it \bibinfo{journal}{Nature Communications}\/},  {\it \bibinfo{volume}{10}\/}, \bibinfo{pages}{5150}.
\bibitem[{Hashizume et~al.(2023)Hashizume, Ozawa \& Bei-Wen}]{haoz23}
\bibinfo{author}{Hashizume, T.}, \bibinfo{author}{Ozawa, Y.}, \& \bibinfo{author}{Bei-Wen, Y.} (\bibinfo{year}{2023}).
\newblock \bibinfo{title}{Employing active learning in the optimization of culture medium for mammalian cells}.
\newblock {\it \bibinfo{journal}{npj Systems Biology and Applications}\/},  {\it \bibinfo{volume}{9}\/}, \bibinfo{pages}{20}.
\bibitem[{He \& Yang(2008)}]{heya08}
\bibinfo{author}{He, X.~B.}, \& \bibinfo{author}{Yang, Y.~P.} (\bibinfo{year}{2008}).
\newblock \bibinfo{title}{Variable mwpca for adaptive process monitoring}.
\newblock {\it \bibinfo{journal}{Industrial {\&} engineering chemistry research}\/},  {\it \bibinfo{volume}{47}\/}, \bibinfo{pages}{419--427}.
\bibitem[{He et~al.(2022)He, Luo \& Ranzi}]{helu22}
\bibinfo{author}{He, Y.}, \bibinfo{author}{Luo, F.}, \& \bibinfo{author}{Ranzi, G.} (\bibinfo{year}{2022}).
\newblock \bibinfo{title}{Transferrable model-agnostic meta-learning for short-term household load forecasting with limited training data}.
\newblock {\it \bibinfo{journal}{IEEE Transactions on Power Systems}\/},  {\it \bibinfo{volume}{37}\/}, \bibinfo{pages}{3177--3180}.
\bibitem[{Helleckes et~al.(2023)Helleckes, Hemmerich, Wiechert, von Lieres \& Gr{\"u}nberger}]{hehe23}
\bibinfo{author}{Helleckes, L.~M.}, \bibinfo{author}{Hemmerich, J.}, \bibinfo{author}{Wiechert, W.}, \bibinfo{author}{von Lieres, E.}, \& \bibinfo{author}{Gr{\"u}nberger, A.} (\bibinfo{year}{2023}).
\newblock \bibinfo{title}{Machine learning in bioprocess development: from promise to practice}.
\newblock {\it \bibinfo{journal}{Trends in Biotechnology}\/},  {\it \bibinfo{volume}{41}\/}, \bibinfo{pages}{817--835}.
\bibitem[{Helleckes et~al.(2024)Helleckes, Wirnsperger, Polak, Guillén-Gosálbez, Butté \& von Stosch}]{hewi24}
\bibinfo{author}{Helleckes, L.~M.}, \bibinfo{author}{Wirnsperger, C.}, \bibinfo{author}{Polak, J.}, \bibinfo{author}{Guillén-Gosálbez, G.}, \bibinfo{author}{Butté, A.}, \& \bibinfo{author}{von Stosch, M.} (\bibinfo{year}{2024}).
\newblock \bibinfo{title}{Novel calibration design improves knowledge transfer across products for the characterization of pharmaceutical bioprocesses}.
\newblock {\it \bibinfo{journal}{Biotechnology Journal}\/},  {\it \bibinfo{volume}{19}\/}, \bibinfo{pages}{2400080}.
\bibitem[{Hoffmann et~al.(2022)Hoffmann, Borgeaud, Mensch, Buchatskaya, Cai, Rutherford, Casas, Hendricks, Welbl, Clark, Hennigan, Noland, Millican, Driessche, Damoc, Guy, Osindero, Simonyan, Elsen, Rae, Vinyals \& Sifre}]{hobo22}
\bibinfo{author}{Hoffmann, J.}, \bibinfo{author}{Borgeaud, S.}, \bibinfo{author}{Mensch, A.}, \bibinfo{author}{Buchatskaya, E.}, \bibinfo{author}{Cai, T.}, \bibinfo{author}{Rutherford, E.}, \bibinfo{author}{Casas, D. d.~L.}, \bibinfo{author}{Hendricks, L.~A.}, \bibinfo{author}{Welbl, J.}, \bibinfo{author}{Clark, A.}, \bibinfo{author}{Hennigan, T.}, \bibinfo{author}{Noland, E.}, \bibinfo{author}{Millican, K.}, \bibinfo{author}{Driessche, G. v.~d.}, \bibinfo{author}{Damoc, B.}, \bibinfo{author}{Guy, A.}, \bibinfo{author}{Osindero, S.}, \bibinfo{author}{Simonyan, K.}, \bibinfo{author}{Elsen, E.}, \bibinfo{author}{Rae, J.~W.}, \bibinfo{author}{Vinyals, O.}, \& \bibinfo{author}{Sifre, L.} (\bibinfo{year}{2022}).
\newblock \bibinfo{title}{Training compute-optimal large language models}.
\newblock {\it \bibinfo{journal}{arXiv preprint arXiv:2203.15556}\/}, .
\bibitem[{Hong et~al.(2021)Hong, Lu, Ou, Wolfrum, Springs, Sinskey \& Braatz}]{holu21}
\bibinfo{author}{Hong, M.~S.}, \bibinfo{author}{Lu, A.~E.}, \bibinfo{author}{Ou, R.~W.}, \bibinfo{author}{Wolfrum, J.~M.}, \bibinfo{author}{Springs, S.~L.}, \bibinfo{author}{Sinskey, A.~J.}, \& \bibinfo{author}{Braatz, R.~D.} (\bibinfo{year}{2021}).
\newblock \bibinfo{title}{Model-based control for column-based continuous viral inactivation of biopharmaceuticals}.
\newblock {\it \bibinfo{journal}{Biotechnology and Bioengineering}\/},  {\it \bibinfo{volume}{118}\/}, \bibinfo{pages}{3215 – 3224}.
\bibitem[{Hotelling(1936)}]{ho36}
\bibinfo{author}{Hotelling, H.} (\bibinfo{year}{1936}).
\newblock \bibinfo{title}{Relations between two sets of variates}.
\newblock {\it \bibinfo{journal}{Biometrika}\/},  {\it \bibinfo{volume}{28}\/}, \bibinfo{pages}{321--377}.
\bibitem[{Hsiao et~al.(2021)Hsiao, Kang \& Wong}]{hska21}
\bibinfo{author}{Hsiao, Y.-D.}, \bibinfo{author}{Kang, J.-L.}, \& \bibinfo{author}{Wong, D. S.-H.} (\bibinfo{year}{2021}).
\newblock \bibinfo{title}{Development of robust and physically interpretable soft sensor for industrial distillation column using transfer learning with small datasets}.
\newblock {\it \bibinfo{journal}{Processes}\/},  {\it \bibinfo{volume}{9}\/}.
\newblock \bibinfo{note}{; All Open Access, Gold Open Access, Green Open Access}.
\bibitem[{Hua et~al.(2023)Hua, Zhang, Sun, Li, Xiong \& Nazir}]{huzh23}
\bibinfo{author}{Hua, L.}, \bibinfo{author}{Zhang, C.}, \bibinfo{author}{Sun, W.}, \bibinfo{author}{Li, Y.}, \bibinfo{author}{Xiong, J.}, \& \bibinfo{author}{Nazir, M.~S.} (\bibinfo{year}{2023}).
\newblock \bibinfo{title}{An evolutionary deep learning soft sensor model based on random forest feature selection technique for penicillin fermentation process}.
\newblock {\it \bibinfo{journal}{ISA Transactions}\/},  {\it \bibinfo{volume}{136}\/}, \bibinfo{pages}{139 – 151}.
\bibitem[{Huang et~al.(1998)Huang, Shen, Long, Wu, Shih, Zheng, Yen, Tung \& Liu}]{hush98}
\bibinfo{author}{Huang, N.~E.}, \bibinfo{author}{Shen, Z.}, \bibinfo{author}{Long, S.~R.}, \bibinfo{author}{Wu, M.~C.}, \bibinfo{author}{Shih, H.~H.}, \bibinfo{author}{Zheng, Q.}, \bibinfo{author}{Yen, N.-C.}, \bibinfo{author}{Tung, C.~C.}, \& \bibinfo{author}{Liu, H.~H.} (\bibinfo{year}{1998}).
\newblock \bibinfo{title}{The empirical mode decomposition and the hilbert spectrum for nonlinear and non-stationary time series analysis}.
\newblock {\it \bibinfo{journal}{Proceedings of the Royal Society. A, Mathematical, physical, and engineering sciences}\/},  {\it \bibinfo{volume}{454}\/}, \bibinfo{pages}{903--995}.
\bibitem[{Huisman et~al.(2021)Huisman, van Rijn \& Plaat}]{huva21}
\bibinfo{author}{Huisman, M.}, \bibinfo{author}{van Rijn, J.~N.}, \& \bibinfo{author}{Plaat, A.} (\bibinfo{year}{2021}).
\newblock \bibinfo{title}{A survey of deep meta-learning}.
\newblock {\it \bibinfo{journal}{Artificial Intelligence Review}\/},  {\it \bibinfo{volume}{54}\/}, \bibinfo{pages}{4483 – 4541}.
\newblock \bibinfo{note}{; All Open Access, Green Open Access, Hybrid Gold Open Access}.
\bibitem[{Hutter et~al.(2021)Hutter, Stosch, Cruz~Bournazou \& Butté}]{hust21}
\bibinfo{author}{Hutter, C.}, \bibinfo{author}{Stosch, M.}, \bibinfo{author}{Cruz~Bournazou, M.~N.}, \& \bibinfo{author}{Butté, A.} (\bibinfo{year}{2021}).
\newblock \bibinfo{title}{Knowledge transfer across cell lines using hybrid gaussian process models with entity embedding vectors}.
\newblock {\it \bibinfo{journal}{Biotechnology and bioengineering}\/},  {\it \bibinfo{volume}{118}\/}, \bibinfo{pages}{4389--4401}.
\bibitem[{Ignova et~al.(1997)Ignova, Glassey, Ward \& Montague}]{iggl97}
\bibinfo{author}{Ignova, M.}, \bibinfo{author}{Glassey, J.}, \bibinfo{author}{Ward, A.}, \& \bibinfo{author}{Montague, G.} (\bibinfo{year}{1997}).
\newblock \bibinfo{title}{Multivariate statistical methods in bioprocess fault detection and performance forecasting}.
\newblock {\it \bibinfo{journal}{Transactions of the Institute of Measurement \& Control}\/},  {\it \bibinfo{volume}{19}\/}, \bibinfo{pages}{271--279}.
\bibitem[{Isoko et~al.(2024)Isoko, Cordiner, Kis \& Moghadam}]{isco24}
\bibinfo{author}{Isoko, K.}, \bibinfo{author}{Cordiner, J.~L.}, \bibinfo{author}{Kis, Z.}, \& \bibinfo{author}{Moghadam, P.~Z.} (\bibinfo{year}{2024}).
\newblock \bibinfo{title}{Bioprocessing 4.0: a pragmatic review and future perspectives}.
\newblock {\it \bibinfo{journal}{Digital discovery}\/},  {\it \bibinfo{volume}{3}\/}, \bibinfo{pages}{1662--1681}.
\bibitem[{Iwana \& Uchida(2020)}]{iwuc20}
\bibinfo{author}{Iwana, B.~K.}, \& \bibinfo{author}{Uchida, S.} (\bibinfo{year}{2020}).
\newblock \bibinfo{title}{Time series data augmentation for neural networks by time warping with a discriminative teacher}.
\bibitem[{Iwana \& Uchida(2021)}]{iwuc21}
\bibinfo{author}{Iwana, B.~K.}, \& \bibinfo{author}{Uchida, S.} (\bibinfo{year}{2021}).
\newblock \bibinfo{title}{An empirical survey of data augmentation for time series classification with neural networks}.
\newblock {\it \bibinfo{journal}{PLoS ONE}\/},  {\it \bibinfo{volume}{16}\/}, \bibinfo{pages}{e0254841}.
\bibitem[{Ji et~al.(2012)Ji, Wang, Chen, Liu, Zhang \& Yan}]{jiwa12}
\bibinfo{author}{Ji, J.}, \bibinfo{author}{Wang, H.}, \bibinfo{author}{Chen, K.}, \bibinfo{author}{Liu, Y.}, \bibinfo{author}{Zhang, N.}, \& \bibinfo{author}{Yan, J.} (\bibinfo{year}{2012}).
\newblock \bibinfo{title}{Recursive weighted kernel regression for semi-supervised soft-sensing modeling of fed-batch processes}.
\newblock {\it \bibinfo{journal}{Journal of the Taiwan Institute of Chemical Engineers}\/},  {\it \bibinfo{volume}{43}\/}, \bibinfo{pages}{67--76}.
\bibitem[{Jiang \& Ge(2022)}]{jige22}
\bibinfo{author}{Jiang, X.}, \& \bibinfo{author}{Ge, Z.} (\bibinfo{year}{2022}).
\newblock \bibinfo{title}{Improving the performance of just-in-time learning-based soft sensor through data augmentation}.
\newblock {\it \bibinfo{journal}{IEEE Transactions on Industrial Electronics}\/},  {\it \bibinfo{volume}{69}\/}, \bibinfo{pages}{13716 – 13726}.
\bibitem[{Jin et~al.(2015)Jin, Chen, Yang, Zhang, Wang \& Wu}]{jich15}
\bibinfo{author}{Jin, H.}, \bibinfo{author}{Chen, X.}, \bibinfo{author}{Yang, J.}, \bibinfo{author}{Zhang, H.}, \bibinfo{author}{Wang, L.}, \& \bibinfo{author}{Wu, L.} (\bibinfo{year}{2015}).
\newblock \bibinfo{title}{Multi-model adaptive soft sensor modeling method using local learning and online support vector regression for nonlinear time-variant batch processes}.
\newblock {\it \bibinfo{journal}{Chemical Engineering Science}\/},  {\it \bibinfo{volume}{131}\/}, \bibinfo{pages}{282--303}.
\bibitem[{{Joe Qin}(1998)}]{qi98}
\bibinfo{author}{{Joe Qin}, S.} (\bibinfo{year}{1998}).
\newblock \bibinfo{title}{Recursive pls algorithms for adaptive data modeling}.
\newblock {\it \bibinfo{journal}{Computers \& Chemical Engineering}\/},  {\it \bibinfo{volume}{22}\/}, \bibinfo{pages}{503--514}.
\bibitem[{Joosten et~al.(2008)Joosten, Cohen, Emsley, Mooij, Lamzin \& Perrakis}]{joco08}
\bibinfo{author}{Joosten, K.}, \bibinfo{author}{Cohen, S.~X.}, \bibinfo{author}{Emsley, P.}, \bibinfo{author}{Mooij, W.}, \bibinfo{author}{Lamzin, V.~S.}, \& \bibinfo{author}{Perrakis, A.} (\bibinfo{year}{2008}).
\newblock \bibinfo{title}{A knowledge-driven approach for crystallographic protein model completion}.
\newblock {\it \bibinfo{journal}{Acta crystallographica. Section D, Biological crystallography.}\/},  {\it \bibinfo{volume}{64}\/}, \bibinfo{pages}{416--424}.
\bibitem[{K et~al.(2022)K, Tayal, George, Singla, Kose, Singla, K, Tayal \& George}]{kata22}
\bibinfo{author}{K, H.}, \bibinfo{author}{Tayal, S.}, \bibinfo{author}{George, P.~M.}, \bibinfo{author}{Singla, P.}, \bibinfo{author}{Kose, U.}, \bibinfo{author}{Singla, P.}, \bibinfo{author}{K, H.}, \bibinfo{author}{Tayal, S.}, \& \bibinfo{author}{George, P.~M.} (\bibinfo{year}{2022}).
\newblock {\it \bibinfo{title}{Bayesian Reasoning and Gaussian Processes for Machine Learning Applications}\/}.
\newblock (\bibinfo{edition}{First edition} ed.).
\newblock \bibinfo{address}{United Kingdom}: \bibinfo{publisher}{Chapman \& Hall}.
\bibitem[{Kadlec(2009)}]{ka09}
\bibinfo{author}{Kadlec, P.} (\bibinfo{year}{2009}).
\newblock {\it \bibinfo{title}{On robust and adaptive soft sensors.}\/}.
\newblock Ph.D. thesis Bournemouth University.
\bibitem[{Kadlec \& Gabrys(2009{\natexlab{a}})}]{kaga09d}
\bibinfo{author}{Kadlec, P.}, \& \bibinfo{author}{Gabrys, B.} (\bibinfo{year}{2009}{\natexlab{a}}).
\newblock \bibinfo{title}{Architecture for development of adaptive on-line prediction models}.
\newblock {\it \bibinfo{journal}{Memetic Computing}\/},  {\it \bibinfo{volume}{1}\/}, \bibinfo{pages}{241--269}.
\bibitem[{Kadlec \& Gabrys(2009{\natexlab{b}})}]{kaga09b}
\bibinfo{author}{Kadlec, P.}, \& \bibinfo{author}{Gabrys, B.} (\bibinfo{year}{2009}{\natexlab{b}}).
\newblock \bibinfo{title}{Evolving on-line prediction model dealing with industrial data sets}.
\newblock In {\it \bibinfo{booktitle}{2009 {IEEE} Workshop on Evolving and Self-Developing Intelligent Systems}\/}.
\newblock \bibinfo{publisher}{{IEEE}}.
\bibitem[{Kadlec \& Gabrys(2009{\natexlab{c}})}]{kaga09c}
\bibinfo{author}{Kadlec, P.}, \& \bibinfo{author}{Gabrys, B.} (\bibinfo{year}{2009}{\natexlab{c}}).
\newblock \bibinfo{title}{Soft sensor based on adaptive local learning}.
\newblock In {\it \bibinfo{booktitle}{Advances in Neuro-Information Processing}\/} (pp. \bibinfo{pages}{1172--1179}).
\newblock \bibinfo{publisher}{Springer Berlin Heidelberg}.
\bibitem[{Kadlec \& Gabrys(2009{\natexlab{d}})}]{kaga09a}
\bibinfo{author}{Kadlec, P.}, \& \bibinfo{author}{Gabrys, B.} (\bibinfo{year}{2009}{\natexlab{d}}).
\newblock \bibinfo{title}{Soft sensors: where are we and what are the current and future challenges?}
\newblock {\it \bibinfo{journal}{{IFAC} Proceedings Volumes}\/},  {\it \bibinfo{volume}{42}\/}, \bibinfo{pages}{572--577}.
\bibitem[{Kadlec \& Gabrys(2010{\natexlab{a}})}]{kaga10a}
\bibinfo{author}{Kadlec, P.}, \& \bibinfo{author}{Gabrys, B.} (\bibinfo{year}{2010}{\natexlab{a}}).
\newblock \bibinfo{title}{Adaptive on-line prediction soft sensing without historical data}.
\newblock In {\it \bibinfo{booktitle}{The 2010 International Joint Conference on Neural Networks ({IJCNN})}\/}.
\newblock \bibinfo{publisher}{{IEEE}}.
\bibitem[{Kadlec \& Gabrys(2010{\natexlab{b}})}]{kaga10}
\bibinfo{author}{Kadlec, P.}, \& \bibinfo{author}{Gabrys, B.} (\bibinfo{year}{2010}{\natexlab{b}}).
\newblock \bibinfo{title}{Local learning-based adaptive soft sensor for catalyst activation prediction}.
\newblock {\it \bibinfo{journal}{{AIChE} Journal}\/},  {\it \bibinfo{volume}{57}\/}, \bibinfo{pages}{1288--1301}.
\bibitem[{Kadlec et~al.(2009)Kadlec, Gabrys \& Strandt}]{kaga09}
\bibinfo{author}{Kadlec, P.}, \bibinfo{author}{Gabrys, B.}, \& \bibinfo{author}{Strandt, S.} (\bibinfo{year}{2009}).
\newblock \bibinfo{title}{Data-driven soft sensors in the process industry}.
\newblock {\it \bibinfo{journal}{Computers {\&} chemical engineering}\/},  {\it \bibinfo{volume}{33}\/}, \bibinfo{pages}{795--814}.
\bibitem[{Kadlec et~al.(2011)Kadlec, Grbić \& Gabrys}]{kagr11}
\bibinfo{author}{Kadlec, P.}, \bibinfo{author}{Grbić, R.}, \& \bibinfo{author}{Gabrys, B.} (\bibinfo{year}{2011}).
\newblock \bibinfo{title}{Review of adaptation mechanisms for data-driven soft sensors}.
\newblock {\it \bibinfo{journal}{Computers and Chemical Engineering}\/},  {\it \bibinfo{volume}{35}\/}, \bibinfo{pages}{1 – 24}.
\bibitem[{Kamycki \& Kapuscinski(2020)}]{kaka20}
\bibinfo{author}{Kamycki, K.}, \& \bibinfo{author}{Kapuscinski, T.} (\bibinfo{year}{2020}).
\newblock \bibinfo{title}{Data augmentation with suboptimal warping for time-series classification}.
\newblock {\it \bibinfo{journal}{Sensors}\/},  {\it \bibinfo{volume}{20}\/}, \bibinfo{pages}{98}.
\bibitem[{Kaneko \& Funatsu(2013)}]{kafu13}
\bibinfo{author}{Kaneko, H.}, \& \bibinfo{author}{Funatsu, K.} (\bibinfo{year}{2013}).
\newblock \bibinfo{title}{Classification of the degradation of soft sensor models and discussion on adaptive models}.
\newblock {\it \bibinfo{journal}{AIChE Journal}\/},  {\it \bibinfo{volume}{59}\/}, \bibinfo{pages}{2339--2347}.
\bibitem[{Kano \& Fujiwara(2013)}]{kafU13a}
\bibinfo{author}{Kano, M.}, \& \bibinfo{author}{Fujiwara, K.} (\bibinfo{year}{2013}).
\newblock \bibinfo{title}{Virtual sensing technology in process industries: trends and challenges revealed by recent industrial applications}.
\newblock {\it \bibinfo{journal}{Journal of chemical engineering of Japan}\/},  {\it \bibinfo{volume}{46}\/}, \bibinfo{pages}{1--17}.
\bibitem[{Kasemiire et~al.(2021)Kasemiire, Avohou, De~Bleye, Sacre, Dumont, Hubert \& Ziemons}]{kaav21}
\bibinfo{author}{Kasemiire, A.}, \bibinfo{author}{Avohou, H.~T.}, \bibinfo{author}{De~Bleye, C.}, \bibinfo{author}{Sacre, P.-Y.}, \bibinfo{author}{Dumont, E.}, \bibinfo{author}{Hubert, P.}, \& \bibinfo{author}{Ziemons, E.} (\bibinfo{year}{2021}).
\newblock \bibinfo{title}{Design of experiments and design space approaches in the pharmaceutical bioprocess optimization}.
\newblock {\it \bibinfo{journal}{European journal of pharmaceutics and biopharmaceutics}\/},  {\it \bibinfo{volume}{166}\/}, \bibinfo{pages}{144--154}.
\bibitem[{Kawohl et~al.(2007)Kawohl, Heine \& King}]{kahe07}
\bibinfo{author}{Kawohl, M.}, \bibinfo{author}{Heine, T.}, \& \bibinfo{author}{King, R.} (\bibinfo{year}{2007}).
\newblock \bibinfo{title}{Model based estimation and optimal control of fed-batch fermentation processes for the production of antibiotics}.
\newblock {\it \bibinfo{journal}{Chemical Engineering and Processing: Process Intensification}\/},  {\it \bibinfo{volume}{46}\/}, \bibinfo{pages}{1223 – 1241}.
\bibitem[{Kedziora et~al.(2024)Kedziora, Musial \& Gabrys}]{kemu24}
\bibinfo{author}{Kedziora, D.~J.}, \bibinfo{author}{Musial, K.}, \& \bibinfo{author}{Gabrys, B.} (\bibinfo{year}{2024}).
\newblock \bibinfo{title}{Autonoml: Towards an integrated framework for autonomous machine learning}.
\newblock {\it \bibinfo{journal}{Foundations and Trends® in Machine Learning}\/},  {\it \bibinfo{volume}{17}\/}, \bibinfo{pages}{590--766}.
\bibitem[{Khatibisepehr et~al.(2013)Khatibisepehr, Huang \& Khare}]{khhu13}
\bibinfo{author}{Khatibisepehr, S.}, \bibinfo{author}{Huang, B.}, \& \bibinfo{author}{Khare, S.} (\bibinfo{year}{2013}).
\newblock \bibinfo{title}{Design of inferential sensors in the process industry: A review of bayesian methods}.
\newblock {\it \bibinfo{journal}{Journal of process control}\/},  {\it \bibinfo{volume}{23}\/}, \bibinfo{pages}{1575--1596}.
\bibitem[{Khuat et~al.(2025)Khuat, Bassett, Otte \& Gabrys}]{turo2024}
\bibinfo{author}{Khuat, T.~T.}, \bibinfo{author}{Bassett, R.}, \bibinfo{author}{Otte, E.}, \& \bibinfo{author}{Gabrys, B.} (\bibinfo{year}{2025}).
\newblock \bibinfo{title}{Online machine learning for real-time cell culture process monitoring}.
\newblock In {\it \bibinfo{booktitle}{Proceedings of the 37th Australasian Joint Conference on Artificial Intelligence}\/} (pp. \bibinfo{pages}{363--375}).
\bibitem[{Khuat et~al.(2024)Khuat, Bassett, Otte, Grevis-James \& Gabrys}]{khba24}
\bibinfo{author}{Khuat, T.~T.}, \bibinfo{author}{Bassett, R.}, \bibinfo{author}{Otte, E.}, \bibinfo{author}{Grevis-James, A.}, \& \bibinfo{author}{Gabrys, B.} (\bibinfo{year}{2024}).
\newblock \bibinfo{title}{Applications of machine learning in antibody discovery, process development, manufacturing and formulation: Current trends, challenges, and opportunities}.
\newblock {\it \bibinfo{journal}{Computers \& Chemical Engineering}\/},  {\it \bibinfo{volume}{182}\/}, \bibinfo{pages}{108585}.
\bibitem[{Khuat et~al.(2023)Khuat, Kedziora, Gabrys et~al.}]{khke23}
\bibinfo{author}{Khuat, T.~T.}, \bibinfo{author}{Kedziora, D.~J.}, \bibinfo{author}{Gabrys, B.} et~al. (\bibinfo{year}{2023}).
\newblock \bibinfo{title}{The roles and modes of human interactions with automated machine learning systems: A critical review and perspectives}.
\newblock {\it \bibinfo{journal}{Foundations and Trends{\textregistered} in Human--Computer Interaction}\/},  {\it \bibinfo{volume}{17}\/}, \bibinfo{pages}{195--387}.
\bibitem[{King et~al.(2004)King, Whelan, Jones, Reiser \& al}]{kiwh04}
\bibinfo{author}{King, R.~D.}, \bibinfo{author}{Whelan, K.~E.}, \bibinfo{author}{Jones, F.~M.}, \bibinfo{author}{Reiser, P. G.~K.}, \& \bibinfo{author}{al, e.} (\bibinfo{year}{2004}).
\newblock \bibinfo{title}{Functional genomic hypothesis generation and experimentation by a robot scientist}.
\newblock {\it \bibinfo{journal}{Nature}\/},  {\it \bibinfo{volume}{427}\/}, \bibinfo{pages}{247--52}.
\bibitem[{Kingma \& Welling(2022)}]{kiwe22}
\bibinfo{author}{Kingma, D.~P.}, \& \bibinfo{author}{Welling, M.} (\bibinfo{year}{2022}).
\newblock \bibinfo{title}{Auto-encoding variational bayes}.
\newblock {\it \bibinfo{journal}{arXiv preprint arXiv:1312.6114}\/}, .
\bibitem[{Kukal et~al.(2015)Kukal, Mareš, Náhlík, Hrnčiřík \& Klimt}]{kuma15}
\bibinfo{author}{Kukal, J.}, \bibinfo{author}{Mareš, J.}, \bibinfo{author}{Náhlík, J.}, \bibinfo{author}{Hrnčiřík, P.}, \& \bibinfo{author}{Klimt, M.} (\bibinfo{year}{2015}).
\newblock \bibinfo{title}{Automated classification of bioprocess based on optimum compromise whitening and clustering}.
\newblock {\it \bibinfo{journal}{Chemical and Biochemical Engineering Quarterly}\/},  {\it \bibinfo{volume}{29}\/}, \bibinfo{pages}{533--539}.
\bibitem[{Kulkarni et~al.(2004)Kulkarni, Chaudhary, Nandi, Tambe \& Kulkarni}]{kuch04}
\bibinfo{author}{Kulkarni, S.}, \bibinfo{author}{Chaudhary, A.}, \bibinfo{author}{Nandi, S.}, \bibinfo{author}{Tambe, S.}, \& \bibinfo{author}{Kulkarni, B.} (\bibinfo{year}{2004}).
\newblock \bibinfo{title}{Modeling and monitoring of batch processes using principal component analysis (pca) assisted generalized regression neural networks (grnn)}.
\newblock {\it \bibinfo{journal}{Biochemical Engineering Journal}\/},  {\it \bibinfo{volume}{18}\/}, \bibinfo{pages}{193--210}.
\bibitem[{Lange et~al.(2024)Lange, Thiele, Santolin, Riedel, Borisyak, Neubauer \& Cruz-Bournazou}]{lath24}
\bibinfo{author}{Lange, C.}, \bibinfo{author}{Thiele, I.}, \bibinfo{author}{Santolin, L.}, \bibinfo{author}{Riedel, S.~L.}, \bibinfo{author}{Borisyak, M.}, \bibinfo{author}{Neubauer, P.}, \& \bibinfo{author}{Cruz-Bournazou, M.~N.} (\bibinfo{year}{2024}).
\newblock \bibinfo{title}{Data augmentation scheme for raman spectra with highly correlated annotations}.
\newblock {\it \bibinfo{journal}{Computer Aided Chemical Engineering}\/},  {\it \bibinfo{volume}{53}\/}, \bibinfo{pages}{3055--3060}.
\bibitem[{Lateh et~al.(2017)Lateh, Muda, Yusof, Muda \& Azmi}]{lamu17}
\bibinfo{author}{Lateh, M.~A.}, \bibinfo{author}{Muda, A.~K.}, \bibinfo{author}{Yusof, Z. I.~M.}, \bibinfo{author}{Muda, N.~A.}, \& \bibinfo{author}{Azmi, M.~S.} (\bibinfo{year}{2017}).
\newblock \bibinfo{title}{Handling a small dataset problem in prediction model by employ artificial data generation approach: A review}.
\newblock In {\it \bibinfo{booktitle}{Journal of Physics: Conference Series}\/} (p. \bibinfo{pages}{012016}).
\newblock \bibinfo{organization}{IOP Publishing} volume \bibinfo{volume}{892}.
\bibitem[{Le~Guennec et~al.(2016)Le~Guennec, Malinowski \& Tavenard}]{lema16}
\bibinfo{author}{Le~Guennec, A.}, \bibinfo{author}{Malinowski, S.}, \& \bibinfo{author}{Tavenard, R.} (\bibinfo{year}{2016}).
\newblock \bibinfo{title}{{Data Augmentation for Time Series Classification using Convolutional Neural Networks}}.
\newblock In {\it \bibinfo{booktitle}{{ECML/PKDD Workshop on Advanced Analytics and Learning on Temporal Data}}\/}.
\newblock \bibinfo{address}{Riva Del Garda, Italy}.
\bibitem[{Lee \& Seung(1999)}]{lese99}
\bibinfo{author}{Lee, D.~D.}, \& \bibinfo{author}{Seung, H.~S.} (\bibinfo{year}{1999}).
\newblock \bibinfo{title}{Learning the parts of objects by non-negative matrix factorization}.
\newblock {\it \bibinfo{journal}{Nature}\/},  {\it \bibinfo{volume}{401}\/}, \bibinfo{pages}{788--791}.
\bibitem[{Lee et~al.(2005)Lee, Joung, Lee, Park \& Woo}]{lejo05}
\bibinfo{author}{Lee, M.~W.}, \bibinfo{author}{Joung, J.~Y.}, \bibinfo{author}{Lee, D.~S.}, \bibinfo{author}{Park, J.~M.}, \& \bibinfo{author}{Woo, S.~H.} (\bibinfo{year}{2005}).
\newblock \bibinfo{title}{Application of a moving-window-adaptive neural network to the modeling of a full-scale anaerobic filter process}.
\newblock {\it \bibinfo{journal}{Industrial {\&} engineering chemistry research}\/},  {\it \bibinfo{volume}{44}\/}, \bibinfo{pages}{3973--3982}.
\bibitem[{Lemke et~al.(2015)Lemke, Budka \& Gabrys}]{lebu15}
\bibinfo{author}{Lemke, C.}, \bibinfo{author}{Budka, M.}, \& \bibinfo{author}{Gabrys, B.} (\bibinfo{year}{2015}).
\newblock \bibinfo{title}{Metalearning: a survey of trends and technologies}.
\newblock {\it \bibinfo{journal}{Artificial Intelligence Review}\/},  {\it \bibinfo{volume}{44}\/}, \bibinfo{pages}{117--130}.
\bibitem[{Li et~al.(2000)Li, Yue, Valle-Cervantes \& Qin}]{liyu00}
\bibinfo{author}{Li, W.}, \bibinfo{author}{Yue, H.}, \bibinfo{author}{Valle-Cervantes, S.}, \& \bibinfo{author}{Qin, S.} (\bibinfo{year}{2000}).
\newblock \bibinfo{title}{Recursive pca for adaptive process monitoring}.
\newblock {\it \bibinfo{journal}{Journal of Process Control}\/},  {\it \bibinfo{volume}{10}\/}, \bibinfo{pages}{471--486}.
\bibitem[{Li et~al.(2018)Li, Deng, Wang, Lv \& Wu}]{lide18}
\bibinfo{author}{Li, X.}, \bibinfo{author}{Deng, S.}, \bibinfo{author}{Wang, S.}, \bibinfo{author}{Lv, Z.}, \& \bibinfo{author}{Wu, L.} (\bibinfo{year}{2018}).
\newblock \bibinfo{title}{Review of small data learning methods}.
\newblock In {\it \bibinfo{booktitle}{Proceedings of the IEEE 42nd annual computer software and applications conference (COMPSAC)}\/} (pp. \bibinfo{pages}{106--109}).
\newblock \bibinfo{organization}{IEEE} volume~\bibinfo{volume}{2}.
\bibitem[{Liu(2007)}]{li07}
\bibinfo{author}{Liu, J.} (\bibinfo{year}{2007}).
\newblock \bibinfo{title}{On-line soft sensor for polyethylene process with multiple production grades}.
\newblock {\it \bibinfo{journal}{Control Engineering Practice}\/},  {\it \bibinfo{volume}{15}\/}, \bibinfo{pages}{769--778}.
\bibitem[{Liu et~al.(2021)Liu, Hou \& Chen}]{liho21}
\bibinfo{author}{Liu, J.}, \bibinfo{author}{Hou, J.}, \& \bibinfo{author}{Chen, J.} (\bibinfo{year}{2021}).
\newblock \bibinfo{title}{Dual-layer feature extraction based soft sensor methods and applications to industrial polyethylene processes}.
\newblock {\it \bibinfo{journal}{Computers and Chemical Engineering}\/},  {\it \bibinfo{volume}{154}\/}.
\bibitem[{Liu et~al.(2015)Liu, Bassalo, Zeitoun \& Gill}]{liba15}
\bibinfo{author}{Liu, R.}, \bibinfo{author}{Bassalo, M.~C.}, \bibinfo{author}{Zeitoun, R.~I.}, \& \bibinfo{author}{Gill, R.~T.} (\bibinfo{year}{2015}).
\newblock \bibinfo{title}{Genome scale engineering techniques for metabolic engineering}.
\newblock {\it \bibinfo{journal}{Metabolic engineering}\/},  {\it \bibinfo{volume}{32}\/}, \bibinfo{pages}{143--154}.
\bibitem[{Liu et~al.(2009)Liu, Kruger, Littler, Xie \& Wang}]{likr09}
\bibinfo{author}{Liu, X.}, \bibinfo{author}{Kruger, U.}, \bibinfo{author}{Littler, T.}, \bibinfo{author}{Xie, L.}, \& \bibinfo{author}{Wang, S.} (\bibinfo{year}{2009}).
\newblock \bibinfo{title}{Moving window kernel pca for adaptive monitoring of nonlinear processes}.
\newblock {\it \bibinfo{journal}{Chemometrics and intelligent laboratory systems}\/},  {\it \bibinfo{volume}{96}\/}, \bibinfo{pages}{132--143}.
\bibitem[{Liu \& Gunawan(2017)}]{ligu17}
\bibinfo{author}{Liu, Y.}, \& \bibinfo{author}{Gunawan, R.} (\bibinfo{year}{2017}).
\newblock \bibinfo{title}{Bioprocess optimization under uncertainty using ensemble modeling}.
\newblock {\it \bibinfo{journal}{Journal of Biotechnology}\/},  {\it \bibinfo{volume}{244}\/}, \bibinfo{pages}{34 – 44}.
\bibitem[{Liu et~al.(2019)Liu, Yang, Liu, Chen \& Yao}]{liya19}
\bibinfo{author}{Liu, Y.}, \bibinfo{author}{Yang, C.}, \bibinfo{author}{Liu, K.}, \bibinfo{author}{Chen, B.}, \& \bibinfo{author}{Yao, Y.} (\bibinfo{year}{2019}).
\newblock \bibinfo{title}{Domain adaptation transfer learning soft sensor for product quality prediction}.
\newblock {\it \bibinfo{journal}{Chemometrics and Intelligent Laboratory Systems}\/},  {\it \bibinfo{volume}{192}\/}.
\bibitem[{Liu et~al.(2020)Liu, Yang, Zhang, Dai \& Yao}]{liya20}
\bibinfo{author}{Liu, Y.}, \bibinfo{author}{Yang, C.}, \bibinfo{author}{Zhang, M.}, \bibinfo{author}{Dai, Y.}, \& \bibinfo{author}{Yao, Y.} (\bibinfo{year}{2020}).
\newblock \bibinfo{title}{Development of adversarial transfer learning soft sensor for multigrade processes}.
\newblock {\it \bibinfo{journal}{Industrial and Engineering Chemistry Research}\/},  {\it \bibinfo{volume}{59}\/}, \bibinfo{pages}{16330 – 16345}.
\bibitem[{Lu et~al.(2023)Lu, Gong, Ye, Zhang \& Zhang}]{lugo23}
\bibinfo{author}{Lu, J.}, \bibinfo{author}{Gong, P.}, \bibinfo{author}{Ye, J.}, \bibinfo{author}{Zhang, J.}, \& \bibinfo{author}{Zhang, C.} (\bibinfo{year}{2023}).
\newblock \bibinfo{title}{A survey on machine learning from few samples}.
\newblock {\it \bibinfo{journal}{Pattern Recognition}\/},  {\it \bibinfo{volume}{139}\/}, \bibinfo{pages}{109480}.
\bibitem[{Magris \& Iosifidis(2023)}]{maio23}
\bibinfo{author}{Magris, M.}, \& \bibinfo{author}{Iosifidis, A.} (\bibinfo{year}{2023}).
\newblock \bibinfo{title}{Bayesian learning for neural networks: an algorithmic survey}.
\newblock {\it \bibinfo{journal}{Artificial Intelligence Review}\/},  {\it \bibinfo{volume}{56}\/}, \bibinfo{pages}{11773--11823}.
\bibitem[{Martínez et~al.(2011)Martínez, Cristaldi, Grau \& Lopes}]{macr11}
\bibinfo{author}{Martínez, E.}, \bibinfo{author}{Cristaldi, M.}, \bibinfo{author}{Grau, R.}, \& \bibinfo{author}{Lopes, J.} (\bibinfo{year}{2011}).
\newblock \bibinfo{title}{Dynamic optimization of bioreactors using probabilistic tendency models and bayesian active learning}.
\newblock {\it \bibinfo{journal}{Computer Aided Chemical Engineering}\/},  {\it \bibinfo{volume}{29}\/}, \bibinfo{pages}{783 – 787}.
\bibitem[{Mohanty et~al.(2022)Mohanty, Sutherland, Bezbradica \& Javidnia}]{mosu22}
\bibinfo{author}{Mohanty, A.}, \bibinfo{author}{Sutherland, A.}, \bibinfo{author}{Bezbradica, M.}, \& \bibinfo{author}{Javidnia, H.} (\bibinfo{year}{2022}).
\newblock \bibinfo{title}{Skin disease analysis with limited data in particular rosacea: A review and recommended framework}.
\newblock {\it \bibinfo{journal}{IEEE Access}\/},  {\it \bibinfo{volume}{10}\/}, \bibinfo{pages}{39045 – 39068}.
\bibitem[{Mumuni \& Mumuni(2025)}]{mumu25}
\bibinfo{author}{Mumuni, A.}, \& \bibinfo{author}{Mumuni, F.} (\bibinfo{year}{2025}).
\newblock \bibinfo{title}{Data augmentation with automated machine learning: approaches and performance comparison with classical data augmentation methods}.
\bibitem[{Murtaza et~al.(2023)Murtaza, Ahmed, Khan, Murtaza, Zafar \& Bano}]{muah23}
\bibinfo{author}{Murtaza, H.}, \bibinfo{author}{Ahmed, M.}, \bibinfo{author}{Khan, N.~F.}, \bibinfo{author}{Murtaza, G.}, \bibinfo{author}{Zafar, S.}, \& \bibinfo{author}{Bano, A.} (\bibinfo{year}{2023}).
\newblock \bibinfo{title}{Synthetic data generation: State of the art in health care domain}.
\newblock {\it \bibinfo{journal}{Computer Science Review}\/},  {\it \bibinfo{volume}{48}\/}, \bibinfo{pages}{100546}.
\bibitem[{Nagrath et~al.(2004)Nagrath, Messac, Bequette \& Cramer}]{name04}
\bibinfo{author}{Nagrath, D.}, \bibinfo{author}{Messac, A.}, \bibinfo{author}{Bequette, B.~W.}, \& \bibinfo{author}{Cramer, S.~M.} (\bibinfo{year}{2004}).
\newblock \bibinfo{title}{A hybrid model framework for the optimization of preparative chromatographic processes}.
\newblock {\it \bibinfo{journal}{Biotechnology progress}\/},  {\it \bibinfo{volume}{20}\/}, \bibinfo{pages}{162--178}.
\bibitem[{Naji et~al.(2019)Naji, Jalab \& Kareem}]{naja19}
\bibinfo{author}{Naji, S.}, \bibinfo{author}{Jalab, H.~A.}, \& \bibinfo{author}{Kareem, S.~A.} (\bibinfo{year}{2019}).
\newblock \bibinfo{title}{A survey on skin detection in colored images}.
\newblock {\it \bibinfo{journal}{The Artificial intelligence review}\/},  {\it \bibinfo{volume}{52}\/}, \bibinfo{pages}{1041--1087}.
\bibitem[{Nam et~al.(2020)Nam, Bu, Park, Seo, Jo \& Jeong}]{nabu20}
\bibinfo{author}{Nam, G.-H.}, \bibinfo{author}{Bu, S.-J.}, \bibinfo{author}{Park, N.-M.}, \bibinfo{author}{Seo, J.-Y.}, \bibinfo{author}{Jo, H.-C.}, \& \bibinfo{author}{Jeong, W.-T.} (\bibinfo{year}{2020}).
\newblock \bibinfo{title}{Data augmentation using empirical mode decomposition on neural networks to classify impact noise in vehicle}.
\newblock In {\it \bibinfo{booktitle}{ICASSP 2020 - 2020 IEEE International Conference on Acoustics, Speech and Signal Processing (ICASSP)}\/} (pp. \bibinfo{pages}{731--735}).
\bibitem[{Narayanan et~al.(2021{\natexlab{a}})Narayanan, Dingfelder, Condado~Morales, Patel, Heding, Bjelke, Egebjerg, Butte, Sokolov, Lorenzen \& Arosio}]{nadi21}
\bibinfo{author}{Narayanan, H.}, \bibinfo{author}{Dingfelder, F.}, \bibinfo{author}{Condado~Morales, I.}, \bibinfo{author}{Patel, B.}, \bibinfo{author}{Heding, K.~E.}, \bibinfo{author}{Bjelke, J.~R.}, \bibinfo{author}{Egebjerg, T.}, \bibinfo{author}{Butte, A.}, \bibinfo{author}{Sokolov, M.}, \bibinfo{author}{Lorenzen, N.}, \& \bibinfo{author}{Arosio, P.} (\bibinfo{year}{2021}{\natexlab{a}}).
\newblock \bibinfo{title}{Design of biopharmaceutical formulations accelerated by machine learning}.
\newblock {\it \bibinfo{journal}{Molecular pharmaceutics}\/},  {\it \bibinfo{volume}{18}\/}, \bibinfo{pages}{3843--3853}.
\bibitem[{Narayanan et~al.(2022)Narayanan, Luna, Sokolov, Butt{\'{e}} \& Morbidelli}]{nalu22}
\bibinfo{author}{Narayanan, H.}, \bibinfo{author}{Luna, M.}, \bibinfo{author}{Sokolov, M.}, \bibinfo{author}{Butt{\'{e}}, A.}, \& \bibinfo{author}{Morbidelli, M.} (\bibinfo{year}{2022}).
\newblock \bibinfo{title}{Hybrid models based on machine learning and an increasing degree of process knowledge: Application to cell culture processes}.
\newblock {\it \bibinfo{journal}{Industrial \& Engineering Chemistry Research}\/},  {\it \bibinfo{volume}{61}\/}, \bibinfo{pages}{8658--8672}.
\bibitem[{Narayanan et~al.(2021{\natexlab{b}})Narayanan, Seidler, Luna, Sokolov, Morbidelli \& Butt{\'e}}]{nase21}
\bibinfo{author}{Narayanan, H.}, \bibinfo{author}{Seidler, T.}, \bibinfo{author}{Luna, M.~F.}, \bibinfo{author}{Sokolov, M.}, \bibinfo{author}{Morbidelli, M.}, \& \bibinfo{author}{Butt{\'e}, A.} (\bibinfo{year}{2021}{\natexlab{b}}).
\newblock \bibinfo{title}{Hybrid models for the simulation and prediction of chromatographic processes for protein capture}.
\newblock {\it \bibinfo{journal}{Journal of Chromatography A}\/},  {\it \bibinfo{volume}{1650}\/}, \bibinfo{pages}{462248}.
\bibitem[{Narayanan et~al.(2019)Narayanan, Sokolov, Morbidelli \& Butt{\'{e}}}]{naso19}
\bibinfo{author}{Narayanan, H.}, \bibinfo{author}{Sokolov, M.}, \bibinfo{author}{Morbidelli, M.}, \& \bibinfo{author}{Butt{\'{e}}, A.} (\bibinfo{year}{2019}).
\newblock \bibinfo{title}{A new generation of predictive models: The added value of hybrid models for manufacturing processes of therapeutic proteins}.
\newblock {\it \bibinfo{journal}{Biotechnology and Bioengineering}\/},  {\it \bibinfo{volume}{116}\/}, \bibinfo{pages}{2540--2549}.
\bibitem[{Ng(2022)}]{Ng22}
\bibinfo{author}{Ng, A.} (\bibinfo{year}{2022}).
\newblock \bibinfo{title}{Data centric ai development from big data to good data}.
\newblock \bibinfo{howpublished}{Video}.
\bibitem[{Odeh-Couvertier et~al.(2022)Odeh-Couvertier, Dwarshuis, Colonna, Levine, Edison, Kotanchek, Roy \& Torres-Garcia}]{oddw22}
\bibinfo{author}{Odeh-Couvertier, V.~Y.}, \bibinfo{author}{Dwarshuis, N.~J.}, \bibinfo{author}{Colonna, M.~B.}, \bibinfo{author}{Levine, B.~L.}, \bibinfo{author}{Edison, A.~S.}, \bibinfo{author}{Kotanchek, T.}, \bibinfo{author}{Roy, K.}, \& \bibinfo{author}{Torres-Garcia, W.} (\bibinfo{year}{2022}).
\newblock \bibinfo{title}{Predicting t-cell quality during manufacturing through an artificial intelligence-based integrative multiomics analytical platform}.
\newblock {\it \bibinfo{journal}{Bioengineering and Translational Medicine}\/},  {\it \bibinfo{volume}{7}\/}.
\bibitem[{Odongo \& Han(2018)}]{odha18}
\bibinfo{author}{Odongo, S.~E.}, \& \bibinfo{author}{Han, D.~S.} (\bibinfo{year}{2018}).
\newblock \bibinfo{title}{Feature representation and data augmentation for human activity classification based on wearable imu sensor data using a deep lstm neural network}.
\newblock {\it \bibinfo{journal}{Sensors}\/},  {\it \bibinfo{volume}{18}\/}.
\bibitem[{Ohashi et~al.(2017)Ohashi, Al-Nasser, Ahmed, Akiyama, Sato, Nguyen, Nakamura \& Dengel}]{ohal17}
\bibinfo{author}{Ohashi, H.}, \bibinfo{author}{Al-Nasser, M.}, \bibinfo{author}{Ahmed, S.}, \bibinfo{author}{Akiyama, T.}, \bibinfo{author}{Sato, T.}, \bibinfo{author}{Nguyen, P.}, \bibinfo{author}{Nakamura, K.}, \& \bibinfo{author}{Dengel, A.} (\bibinfo{year}{2017}).
\newblock \bibinfo{title}{Augmenting wearable sensor data with physical constraint for dnn-based human-action recognition}.
\newblock In {\it \bibinfo{booktitle}{ICML 2017 times series workshop}\/} (pp. \bibinfo{pages}{6--11}).
\bibitem[{Oyetunde et~al.(2019)Oyetunde, Liu, Martin \& Tang}]{oyli19}
\bibinfo{author}{Oyetunde, T.}, \bibinfo{author}{Liu, D.}, \bibinfo{author}{Martin, H.~G.}, \& \bibinfo{author}{Tang, Y.~J.} (\bibinfo{year}{2019}).
\newblock \bibinfo{title}{Machine learning framework for assessment of microbial factory performance}.
\newblock {\it \bibinfo{journal}{PLoS ONE}\/},  {\it \bibinfo{volume}{14}\/}.
\bibitem[{Pan et~al.(2020)Pan, Li \& Fang}]{Pan2020}
\bibinfo{author}{Pan, Q.}, \bibinfo{author}{Li, X.}, \& \bibinfo{author}{Fang, L.} (\bibinfo{year}{2020}).
\newblock \bibinfo{title}{Data augmentation for deep learning-based ecg analysis}.
\newblock In \bibinfo{editor}{C.~Liu}, \& \bibinfo{editor}{J.~Li} (Eds.), {\it \bibinfo{booktitle}{Feature Engineering and Computational Intelligence in ECG Monitoring}\/} (pp. \bibinfo{pages}{91--111}).
\newblock \bibinfo{publisher}{Springer}.
\bibitem[{Pan \& Yang(2010)}]{paya10}
\bibinfo{author}{Pan, S.~J.}, \& \bibinfo{author}{Yang, Q.} (\bibinfo{year}{2010}).
\newblock \bibinfo{title}{A survey on transfer learning}.
\newblock {\it \bibinfo{journal}{IEEE Transactions on Knowledge and Data Engineering}\/},  {\it \bibinfo{volume}{22}\/}, \bibinfo{pages}{1345 – 1359}.
\bibitem[{Pan et~al.(2022)Pan, Chen, Zhang, Liu, He \& Lv}]{pach22}
\bibinfo{author}{Pan, T.}, \bibinfo{author}{Chen, J.}, \bibinfo{author}{Zhang, T.}, \bibinfo{author}{Liu, S.}, \bibinfo{author}{He, S.}, \& \bibinfo{author}{Lv, H.} (\bibinfo{year}{2022}).
\newblock \bibinfo{title}{Generative adversarial network in mechanical fault diagnosis under small sample: A systematic review on applications and future perspectives}.
\newblock {\it \bibinfo{journal}{ISA Transactions}\/},  {\it \bibinfo{volume}{128}\/}, \bibinfo{pages}{1 – 10}.
\bibitem[{Parnami \& Lee(2022)}]{pale22}
\bibinfo{author}{Parnami, A.}, \& \bibinfo{author}{Lee, M.} (\bibinfo{year}{2022}).
\newblock \bibinfo{title}{Learning from few examples: A summary of approaches to few-shot learning}.
\newblock {\it \bibinfo{journal}{ArXiv}\/},  {\it \bibinfo{volume}{abs/2203.04291}\/}.
\bibitem[{Pearson(1901)}]{pe01}
\bibinfo{author}{Pearson, K.} (\bibinfo{year}{1901}).
\newblock \bibinfo{title}{Liii. on lines and planes of closest fit to systems of points in space}.
\newblock {\it \bibinfo{journal}{The London, Edinburgh, and Dublin Philosophical Magazine and Journal of Science}\/},  {\it \bibinfo{volume}{2}\/}, \bibinfo{pages}{559--572}.
\bibitem[{Petitjean et~al.(2011)Petitjean, Ketterlin \& Gançarski}]{peke11}
\bibinfo{author}{Petitjean, F.}, \bibinfo{author}{Ketterlin, A.}, \& \bibinfo{author}{Gançarski, P.} (\bibinfo{year}{2011}).
\newblock \bibinfo{title}{A global averaging method for dynamic time warping, with applications to clustering}.
\newblock {\it \bibinfo{journal}{Pattern Recognition}\/},  {\it \bibinfo{volume}{44}\/}, \bibinfo{pages}{678--693}.
\bibitem[{Petsagkourakis et~al.(2020)Petsagkourakis, Sandoval, Bradford, Zhang \& del Rio-Chanona}]{pesa20}
\bibinfo{author}{Petsagkourakis, P.}, \bibinfo{author}{Sandoval, I.}, \bibinfo{author}{Bradford, E.}, \bibinfo{author}{Zhang, D.}, \& \bibinfo{author}{del Rio-Chanona, E.} (\bibinfo{year}{2020}).
\newblock \bibinfo{title}{Reinforcement learning for batch bioprocess optimization}.
\newblock {\it \bibinfo{journal}{Computers \& chemical engineering}\/},  {\it \bibinfo{volume}{133}\/}, \bibinfo{pages}{106649}.
\bibitem[{Pinto et~al.(2022)Pinto, Mestre, Ramos, Costa, Striedner \& Oliveira}]{pime22}
\bibinfo{author}{Pinto, J.}, \bibinfo{author}{Mestre, M.}, \bibinfo{author}{Ramos, J.}, \bibinfo{author}{Costa, R.~S.}, \bibinfo{author}{Striedner, G.}, \& \bibinfo{author}{Oliveira, R.} (\bibinfo{year}{2022}).
\newblock \bibinfo{title}{A general deep hybrid model for bioreactor systems: Combining first principles with deep neural networks}.
\newblock {\it \bibinfo{journal}{Computers and Chemical Engineering}\/},  {\it \bibinfo{volume}{165}\/}, \bibinfo{pages}{107952}.
\bibitem[{Pokhrel(2020)}]{po20}
\bibinfo{author}{Pokhrel, S.~R.} (\bibinfo{year}{2020}).
\newblock \bibinfo{title}{Federated learning meets blockchain at 6g edge: A drone-assisted networking for disaster response}.
\newblock In {\it \bibinfo{booktitle}{Proceedings of the 2nd ACM MobiCom Workshop on Drone Assisted Wireless Communications for 5G and Beyond}\/} DroneCom '20 (p. \bibinfo{pages}{49–54}).
\newblock \bibinfo{publisher}{Association for Computing Machinery}.
\bibitem[{Polikar et~al.(2001)Polikar, Udpa, Udpa \& Honavar}]{poud01}
\bibinfo{author}{Polikar, R.}, \bibinfo{author}{Udpa, L.}, \bibinfo{author}{Udpa, S.~S.}, \& \bibinfo{author}{Honavar, V.} (\bibinfo{year}{2001}).
\newblock \bibinfo{title}{Learn++: An incremental learning algorithm for supervised neural networks}.
\newblock {\it \bibinfo{journal}{IEEE Transactions on Systems, Man and Cybernetics Part C: Applications and Reviews}\/},  {\it \bibinfo{volume}{31}\/}, \bibinfo{pages}{497 – 508}.
\bibitem[{Putra et~al.(2021)Putra, Chen, Prayitno, Ogiela, Chou, Weng \& Shae}]{puch21}
\bibinfo{author}{Putra, K.~T.}, \bibinfo{author}{Chen, H.-C.}, \bibinfo{author}{Prayitno}, \bibinfo{author}{Ogiela, M.~R.}, \bibinfo{author}{Chou, C.-L.}, \bibinfo{author}{Weng, C.-E.}, \& \bibinfo{author}{Shae, Z.-Y.} (\bibinfo{year}{2021}).
\newblock \bibinfo{title}{Federated compressed learning edge computing framework with ensuring data privacy for pm2.5 prediction in smart city sensing applications}.
\newblock {\it \bibinfo{journal}{Sensors}\/},  {\it \bibinfo{volume}{21}\/}.
\bibitem[{Qiu et~al.(2022)Qiu, Wang, Zhou, Wang \& Guo}]{qiwa22}
\bibinfo{author}{Qiu, K.}, \bibinfo{author}{Wang, J.}, \bibinfo{author}{Zhou, X.}, \bibinfo{author}{Wang, R.}, \& \bibinfo{author}{Guo, Y.} (\bibinfo{year}{2022}).
\newblock \bibinfo{title}{Soft sensor based on localized semi-supervised relevance vector machine for penicillin fermentation process with asymmetric data}.
\newblock {\it \bibinfo{journal}{Measurement: Journal of the International Measurement Confederation}\/},  {\it \bibinfo{volume}{202}\/}.
\bibitem[{Quintero et~al.(2008)Quintero, Amicarelli, Di~Sciascio \& Scaglia}]{quam08}
\bibinfo{author}{Quintero, O.~L.}, \bibinfo{author}{Amicarelli, A.~A.}, \bibinfo{author}{Di~Sciascio, F.}, \& \bibinfo{author}{Scaglia, G.} (\bibinfo{year}{2008}).
\newblock \bibinfo{title}{State estimation in alcoholic continuous fermentation of zymomonas mobilis using recursive bayesian filtering: A simulation approach}.
\newblock {\it \bibinfo{journal}{BioResources}\/},  {\it \bibinfo{volume}{3}\/}, \bibinfo{pages}{316 – 334}.
\bibitem[{Quintero et~al.(2009)Quintero, Amicarelli, Scaglia \& di~Sciascio}]{quam09}
\bibinfo{author}{Quintero, O.~L.}, \bibinfo{author}{Amicarelli, A.~A.}, \bibinfo{author}{Scaglia, G.}, \& \bibinfo{author}{di~Sciascio, F.} (\bibinfo{year}{2009}).
\newblock \bibinfo{title}{Control based on numerical methods and recursive bayesian estimation in a continuous alcoholic fermentation process}.
\newblock {\it \bibinfo{journal}{BioResources}\/},  {\it \bibinfo{volume}{4}\/}, \bibinfo{pages}{1372 – 1395}.
\bibitem[{Radivojevi{\'{c}} et~al.(2020)Radivojevi{\'{c}}, Costello, Workman \& Garcia~Martin}]{raco20}
\bibinfo{author}{Radivojevi{\'{c}}, T.}, \bibinfo{author}{Costello, Z.}, \bibinfo{author}{Workman, K.}, \& \bibinfo{author}{Garcia~Martin, H.} (\bibinfo{year}{2020}).
\newblock \bibinfo{title}{A machine learning automated recommendation tool for synthetic biology}.
\newblock {\it \bibinfo{journal}{Nature Communications}\/},  {\it \bibinfo{volume}{11}\/}, \bibinfo{pages}{4879}.
\bibitem[{Raschka(2018)}]{ra18}
\bibinfo{author}{Raschka, S.} (\bibinfo{year}{2018}).
\newblock \bibinfo{title}{Model evaluation, model selection, and algorithm selection in machine learning}.
\newblock {\it \bibinfo{journal}{ArXiv}\/},  {\it \bibinfo{volume}{abs/1811.12808}\/}.
\bibitem[{Rashedi et~al.(2024)Rashedi, Khodabandehlou, Wang, Demers, Tulsyan, Garvin \& Undey}]{rakh24}
\bibinfo{author}{Rashedi, M.}, \bibinfo{author}{Khodabandehlou, H.}, \bibinfo{author}{Wang, T.}, \bibinfo{author}{Demers, M.}, \bibinfo{author}{Tulsyan, A.}, \bibinfo{author}{Garvin, C.}, \& \bibinfo{author}{Undey, C.} (\bibinfo{year}{2024}).
\newblock \bibinfo{title}{Integration of just-in-time learning with variational autoencoder for cell culture process monitoring based on raman spectroscopy}.
\newblock {\it \bibinfo{journal}{Biotechnology and Bioengineering}\/},  {\it \bibinfo{volume}{121}\/}, \bibinfo{pages}{2205--2224}.
\bibitem[{Rashid \& Louis(2019)}]{ralo19}
\bibinfo{author}{Rashid, K.~M.}, \& \bibinfo{author}{Louis, J.} (\bibinfo{year}{2019}).
\newblock \bibinfo{title}{Time-warping: A time series data augmentation of imu data for construction equipment activity identification}.
\newblock In {\it \bibinfo{booktitle}{Proceedings of the 36th International Symposium on Automation and Robotics in Construction (ISARC)}\/} (pp. \bibinfo{pages}{651--657}).
\bibitem[{Rathinavelu et~al.(2023)Rathinavelu, Pavan \& Sivaprakasam}]{rapa23}
\bibinfo{author}{Rathinavelu, S.}, \bibinfo{author}{Pavan, S.~S.}, \& \bibinfo{author}{Sivaprakasam, S.} (\bibinfo{year}{2023}).
\newblock \bibinfo{title}{Hybrid model-based framework for soft sensing and forecasting key process variables in the production of hyaluronic acid by streptococcus zooepidemicus}.
\newblock {\it \bibinfo{journal}{Biotechnology and Bioprocess Engineering}\/},  {\it \bibinfo{volume}{28}\/}, \bibinfo{pages}{203 – 214}.
\bibitem[{Ray et~al.(2021)Ray, Reddy \& Banerjee}]{rare21}
\bibinfo{author}{Ray, P.}, \bibinfo{author}{Reddy, S.~S.}, \& \bibinfo{author}{Banerjee, T.} (\bibinfo{year}{2021}).
\newblock \bibinfo{title}{Various dimension reduction techniques for high dimensional data analysis: a review}.
\newblock {\it \bibinfo{journal}{Artificial Intelligence Review}\/},  {\it \bibinfo{volume}{54}\/}, \bibinfo{pages}{3473 – 3515}.
\bibitem[{Rieke et~al.(2020)Rieke, Hancox, Li, Milletar{\`i}, Roth, Albarqouni, Bakas, Galtier, Landman, Maier-Hein, Ourselin, Sheller, Summers, Trask, Xu, Baust \& Cardoso}]{riha20}
\bibinfo{author}{Rieke, N.}, \bibinfo{author}{Hancox, J.}, \bibinfo{author}{Li, W.}, \bibinfo{author}{Milletar{\`i}, F.}, \bibinfo{author}{Roth, H.~R.}, \bibinfo{author}{Albarqouni, S.}, \bibinfo{author}{Bakas, S.}, \bibinfo{author}{Galtier, M.~N.}, \bibinfo{author}{Landman, B.~A.}, \bibinfo{author}{Maier-Hein, K.}, \bibinfo{author}{Ourselin, S.}, \bibinfo{author}{Sheller, M.}, \bibinfo{author}{Summers, R.~M.}, \bibinfo{author}{Trask, A.}, \bibinfo{author}{Xu, D.}, \bibinfo{author}{Baust, M.}, \& \bibinfo{author}{Cardoso, M.~J.} (\bibinfo{year}{2020}).
\newblock \bibinfo{title}{The future of digital health with federated learning}.
\newblock {\it \bibinfo{journal}{npj Digital Medicine}\/},  {\it \bibinfo{volume}{3}\/}, \bibinfo{pages}{119}.
\bibitem[{del Rio-Chanona et~al.(2016)del Rio-Chanona, Manirafasha, Zhang, Yue \& Jing}]{dema16}
\bibinfo{author}{del Rio-Chanona, E.~A.}, \bibinfo{author}{Manirafasha, E.}, \bibinfo{author}{Zhang, D.}, \bibinfo{author}{Yue, Q.}, \& \bibinfo{author}{Jing, K.} (\bibinfo{year}{2016}).
\newblock \bibinfo{title}{Dynamic modeling and optimization of cyanobacterial c-phycocyanin production process by artificial neural network}.
\newblock {\it \bibinfo{journal}{Algal Research}\/},  {\it \bibinfo{volume}{13}\/}, \bibinfo{pages}{7 – 15}.
\bibitem[{Rodríguez et~al.(2008)Rodríguez, Roca, Lema \& Bernard}]{roro08}
\bibinfo{author}{Rodríguez, J.}, \bibinfo{author}{Roca, E.}, \bibinfo{author}{Lema, J.~M.}, \& \bibinfo{author}{Bernard, O.} (\bibinfo{year}{2008}).
\newblock \bibinfo{title}{Determination of the adequate minimum model complexity required in anaerobic bioprocesses using experimental data}.
\newblock {\it \bibinfo{journal}{Journal of Chemical Technology and Biotechnology}\/},  {\it \bibinfo{volume}{83}\/}, \bibinfo{pages}{1694 – 1702}.
\bibitem[{Roell et~al.(2022)Roell, Sathish, Wan, Cheng, Wen, Tang \& Bao}]{rosa22}
\bibinfo{author}{Roell, G.~W.}, \bibinfo{author}{Sathish, A.}, \bibinfo{author}{Wan, N.}, \bibinfo{author}{Cheng, Q.}, \bibinfo{author}{Wen, Z.}, \bibinfo{author}{Tang, Y.~J.}, \& \bibinfo{author}{Bao, F.~S.} (\bibinfo{year}{2022}).
\newblock \bibinfo{title}{A comparative evaluation of machine learning algorithms for predicting syngas fermentation outcomes}.
\newblock {\it \bibinfo{journal}{Biochemical Engineering Journal}\/},  {\it \bibinfo{volume}{186}\/}, \bibinfo{pages}{108578}.
\bibitem[{Roger et~al.(2022)Roger, Farinas \& Pinquier}]{rofa22}
\bibinfo{author}{Roger, V.}, \bibinfo{author}{Farinas, J.}, \& \bibinfo{author}{Pinquier, J.} (\bibinfo{year}{2022}).
\newblock \bibinfo{title}{Deep neural networks for automatic speech processing: a survey from large corpora to limited data}.
\newblock {\it \bibinfo{journal}{Eurasip Journal on Audio, Speech, and Music Processing}\/},  {\it \bibinfo{volume}{2022}\/}.
\bibitem[{Rogers et~al.(2022)Rogers, Vega-Ramon, Yan, del Río-Chanona, Jing \& Zhang}]{rove22}
\bibinfo{author}{Rogers, A.}, \bibinfo{author}{Vega-Ramon, F.}, \bibinfo{author}{Yan, J.}, \bibinfo{author}{del Río-Chanona, E.}, \bibinfo{author}{Jing, K.}, \& \bibinfo{author}{Zhang, D.} (\bibinfo{year}{2022}).
\newblock \bibinfo{title}{A transfer learning approach for predictive modeling of bioprocesses using small data}.
\newblock {\it \bibinfo{journal}{Biotechnology and Bioengineering}\/},  {\it \bibinfo{volume}{119}\/}, \bibinfo{pages}{411--422}.
\bibitem[{Rosa et~al.(2010)Rosa, Ferreira, Azevedo \& Aires-Barros}]{rofe10}
\bibinfo{author}{Rosa, P.}, \bibinfo{author}{Ferreira, I.}, \bibinfo{author}{Azevedo, A.}, \& \bibinfo{author}{Aires-Barros, M.} (\bibinfo{year}{2010}).
\newblock \bibinfo{title}{Aqueous two-phase systems: A viable platform in the manufacturing of biopharmaceuticals}.
\newblock {\it \bibinfo{journal}{Journal of Chromatography A}\/},  {\it \bibinfo{volume}{1217}\/}, \bibinfo{pages}{2296--2305}.
\bibitem[{Roweis \& Saul(2000)}]{rosa00}
\bibinfo{author}{Roweis, S.~T.}, \& \bibinfo{author}{Saul, L.~K.} (\bibinfo{year}{2000}).
\newblock \bibinfo{title}{Nonlinear dimensionality reduction by locally linear embedding}.
\newblock {\it \bibinfo{journal}{Science}\/},  {\it \bibinfo{volume}{290}\/}, \bibinfo{pages}{2323--2326}.
\bibitem[{Rómoli et~al.(2016)Rómoli, Amicarelli, Ortiz, Scaglia \& di~Sciascio}]{roam16}
\bibinfo{author}{Rómoli, S.}, \bibinfo{author}{Amicarelli, A.~N.}, \bibinfo{author}{Ortiz, O.~A.}, \bibinfo{author}{Scaglia, G. J.~E.}, \& \bibinfo{author}{di~Sciascio, F.} (\bibinfo{year}{2016}).
\newblock \bibinfo{title}{Nonlinear control of the dissolved oxygen concentration integrated with a biomass estimator for production of bacillus thuringiensis $\delta$-endotoxins}.
\newblock {\it \bibinfo{journal}{Computers and Chemical Engineering}\/},  {\it \bibinfo{volume}{93}\/}, \bibinfo{pages}{13 – 24}.
\bibitem[{Sagi \& Rokach(2018)}]{saro18}
\bibinfo{author}{Sagi, O.}, \& \bibinfo{author}{Rokach, L.} (\bibinfo{year}{2018}).
\newblock \bibinfo{title}{Ensemble learning: A survey}.
\newblock {\it \bibinfo{journal}{WIREs Data Mining and Knowledge Discovery}\/},  {\it \bibinfo{volume}{8}\/}, \bibinfo{pages}{e1249}.
\bibitem[{Sakoe \& Chiba(1978)}]{sach78}
\bibinfo{author}{Sakoe, H.}, \& \bibinfo{author}{Chiba, S.} (\bibinfo{year}{1978}).
\newblock \bibinfo{title}{Dynamic programming algorithm optimization for spoken word recognition}.
\newblock {\it \bibinfo{journal}{IEEE Transactions on Acoustics, Speech, and Signal Processing}\/},  {\it \bibinfo{volume}{26}\/}, \bibinfo{pages}{43--49}.
\bibitem[{Salam et~al.(2021)Salam, Azar, Elgendy \& Fouad}]{saaz21}
\bibinfo{author}{Salam, M.~A.}, \bibinfo{author}{Azar, A.~T.}, \bibinfo{author}{Elgendy, M.~S.}, \& \bibinfo{author}{Fouad, K.~M.} (\bibinfo{year}{2021}).
\newblock \bibinfo{title}{The effect of different dimensionality reduction techniques on machine learning overfitting problem}.
\newblock {\it \bibinfo{journal}{International Journal of Advanced Computer Science and Applications}\/},  {\it \bibinfo{volume}{12}\/}.
\bibitem[{Salvador et~al.(2016{\natexlab{a}})Salvador, Budka \& Gabrys}]{sabu16a}
\bibinfo{author}{Salvador, M.~M.}, \bibinfo{author}{Budka, M.}, \& \bibinfo{author}{Gabrys, B.} (\bibinfo{year}{2016}{\natexlab{a}}).
\newblock \bibinfo{title}{Adapting multicomponent predictive systems using hybrid adaptation strategies with auto-weka in process industry}.
\newblock In {\it \bibinfo{booktitle}{Workshop on Automatic Machine Learning}\/} (pp. \bibinfo{pages}{48--57}).
\bibitem[{Salvador et~al.(2016{\natexlab{b}})Salvador, Budka \& Gabrys}]{sabu16}
\bibinfo{author}{Salvador, M.~M.}, \bibinfo{author}{Budka, M.}, \& \bibinfo{author}{Gabrys, B.} (\bibinfo{year}{2016}{\natexlab{b}}).
\newblock \bibinfo{title}{Towards automatic composition of multicomponent predictive systems}.
\newblock In {\it \bibinfo{booktitle}{Lecture Notes in Computer Science}\/} (pp. \bibinfo{pages}{27--39}).
\newblock \bibinfo{publisher}{Springer International Publishing}.
\bibitem[{Salvador et~al.(2019)Salvador, Budka \& Gabrys}]{sabu19}
\bibinfo{author}{Salvador, M.~M.}, \bibinfo{author}{Budka, M.}, \& \bibinfo{author}{Gabrys, B.} (\bibinfo{year}{2019}).
\newblock \bibinfo{title}{Automatic composition and optimization of multicomponent predictive systems with an extended auto-{WEKA}}.
\newblock {\it \bibinfo{journal}{{IEEE} Transactions on Automation Science and Engineering}\/},  {\it \bibinfo{volume}{16}\/}, \bibinfo{pages}{946--959}.
\bibitem[{Sandoval et~al.(2023)Sandoval, Petsagkourakis \& del Rio-Chanona}]{sape23}
\bibinfo{author}{Sandoval, I.~O.}, \bibinfo{author}{Petsagkourakis, P.}, \& \bibinfo{author}{del Rio-Chanona, E.~A.} (\bibinfo{year}{2023}).
\newblock \bibinfo{title}{Neural odes as feedback policies for nonlinear optimal control}.
\newblock {\it \bibinfo{journal}{IFAC-PapersOnLine}\/},  {\it \bibinfo{volume}{56}\/}, \bibinfo{pages}{4816--4821}.
\bibitem[{Saptoro(2014)}]{sa14}
\bibinfo{author}{Saptoro, A.} (\bibinfo{year}{2014}).
\newblock \bibinfo{title}{State of the art in the development of adaptive soft sensors based on just-in-time models}.
\newblock {\it \bibinfo{journal}{Procedia Chemistry}\/},  {\it \bibinfo{volume}{9}\/}, \bibinfo{pages}{226--234}.
\bibitem[{Scriven et~al.(2023)Scriven, Kedziora, Musial \& Gabrys}]{scke23}
\bibinfo{author}{Scriven, A.}, \bibinfo{author}{Kedziora, D.~J.}, \bibinfo{author}{Musial, K.}, \& \bibinfo{author}{Gabrys, B.} (\bibinfo{year}{2023}).
\newblock \bibinfo{title}{The technological emergence of automl: A survey of performant software and applications in the context of industry}.
\newblock {\it \bibinfo{journal}{Foundations and Trends® in Information Systems}\/},  {\it \bibinfo{volume}{7}\/}, \bibinfo{pages}{1--252}.
\bibitem[{Settles(2009)}]{se09}
\bibinfo{author}{Settles, B.} (\bibinfo{year}{2009}).
\newblock {\it \bibinfo{title}{Active Learning Literature Survey}\/}.
\newblock \bibinfo{type}{Computer Sciences Technical Report} \bibinfo{number}{1648} University of Wisconsin--Madison.
\bibitem[{Shahriari et~al.(2016)Shahriari, Swersky, Wang, Adams \& de~Freitas}]{shsw16}
\bibinfo{author}{Shahriari, B.}, \bibinfo{author}{Swersky, K.}, \bibinfo{author}{Wang, Z.}, \bibinfo{author}{Adams, R.~P.}, \& \bibinfo{author}{de~Freitas, N.} (\bibinfo{year}{2016}).
\newblock \bibinfo{title}{Taking the human out of the loop: A review of bayesian optimization}.
\newblock {\it \bibinfo{journal}{Proceedings of the IEEE}\/},  {\it \bibinfo{volume}{104}\/}, \bibinfo{pages}{148--175}.
\bibitem[{Shi \& Xiong(2018)}]{shxi18}
\bibinfo{author}{Shi, X.}, \& \bibinfo{author}{Xiong, W.} (\bibinfo{year}{2018}).
\newblock \bibinfo{title}{Approximate linear dependence criteria with active learning for smart soft sensor design}.
\newblock {\it \bibinfo{journal}{Chemometrics and Intelligent Laboratory Systems}\/},  {\it \bibinfo{volume}{180}\/}, \bibinfo{pages}{88 – 95}.
\bibitem[{Shu et~al.(2018)Shu, Xu \& Meng}]{shxu18}
\bibinfo{author}{Shu, J.}, \bibinfo{author}{Xu, Z.}, \& \bibinfo{author}{Meng, D.} (\bibinfo{year}{2018}).
\newblock \bibinfo{title}{Small sample learning in big data era}.
\newblock {\it \bibinfo{journal}{arXiv preprint arXiv:1808.04572}\/}, .
\bibitem[{Sibley et~al.(2020)Sibley, Woodhams, Hoehse \& Zoro}]{siwo20}
\bibinfo{author}{Sibley, M.}, \bibinfo{author}{Woodhams, A.}, \bibinfo{author}{Hoehse, M.}, \& \bibinfo{author}{Zoro, B.} (\bibinfo{year}{2020}).
\newblock \bibinfo{title}{Novel integrated raman spectroscopy technology for minibioreactors}.
\newblock {\it \bibinfo{journal}{BioProcess Int}\/},  {\it \bibinfo{volume}{18}\/}, \bibinfo{pages}{9}.
\bibitem[{Siegl et~al.(2022)Siegl, Brunner, Geier \& Becker}]{sibr22}
\bibinfo{author}{Siegl, M.}, \bibinfo{author}{Brunner, V.}, \bibinfo{author}{Geier, D.}, \& \bibinfo{author}{Becker, T.} (\bibinfo{year}{2022}).
\newblock \bibinfo{title}{Ensemble-based adaptive soft sensor for fault-tolerant biomass monitoring}.
\newblock {\it \bibinfo{journal}{Engineering in Life Sciences}\/},  {\it \bibinfo{volume}{22}\/}, \bibinfo{pages}{229 – 241}.
\bibitem[{Siegl et~al.(2023)Siegl, Kämpf, Geier, Andreeßen, Max, Zavrel \& Becker}]{sika23}
\bibinfo{author}{Siegl, M.}, \bibinfo{author}{Kämpf, M.}, \bibinfo{author}{Geier, D.}, \bibinfo{author}{Andreeßen, B.}, \bibinfo{author}{Max, S.}, \bibinfo{author}{Zavrel, M.}, \& \bibinfo{author}{Becker, T.} (\bibinfo{year}{2023}).
\newblock \bibinfo{title}{Generalizability of soft sensors for bioprocesses through similarity analysis and phase-dependent recalibration}.
\newblock {\it \bibinfo{journal}{Sensors}\/},  {\it \bibinfo{volume}{23}\/}.
\bibitem[{Simon \& Nazmul~Karim(2001)}]{sina01}
\bibinfo{author}{Simon, L.}, \& \bibinfo{author}{Nazmul~Karim, M.} (\bibinfo{year}{2001}).
\newblock \bibinfo{title}{Probabilistic neural networks using bayesian decision strategies and a modified gompertz model for growth phase classification in the batch culture of bacillus subtilis}.
\newblock {\it \bibinfo{journal}{Biochemical Engineering Journal}\/},  {\it \bibinfo{volume}{7}\/}, \bibinfo{pages}{41 – 48}.
\bibitem[{Sinner et~al.(2022)Sinner, Stiegler, Goldbeck, Seibold, Herwig \& Kager}]{sist22}
\bibinfo{author}{Sinner, P.}, \bibinfo{author}{Stiegler, M.}, \bibinfo{author}{Goldbeck, O.}, \bibinfo{author}{Seibold, G.~M.}, \bibinfo{author}{Herwig, C.}, \& \bibinfo{author}{Kager, J.} (\bibinfo{year}{2022}).
\newblock \bibinfo{title}{Online estimation of changing metabolic capacities in continuous corynebacterium glutamicum cultivations growing on a complex sugar mixture}.
\newblock {\it \bibinfo{journal}{Biotechnology and Bioengineering}\/},  {\it \bibinfo{volume}{119}\/}, \bibinfo{pages}{575 – 590}.
\bibitem[{St{\^\i}ng{\u{a}} \& Petre(2018)}]{stpe18}
\bibinfo{author}{St{\^\i}ng{\u{a}}, F.}, \& \bibinfo{author}{Petre, E.} (\bibinfo{year}{2018}).
\newblock \bibinfo{title}{Estimation based control strategies for an aerobic bioprocess}.
\newblock In {\it \bibinfo{booktitle}{2018 22nd International Conference on System Theory, Control and Computing (ICSTCC)}\/} (pp. \bibinfo{pages}{218--223}).
\newblock \bibinfo{organization}{IEEE}.
\bibitem[{Stonier et~al.(2011)Stonier, Pain, Westlake, Hutchinson, Thornhill \& Farid}]{stpa11}
\bibinfo{author}{Stonier, A.}, \bibinfo{author}{Pain, D.}, \bibinfo{author}{Westlake, A.}, \bibinfo{author}{Hutchinson, N.}, \bibinfo{author}{Thornhill, N.~F.}, \& \bibinfo{author}{Farid, S.~S.} (\bibinfo{year}{2011}).
\newblock \bibinfo{title}{Integration of stochastic simulation with advanced multivariate and visualisation analyses for rapid prediction of facility fit issues in biopharmaceutical processes}.
\newblock {\it \bibinfo{journal}{Computer Aided Chemical Engineering}\/},  {\it \bibinfo{volume}{29}\/}, \bibinfo{pages}{1356 – 1360}.
\bibitem[{Streefland et~al.(2013)Streefland, Martens, Beuvery \& Wijffels}]{stma13}
\bibinfo{author}{Streefland, M.}, \bibinfo{author}{Martens, D.}, \bibinfo{author}{Beuvery, E.}, \& \bibinfo{author}{Wijffels, R.} (\bibinfo{year}{2013}).
\newblock \bibinfo{title}{Process analytical technology (pat) tools for the cultivation step in biopharmaceutical production}.
\newblock {\it \bibinfo{journal}{Engineering in Life Sciences}\/},  {\it \bibinfo{volume}{13}\/}.
\bibitem[{Sugiyama(2006)}]{su06}
\bibinfo{author}{Sugiyama, M.} (\bibinfo{year}{2006}).
\newblock \bibinfo{title}{Local fisher discriminant analysis for supervised dimensionality reduction}.
\newblock In {\it \bibinfo{booktitle}{Proceedings of the 23rd International Conference on Machine Learning}\/} (p. \bibinfo{pages}{905–912}).
\newblock \bibinfo{publisher}{Association for Computing Machinery}.
\bibitem[{Sun \& Ge(2022)}]{suge22}
\bibinfo{author}{Sun, Q.}, \& \bibinfo{author}{Ge, Z.} (\bibinfo{year}{2022}).
\newblock \bibinfo{title}{Gated stacked target-related autoencoder: A novel deep feature extraction and layerwise ensemble method for industrial soft sensor application}.
\newblock {\it \bibinfo{journal}{IEEE Transactions on Cybernetics}\/},  {\it \bibinfo{volume}{52}\/}, \bibinfo{pages}{3457 – 3468}.
\bibitem[{Syu \& Hou(1997)}]{syho97}
\bibinfo{author}{Syu, M.-J.}, \& \bibinfo{author}{Hou, C.-L.} (\bibinfo{year}{1997}).
\newblock \bibinfo{title}{A neural network study on the dynamic identification of a fermentation system}.
\newblock {\it \bibinfo{journal}{Bioprocess Engineering}\/},  {\it \bibinfo{volume}{17}\/}, \bibinfo{pages}{203--213}.
\bibitem[{Tang et~al.(2018)Tang, Li \& Xi}]{tali18}
\bibinfo{author}{Tang, Q.}, \bibinfo{author}{Li, D.}, \& \bibinfo{author}{Xi, Y.} (\bibinfo{year}{2018}).
\newblock \bibinfo{title}{A new active learning strategy for soft sensor modeling based on feature reconstruction and uncertainty evaluation}.
\newblock {\it \bibinfo{journal}{Chemometrics and Intelligent Laboratory Systems}\/},  {\it \bibinfo{volume}{172}\/}, \bibinfo{pages}{43 – 51}.
\bibitem[{Tenenbaum et~al.(2000)Tenenbaum, de~Silva \& Langford}]{tesi00}
\bibinfo{author}{Tenenbaum, J.~B.}, \bibinfo{author}{de~Silva, V.}, \& \bibinfo{author}{Langford, J.~C.} (\bibinfo{year}{2000}).
\newblock \bibinfo{title}{A global geometric framework for nonlinear dimensionality reduction}.
\newblock {\it \bibinfo{journal}{Science}\/},  {\it \bibinfo{volume}{290}\/}, \bibinfo{pages}{2319--2323}.
\bibitem[{Tharwat \& Schenck(2023)}]{thsc23}
\bibinfo{author}{Tharwat, A.}, \& \bibinfo{author}{Schenck, W.} (\bibinfo{year}{2023}).
\newblock \bibinfo{title}{A survey on active learning: State-of-the-art, practical challenges and research directions}.
\newblock {\it \bibinfo{journal}{Mathematics}\/},  {\it \bibinfo{volume}{11}\/}.
\bibitem[{Tibshirani(1996)}]{ro96}
\bibinfo{author}{Tibshirani, R.} (\bibinfo{year}{1996}).
\newblock \bibinfo{title}{Regression shrinkage and selection via the lasso}.
\newblock {\it \bibinfo{journal}{Journal of the Royal Statistical Society. Series B (Methodological)}\/},  {\it \bibinfo{volume}{58}\/}, \bibinfo{pages}{267--288}.
\newblock \bibinfo{note}{Full publication date: 1996}.
\bibitem[{Tran \& Choi(2020)}]{trch20}
\bibinfo{author}{Tran, L.}, \& \bibinfo{author}{Choi, D.} (\bibinfo{year}{2020}).
\newblock \bibinfo{title}{Data augmentation for inertial sensor-based gait deep neural network}.
\newblock {\it \bibinfo{journal}{IEEE Access}\/},  {\it \bibinfo{volume}{8}\/}, \bibinfo{pages}{12364--12378}.
\bibitem[{Tulsyan et~al.(2018)Tulsyan, Garvin \& Ündey}]{tuga18}
\bibinfo{author}{Tulsyan, A.}, \bibinfo{author}{Garvin, C.}, \& \bibinfo{author}{Ündey, C.} (\bibinfo{year}{2018}).
\newblock \bibinfo{title}{Advances in industrial biopharmaceutical batch process monitoring: Machine-learning methods for small data problems}.
\newblock {\it \bibinfo{journal}{Biotechnology and Bioengineering}\/},  {\it \bibinfo{volume}{115}\/}, \bibinfo{pages}{1915--1924}.
\bibitem[{Tulsyan et~al.(2021)Tulsyan, Khodabandehlou, Wang, Schorner, Coufal \& Undey}]{tukh21}
\bibinfo{author}{Tulsyan, A.}, \bibinfo{author}{Khodabandehlou, H.}, \bibinfo{author}{Wang, T.}, \bibinfo{author}{Schorner, G.}, \bibinfo{author}{Coufal, M.}, \& \bibinfo{author}{Undey, C.} (\bibinfo{year}{2021}).
\newblock \bibinfo{title}{Spectroscopic models for real-time monitoring of cell culture processes using spatiotemporal just-in-time gaussian processes}.
\newblock {\it \bibinfo{journal}{AIChE Journal}\/},  {\it \bibinfo{volume}{67}\/}.
\bibitem[{Tulsyan et~al.(2019)Tulsyan, Schorner, Khodabandehlou, Wang, Coufal \& Undey}]{tusc19}
\bibinfo{author}{Tulsyan, A.}, \bibinfo{author}{Schorner, G.}, \bibinfo{author}{Khodabandehlou, H.}, \bibinfo{author}{Wang, T.}, \bibinfo{author}{Coufal, M.}, \& \bibinfo{author}{Undey, C.} (\bibinfo{year}{2019}).
\newblock \bibinfo{title}{A machine-learning approach to calibrate generic raman models for real-time monitoring of cell culture processes}.
\newblock {\it \bibinfo{journal}{Biotechnology and Bioengineering}\/},  {\it \bibinfo{volume}{116}\/}, \bibinfo{pages}{2575 – 2586}.
\bibitem[{Tulsyan et~al.(2020)Tulsyan, Wang, Schorner, Khodabandehlou, Coufal \& Undey}]{tuwa20}
\bibinfo{author}{Tulsyan, A.}, \bibinfo{author}{Wang, T.}, \bibinfo{author}{Schorner, G.}, \bibinfo{author}{Khodabandehlou, H.}, \bibinfo{author}{Coufal, M.}, \& \bibinfo{author}{Undey, C.} (\bibinfo{year}{2020}).
\newblock \bibinfo{title}{Automatic real-time calibration, assessment, and maintenance of generic raman models for online monitoring of cell culture processes}.
\newblock {\it \bibinfo{journal}{Biotechnology and Bioengineering}\/},  {\it \bibinfo{volume}{117}\/}, \bibinfo{pages}{406 – 416}.
\bibitem[{Um et~al.(2017)Um, Pfister, Pichler, Endo, Lang, Hirche, Fietzek \& Kuli\'{c}}]{umpf17}
\bibinfo{author}{Um, T.~T.}, \bibinfo{author}{Pfister, F. M.~J.}, \bibinfo{author}{Pichler, D.}, \bibinfo{author}{Endo, S.}, \bibinfo{author}{Lang, M.}, \bibinfo{author}{Hirche, S.}, \bibinfo{author}{Fietzek, U.}, \& \bibinfo{author}{Kuli\'{c}, D.} (\bibinfo{year}{2017}).
\newblock \bibinfo{title}{Data augmentation of wearable sensor data for parkinson’s disease monitoring using convolutional neural networks}.
\newblock In {\it \bibinfo{booktitle}{Proceedings of the 19th ACM International Conference on Multimodal Interaction}\/} (p. \bibinfo{pages}{216–220}).
\bibitem[{Valiant(1984)}]{va84}
\bibinfo{author}{Valiant, L.~G.} (\bibinfo{year}{1984}).
\newblock \bibinfo{title}{A theory of the learnable}.
\newblock In {\it \bibinfo{booktitle}{Proceedings of the Sixteenth Annual ACM Symposium on Theory of Computing}\/} STOC '84 (p. \bibinfo{pages}{436–445}).
\newblock \bibinfo{address}{New York, NY, USA}.
\bibitem[{Vapnik(1998)}]{va98}
\bibinfo{author}{Vapnik, V.~N.} (\bibinfo{year}{1998}).
\newblock {\it \bibinfo{title}{The Nature of Statistical Learning Theory}\/}.
\newblock \bibinfo{address}{New York, NY}: \bibinfo{publisher}{Springer New York}.
\bibitem[{Wang et~al.(2018{\natexlab{a}})Wang, Yu, Li, Li \& Wang}]{wayu18}
\bibinfo{author}{Wang, H.}, \bibinfo{author}{Yu, D.}, \bibinfo{author}{Li, Y.}, \bibinfo{author}{Li, Z.}, \& \bibinfo{author}{Wang, G.} (\bibinfo{year}{2018}{\natexlab{a}}).
\newblock \bibinfo{title}{Multi-label online streaming feature selection based on spectral granulation and mutual information}.
\newblock In \bibinfo{editor}{H.~S. Nguyen}, \bibinfo{editor}{Q.-T. Ha}, \bibinfo{editor}{T.~Li}, \& \bibinfo{editor}{M.~Przyby{\l}a-Kasperek} (Eds.), {\it \bibinfo{booktitle}{Rough Sets}\/} (pp. \bibinfo{pages}{215--228}).
\newblock \bibinfo{address}{Cham}: \bibinfo{publisher}{Springer International Publishing}.
\bibitem[{Wang et~al.(2018{\natexlab{b}})Wang, Tian, Yu, Liu, Zhan \& Wang}]{wati18}
\bibinfo{author}{Wang, J.}, \bibinfo{author}{Tian, F.}, \bibinfo{author}{Yu, H.}, \bibinfo{author}{Liu, C.~H.}, \bibinfo{author}{Zhan, K.}, \& \bibinfo{author}{Wang, X.} (\bibinfo{year}{2018}{\natexlab{b}}).
\newblock \bibinfo{title}{Diverse non-negative matrix factorization for multiview data representation}.
\newblock {\it \bibinfo{journal}{IEEE Transactions on Cybernetics}\/},  {\it \bibinfo{volume}{48}\/}, \bibinfo{pages}{2620--2632}.
\bibitem[{Wang \& Zhao(2020)}]{wazh20}
\bibinfo{author}{Wang, J.}, \& \bibinfo{author}{Zhao, C.} (\bibinfo{year}{2020}).
\newblock \bibinfo{title}{Mode-cloud data analytics based transfer learning for soft sensor of manufacturing industry with incremental learning ability}.
\newblock {\it \bibinfo{journal}{Control Engineering Practice}\/},  {\it \bibinfo{volume}{98}\/}.
\bibitem[{Wang et~al.(2020)Wang, Yao, Kwok \& Ni}]{waya20}
\bibinfo{author}{Wang, Y.}, \bibinfo{author}{Yao, Q.}, \bibinfo{author}{Kwok, J.~T.}, \& \bibinfo{author}{Ni, L.~M.} (\bibinfo{year}{2020}).
\newblock \bibinfo{title}{Generalizing from a few examples: A survey on few-shot learning}.
\newblock {\it \bibinfo{journal}{ACM Computing Surveys}\/},  {\it \bibinfo{volume}{53}\/}.
\bibitem[{Wang et~al.(2024)Wang, Wang, Liu, Wang, Fu, Lu, Aggarwal, Pei \& Zhou}]{wawa24}
\bibinfo{author}{Wang, Z.}, \bibinfo{author}{Wang, P.}, \bibinfo{author}{Liu, K.}, \bibinfo{author}{Wang, P.}, \bibinfo{author}{Fu, Y.}, \bibinfo{author}{Lu, C.-T.}, \bibinfo{author}{Aggarwal, C.~C.}, \bibinfo{author}{Pei, J.}, \& \bibinfo{author}{Zhou, Y.} (\bibinfo{year}{2024}).
\newblock \bibinfo{title}{A comprehensive survey on data augmentation}.
\bibitem[{Wang et~al.(2016)Wang, Zhang, Chen, Yang, Sun, Kang, Yang \& Liang}]{wazh16}
\bibinfo{author}{Wang, Z.}, \bibinfo{author}{Zhang, Y.}, \bibinfo{author}{Chen, Z.}, \bibinfo{author}{Yang, H.}, \bibinfo{author}{Sun, Y.}, \bibinfo{author}{Kang, J.}, \bibinfo{author}{Yang, Y.}, \& \bibinfo{author}{Liang, X.} (\bibinfo{year}{2016}).
\newblock \bibinfo{title}{Application of relieff algorithm to selecting feature sets for classification of high resolution remote sensing image}.
\newblock In {\it \bibinfo{booktitle}{2016 IEEE International Geoscience and Remote Sensing Symposium (IGARSS)}\/} (pp. \bibinfo{pages}{755--758}).
\bibitem[{Wen et~al.(2018)Wen, Gao, Song, Sun, Xu \& Zhu}]{wega18}
\bibinfo{author}{Wen, Q.}, \bibinfo{author}{Gao, J.}, \bibinfo{author}{Song, X.}, \bibinfo{author}{Sun, L.}, \bibinfo{author}{Xu, H.}, \& \bibinfo{author}{Zhu, S.} (\bibinfo{year}{2018}).
\newblock \bibinfo{title}{Robuststl: A robust seasonal-trend decomposition algorithm for long time series}.
\bibitem[{Widmer \& Kubat(1996)}]{wiku96}
\bibinfo{author}{Widmer, G.}, \& \bibinfo{author}{Kubat, M.} (\bibinfo{year}{1996}).
\newblock \bibinfo{title}{Learning in the presence of concept drift and hidden contexts}.
\newblock {\it \bibinfo{journal}{Machine Learning}\/},  {\it \bibinfo{volume}{23}\/}, \bibinfo{pages}{69--101}.
\bibitem[{Wold(1975)}]{wo75}
\bibinfo{author}{Wold, H.} (\bibinfo{year}{1975}).
\newblock \bibinfo{title}{Soft modelling by latent variables: The non-linear iterative partial least squares (nipals) approach}.
\newblock {\it \bibinfo{journal}{Journal of applied probability}\/},  {\it \bibinfo{volume}{12}\/}, \bibinfo{pages}{117--142}.
\bibitem[{Wu et~al.(2017)Wu, Chan \& Chen}]{wuch17}
\bibinfo{author}{Wu, Q.-Y.}, \bibinfo{author}{Chan, L. L.~T.}, \& \bibinfo{author}{Chen, J.} (\bibinfo{year}{2017}).
\newblock \bibinfo{title}{Active learning dynamic soft sensor with forward-update scheme}.
\newblock In {\it \bibinfo{booktitle}{Proceedings of the 6th International Symposium on Advanced Control of Industrial Processes (AdCONIP)}\/} (pp. \bibinfo{pages}{306--311}).
\bibitem[{Xiaoyong et~al.(2022)Xiaoyong, Yanchao, Yi \& Dexian}]{xiya22}
\bibinfo{author}{Xiaoyong, G.}, \bibinfo{author}{Yanchao, L.}, \bibinfo{author}{Yi, X.}, \& \bibinfo{author}{Dexian, H.} (\bibinfo{year}{2022}).
\newblock \bibinfo{title}{Novel multimodal data fusion soft sensor modeling framework based on meta-learning networks for complex chemical process}.
\newblock {\it \bibinfo{journal}{IFAC-PapersOnLine}\/},  {\it \bibinfo{volume}{55}\/}, \bibinfo{pages}{839--844}.
\bibitem[{Xie et~al.(2022)Xie, Wang, Li, Xie \& Auclair}]{xiwa22}
\bibinfo{author}{Xie, W.}, \bibinfo{author}{Wang, B.}, \bibinfo{author}{Li, C.}, \bibinfo{author}{Xie, D.}, \& \bibinfo{author}{Auclair, J.} (\bibinfo{year}{2022}).
\newblock \bibinfo{title}{Interpretable biomanufacturing process risk and sensitivity analyses for quality-by-design and stability control}.
\newblock {\it \bibinfo{journal}{Naval Research Logistics}\/},  {\it \bibinfo{volume}{69}\/}, \bibinfo{pages}{461 – 483}.
\bibitem[{Yamada \& Kaneko(2022)}]{yaka22}
\bibinfo{author}{Yamada, N.}, \& \bibinfo{author}{Kaneko, H.} (\bibinfo{year}{2022}).
\newblock \bibinfo{title}{Adaptive soft sensor based on transfer learning and ensemble learning for multiple process states}.
\newblock {\it \bibinfo{journal}{Analytical Science Advances}\/},  {\it \bibinfo{volume}{3}\/}, \bibinfo{pages}{205 – 211}.
\newblock \bibinfo{note}{; All Open Access, Bronze Open Access}.
\bibitem[{Yang et~al.(2014)Yang, Farid \& Thornhill}]{yafa14}
\bibinfo{author}{Yang, Y.}, \bibinfo{author}{Farid, S.~S.}, \& \bibinfo{author}{Thornhill, N.~F.} (\bibinfo{year}{2014}).
\newblock \bibinfo{title}{Data mining for rapid prediction of facility fit and debottlenecking of biomanufacturing facilities}.
\newblock {\it \bibinfo{journal}{Journal of Biotechnology}\/},  {\it \bibinfo{volume}{179}\/}, \bibinfo{pages}{17 – 25}.
\bibitem[{Yenice et~al.(2018)Yenice, Adhikari, Wong, Aksakalli, Gumus \& Abbasi}]{yead18}
\bibinfo{author}{Yenice, Z.~D.}, \bibinfo{author}{Adhikari, N.}, \bibinfo{author}{Wong, Y.~K.}, \bibinfo{author}{Aksakalli, V.}, \bibinfo{author}{Gumus, A.~T.}, \& \bibinfo{author}{Abbasi, B.} (\bibinfo{year}{2018}).
\newblock \bibinfo{title}{Spsa-fsr: Simultaneous perturbation stochastic approximation for feature selection and ranking}.
\newblock {\it \bibinfo{journal}{arXiv preprint arXiv:1804.05589}\/}, .
\bibitem[{Yu(2012)}]{yu12}
\bibinfo{author}{Yu, J.} (\bibinfo{year}{2012}).
\newblock \bibinfo{title}{A bayesian inference based two-stage support vector regression framework for soft sensor development in batch bioprocesses}.
\newblock {\it \bibinfo{journal}{Computers and Chemical Engineering}\/},  {\it \bibinfo{volume}{41}\/}, \bibinfo{pages}{134 – 144}.
\bibitem[{Yuan et~al.(2014)Yuan, Ge \& Song}]{yuge14}
\bibinfo{author}{Yuan, X.}, \bibinfo{author}{Ge, Z.}, \& \bibinfo{author}{Song, Z.} (\bibinfo{year}{2014}).
\newblock \bibinfo{title}{Locally weighted kernel principal component regression model for soft sensing of nonlinear time-variant processes}.
\newblock {\it \bibinfo{journal}{Industrial \& Engineering Chemistry Research}\/},  {\it \bibinfo{volume}{53}\/}, \bibinfo{pages}{13736--13749}.
\bibitem[{Zhang et~al.(2021)Zhang, Wang, Castan, Hjalmarsson \& Chotteau}]{zhwa21}
\bibinfo{author}{Zhang, L.}, \bibinfo{author}{Wang, M.}, \bibinfo{author}{Castan, A.}, \bibinfo{author}{Hjalmarsson, H.}, \& \bibinfo{author}{Chotteau, V.} (\bibinfo{year}{2021}).
\newblock \bibinfo{title}{Probabilistic model by bayesian network for the prediction of antibody glycosylation in perfusion and fed-batch cell cultures}.
\newblock {\it \bibinfo{journal}{Biotechnology and Bioengineering}\/},  {\it \bibinfo{volume}{118}\/}, \bibinfo{pages}{3447 – 3459}.
\bibitem[{Zhao et~al.(2007)Zhao, Wang \& Jia}]{zhwa07}
\bibinfo{author}{Zhao, C.}, \bibinfo{author}{Wang, F.}, \& \bibinfo{author}{Jia, M.} (\bibinfo{year}{2007}).
\newblock \bibinfo{title}{Dissimilarity analysis based batch process monitoring using moving windows}.
\newblock {\it \bibinfo{journal}{AIChE journal}\/},  {\it \bibinfo{volume}{53}\/}, \bibinfo{pages}{1267--1277}.
\bibitem[{Zhou et~al.(2019)Zhou, Zhang, Zhang \& Liu}]{zhzh19}
\bibinfo{author}{Zhou, H.}, \bibinfo{author}{Zhang, Y.}, \bibinfo{author}{Zhang, Y.}, \& \bibinfo{author}{Liu, H.} (\bibinfo{year}{2019}).
\newblock \bibinfo{title}{Feature selection based on conditional mutual information: minimum conditional relevance and minimum conditional redundancy}.
\newblock {\it \bibinfo{journal}{Applied Intelligence}\/},  {\it \bibinfo{volume}{49}\/}, \bibinfo{pages}{883--896}.
\bibitem[{Zhu et~al.(2018)Zhu, Ge \& Song}]{zhge18}
\bibinfo{author}{Zhu, J.}, \bibinfo{author}{Ge, Z.}, \& \bibinfo{author}{Song, Z.} (\bibinfo{year}{2018}).
\newblock \bibinfo{title}{Quantum statistic based semi-supervised learning approach for industrial soft sensor development}.
\newblock {\it \bibinfo{journal}{Control Engineering Practice}\/},  {\it \bibinfo{volume}{74}\/}, \bibinfo{pages}{144 – 152}.
\bibitem[{Zhu et~al.(2021)Zhu, Hou, Chen, Gao, Xu \& He}]{zhho21}
\bibinfo{author}{Zhu, Q.-X.}, \bibinfo{author}{Hou, K.-R.}, \bibinfo{author}{Chen, Z.-S.}, \bibinfo{author}{Gao, Z.-S.}, \bibinfo{author}{Xu, Y.}, \& \bibinfo{author}{He, Y.-L.} (\bibinfo{year}{2021}).
\newblock \bibinfo{title}{Novel virtual sample generation using conditional gan for developing soft sensor with small data}.
\newblock {\it \bibinfo{journal}{Engineering Applications of Artificial Intelligence}\/},  {\it \bibinfo{volume}{106}\/}, \bibinfo{pages}{104497}.
\bibitem[{Zhu et~al.(2022)Zhu, Xu, Xu \& He}]{zhxu22}
\bibinfo{author}{Zhu, Q.-X.}, \bibinfo{author}{Xu, T.-X.}, \bibinfo{author}{Xu, Y.}, \& \bibinfo{author}{He, Y.-L.} (\bibinfo{year}{2022}).
\newblock \bibinfo{title}{Improved virtual sample generation method using enhanced conditional generative adversarial networks with cycle structures for soft sensors with limited data}.
\newblock {\it \bibinfo{journal}{Industrial and Engineering Chemistry Research}\/},  {\it \bibinfo{volume}{61}\/}, \bibinfo{pages}{530 – 540}.

\end{thebibliography}

\end{document}